\documentclass[twoside,11pt]{article}

\usepackage{microtype}
\usepackage{graphicx}
\usepackage{epsfig}
\usepackage{booktabs} 
\usepackage{amsmath}
\usepackage{amssymb}
\usepackage{bm}
\usepackage{color}
\usepackage[table,xcdraw]{xcolor}
\usepackage{algorithm}
\usepackage{algorithmic}
\usepackage[labelformat=simple]{subcaption}
\usepackage{multirow}
\usepackage{graphbox} 
\usepackage{array}

\usepackage[abbrvbib, preprint]{jmlr2e}

\newcolumntype{C}[1]{>{\centering\arraybackslash}p{#1}}

\newcommand{\eg}{\textit{e.g., }}
\newcommand{\ie}{\textit{i.e., }}
\newcommand{\etal}{\textit{et al. }}

\jmlrheading{1}{2021}{1-48}{4/00}{10/00}{meila00a}{Kaixuan Wei, Angelica Aviles-Rivero, Jingwei Liang, Ying Fu$^*$, Hua Huang and Carola-Bibiane Sch\"{o}nlieb; $^*$Correspondence}

\ShortHeadings{TFPnP: Tuning-free Plug-and-Play Proximal Algorithm}{Wei, Aviles-Rivero, Liang, Fu, Huang and Sch\"{o}nlieb}
\firstpageno{1}

\begin{document}

\renewcommand{\arraystretch}{1.08}

\title{TFPnP: Tuning-free Plug-and-Play Proximal Algorithms with Applications to Inverse Imaging Problems}

\author{\name Kaixuan Wei \email kaixuan\_wei@bit.edu.cn \\
       \addr School of Computer Science and Technology\\
       Beijing Institute of Technology, Beijing, China       
       \AND
       \name Angelica Aviles-Rivero \email ai323@cam.ac.uk \\
       \addr Department of Pure Mathematics and Mathematical Statistics\\
       University of Cambridge, Cambridge, United Kingdom       
       \AND
       \name Jingwei Liang \email jingwei.liang@sjtu.edu.cn \\
       \addr Institute of Natural Sciences and School of Mathematical Sciences\\
       Shanghai Jiao Tong University, Shanghai, China
       \AND       
       \name Ying Fu$^*$ \email fuying@bit.edu.cn \\
       \addr School of Computer Science and Technology\\
       Beijing Institute of Technology, Beijing, China       
       \AND
       \name Hua Huang \email huahuang@bnu.edu.cn \\
       \addr School of Artificial Intelligence\\
        Beijing Normal University, Beijing, China       
       \AND
       \name Carola-Bibiane Sch\"{o}nlieb \email cbs31@cam.ac.uk \\
       \addr Department of Applied Mathematics and Theoretical Physics\\
       University of Cambridge, Cambridge, United Kingdom        
       }

\editor{ICML 2020 Award Special Issue}

\maketitle

\begin{abstract}
Plug-and-Play (PnP) is a non-convex optimization framework that combines proximal algorithms, for example, the alternating direction method of multipliers (ADMM), with advanced denoising priors.  
Over the past few years, great empirical success has been obtained by PnP algorithms, especially for the ones that integrate deep learning-based denoisers. However, a key challenge of PnP approaches is the need for manual parameter tweaking as it is essential to obtain high-quality results across the high discrepancy in imaging conditions and varying scene content.
In this work, we present a class of tuning-free PnP proximal algorithms that can determine parameters such as denoising strength, termination time, and other optimization-specific parameters automatically. 
A core part of our approach is a policy network for automated parameter search which can be effectively learned via a mixture of model-free and model-based deep reinforcement learning strategies.
We demonstrate, through rigorous numerical and visual experiments, that the learned policy can customize parameters to different settings, and is often more efficient and effective than existing handcrafted criteria. Moreover, we discuss several practical considerations of  PnP denoisers, which together with our learned policy yield state-of-the-art results. This advanced performance is prevalent on both linear and nonlinear exemplar inverse imaging problems, and in particular shows promising results on compressed sensing MRI, sparse-view CT, single-photon imaging, and phase retrieval.
\end{abstract}


\section{Introduction}


\noindent

Over the past decades advances in imaging technologies, \eg data acquisition tools and image processing units, have pushed the development of inverse problems and mathematical image analysis. 
A common challenge in modern inverse imaging problems is their ill-posedness \citep{tikhonov1987ill,engl1996regularization,chambolle2016introduction,mccann2017convolutional,jin2017deep,lucas2018using,arridge2019solving,ongie2020deep}, 
often due to measurement noise and undersampling.
To overcome the challenges imposed by ill-posedness a common strategy is to incorporate prior knowledge in the problem in the form of desired solution properties. This leads to so-called regularization methods \citep{engl1996regularization,benning2018modern}.

A typical forward model in an inverse imaging problem takes the following form
\begin{equation}\label{eq:observation}
y = \mathcal{A} \left( x \right) + \varepsilon , 
\end{equation}
where $\mathcal{A}: \mathbb{R}^N \to \mathbb{R}^M$ is a (typically linear but possibly nonlinear) forward measurement operator, $x \in \mathbb{R}^N$ is the unknown target image of interest which cannot be accessed directly and $\varepsilon \in \mathbb{R}^M$ denotes  measurement noise, \eg white Gaussian noise. 
Depending on the form of the measurement operator $\mathcal{A}$, \eqref{eq:observation} covers numerous imaging applications, including magnetic resonance imaging (MRI) \citep{fessler2010model}, computed tomography (CT) \citep{elbakri2002segmentation}, microscopy \citep{aguet2008model,zheng2013wide-field} and inverse scattering \citep{katz2014non,metzler2017coherent}, to name a few.

To reconstruct $x$ from  noise-contaminated measurements $y$, a routine approach is to approximate $x$ by the solution of the following minimization problem
\begin{equation}\label{eq:min_D}
\min_{x\in\mathbb{R}^N}~ \mathcal{D} \left( x \right) ,
\end{equation}
where $\mathcal{D}$ is called data-fidelity term and measures the discrepancy between the recovered $x$ and the measurement $y$. For instance, when $\varepsilon$ is white Gaussian noise, $\mathcal{D}$ can take the simple  form 
$$\mathcal{D} \left( x \right) = \frac{1}{2} \|\mathcal{A} \left( x \right) - y\|^2 ,$$
and \eqref{eq:min_D} constitutes a least square fitting problem. 
However, the solution obtained by \eqref{eq:min_D} could be meaningless due to the ill-posedness of $\mathcal{A}$. For example, when $\mathcal{A}$ is linear and has a non-trivial null space, the minimization of $\mathcal{D}(x)$ admits infinitely many solutions. 

To overcome the aforementioned challenge, regularization can be added to modify \eqref{eq:min_D}, which is determined by prior information on $x$, promoting certain structures to the solution. For instance, one can consider total variation (TV) regularization \citep{rudin1992nonlinear} which promotes piece-wise constant structure and preserves edges in reconstructed images. As a result, we seek to solve the following regularized reconstruction model
\begin{equation}\label{eq:optimization}
 \mathop{\mathrm{min}}_{x \in \mathbb{R}^N} ~ \mathcal{D} \left( x \right)+ \lambda \mathcal{R} \left( x \right),
\end{equation}
where $\mathcal{R}$ is the  regularization term, and $\lambda$ is a  positive  parameter balancing $\mathcal{R}$ against $\mathcal{D}$. 
In the literature, numerous regularization functions are developed, with representative examples including Tikhonov regularization \citep{tikhonov1977solutions}, total variation \citep{rudin1992nonlinear,osher2005iterative}, wavelets \citep{mallat1989theory,daubechies1992ten}, sparsity promoting dictionary \citep{liao2008sparse,mairal2009online,ravishankar2010mr}, nonlocal self-similarity \citep{mairal2009non,qu2014magnetic} and low rank regularization \citep{fazel2001rank,recht2010guaranteed,semerci2014tensor,gu2017weighted}. Besides Tikhonov regularization which is differentiable, all the others are non-smooth. We refer the reader to \citep{benning2018modern} and the references therein for further details.

\subsection{Proximal Algorithms}
The optimization problem in \eqref{eq:optimization} is usually non-smooth, due to $\mathcal{R}$,  which imposes difficulties for designing efficient numerical algorithms. 
One popular approach to solve \eqref{eq:optimization} is the class of so-called first-order methods \citep{beck2009fast,bauschke2011convex,boyd2011distributed,chambolle2011first,chambolle2016introduction,geman1995nonlinear,esser2010general}.

When $\mathcal{D}$ is smoothly differentiable with its gradient $\nabla D$ being $L$-Lipschitz continuous, \eqref{eq:optimization} can be solved by, \eg proximal gradient descent (a.k.a. Forward--Backward splitting method) \citep{lions1979splitting}, and its accelerated versions such as fast iterative shrinkage/thresholding algorithm (FISTA)~\citep{beck2009fast}. 
The standard proximal gradient descent takes the following iteration
\begin{align}\label{eq:pgm}
x_{k+1} &= \mathrm{Prox}_{\gamma \mathcal{R}} \big( x_{k} - \gamma \nabla \mathcal{D}(x_k) \big) ,
\end{align}
where $\gamma \in ]0, 2/L[$ is the step size, $\mathrm{Prox}_{\gamma \mathcal{R}}(\cdot)$ is called proximal operator \citep{parikh2014proximal} of $\mathcal{R}$ and it is defined by
\begin{equation} \label{eq:proximal}
\mathrm{Prox}_{\gamma \mathcal{R}} \left( v \right)  = \mathop{\mathrm{argmin}}_x \Big\{ \gamma \mathcal{R}(x) + \frac{1}{2} \| x - v \|^2 \Big \}.
\end{equation}
For many widely used non-smooth regularization terms, their corresponding proximal operators have closed-form expressions. For example, for $\mathcal{R} = \|\cdot\|_1$ being the $\ell_1$-norm, its proximal mapping is soft shrinkage thresholding~\citep{daubechies2004iterative}.

When $\mathcal{D}$ is non-smooth, it can no longer be handled by proximal gradient descent.  A widely adopted algorithm for this case is the alternating direction method of multipliers (ADMM) \citep{gabay1976dual}. Let $\mu > 0$, $\sigma^2 = \frac{\lambda}{\mu}$,  then one form of ADMM iteration reads
\begin{equation}\label{eq:admm}
\begin{aligned} 
&x_{k+1} = \mathrm{Prox}_{\sigma^2 \mathcal{R}} \left( z_k - u_k \right) ,\\
&z_{k+1} = \mathrm{Prox}_{\frac{1}{\mu} \mathcal{D}} \left( x_{k+1} + u_k \right) , \\
&u_{k+1} = u_k + x_{k+1} - z_{k+1} . 
\end{aligned}
\end{equation}
We refer to \citep{eckstein1992douglas,boyd2011distributed} for more discussions on ADMM and \citep{bauschke2011convex,chambolle2016introduction} for an overview on first-order methods for non-smooth optimization.

\subsection{Plug-and-Play Priors}

In traditional regularization theory, most regularizers have explicit mathematical expressions \citep{benning2018modern}. Over the past few years, implicit regularization (\ie regularization without explicit mathematical formulation) has been developed and demonstrated superior performance over the traditional ones. 

The foundation of implicit regularization is the mathematical equivalence of the general proximal operator \eqref{eq:proximal} to the regularized denoising.
As a result, 
one can replace the proximal operators $\mathrm{Prox}_{\sigma^2 \mathcal{R}}$  by any off-the-shelf denoisers $\mathcal{H}_{\sigma}$ for noise level $\sigma$, yielding a new framework coined Plug-and-Play (PnP) prior \citep{venkatakrishnan2013plug}. 
Applying this methodology to ADMM, we obtain the following PnP-ADMM scheme
\begin{align} 
&x_{k+1} 
= \mathcal{H}_{\sigma_k} \left( z_k - u_k  \right) , \label{eq:pnp-admm-1}  \\ 
&z_{k+1} = \mathrm{Prox}_{\frac{1}{\mu_k} \mathcal{D}} \left(x_{k+1} + u_k \right) , \label{eq:pnp-admm-2} \\ 
&u_{k+1} = u_k + x_{k+1} - z_{k+1} , \label{eq:pnp-admm-3}
\end{align}
where $k \in \left[0, \tau \right)$ denotes the $k$-th iteration, $\tau$ is the termination time, 
$\sigma_k$ and $\mu_k$ indicate the denoising strength (of the denoiser), and the penalty parameter used in the $k$-th iteration, respectively.

In this formulation, the regularizer $\mathcal{R}$ is implicitly defined by a PnP denoiser $\mathcal{H}_{\sigma}$, which opens a new door to leverage the vast progress made in image denoising for solving more general inverse imaging problems. Plugging the well-known image denoisers, \eg BM3D \citep{dabov2007image} and NLM \citep{buades2005non}, into the aforementioned  optimization algorithms often leads to sizeable performance gain compared to other explicitly defined regularizers, \eg total variation. 
Moreover,  PnP as a stand-alone framework can combine the  benefits of both deep learning based denoisers and optimization methods,  \eg~\citep{Zhang_2017_CVPR,Chang_2017_ICCV,Meinhardt_2017_ICCV}. These highly desirable benefits are in terms of fast and effective inference whilst circumventing the need for expensive network retraining whenever the specifics of the problem in~\eqref{eq:observation} changes.

\begin{figure}[t]
	\centering
	\setlength\tabcolsep{1pt}
	\begin{tabular}{ccccc}
		\rotatebox[origin=c]{90}{CS-MRI}  
		& \includegraphics[align=c,width=0.235\linewidth]{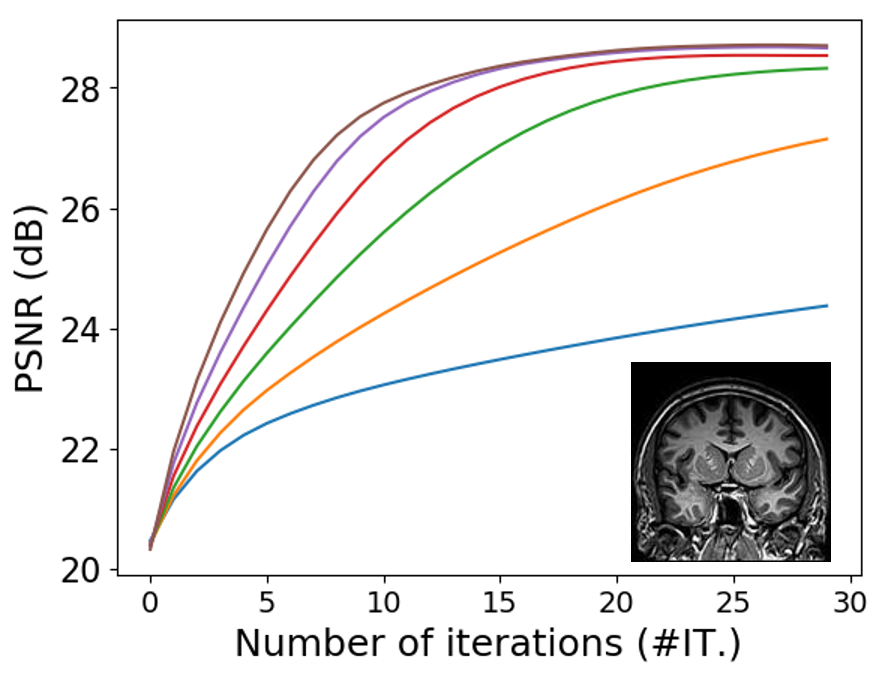}         	
		& \includegraphics[align=c,width=0.235\linewidth]{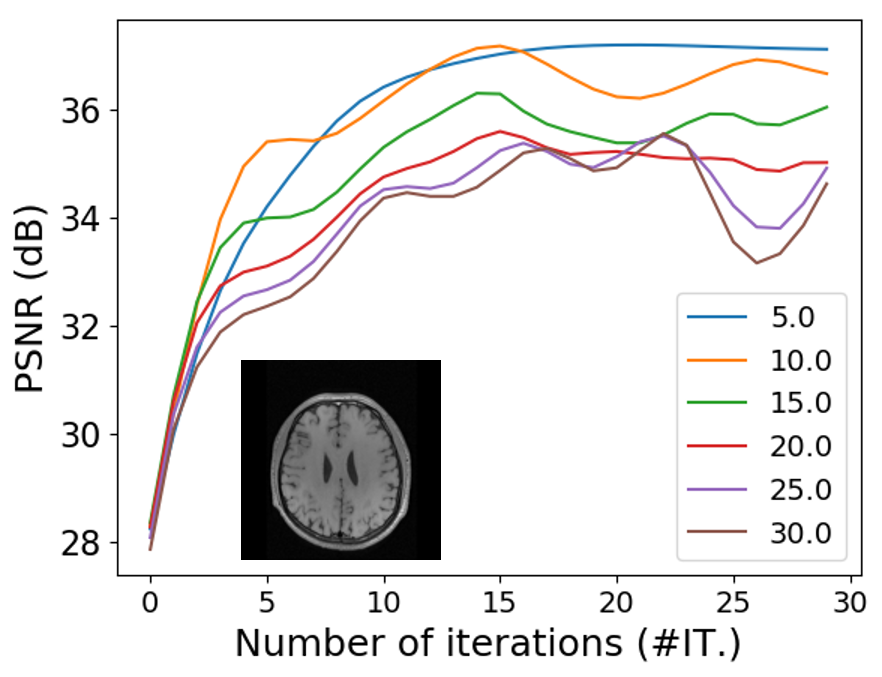} 
		& \includegraphics[align=c,width=0.235\linewidth]{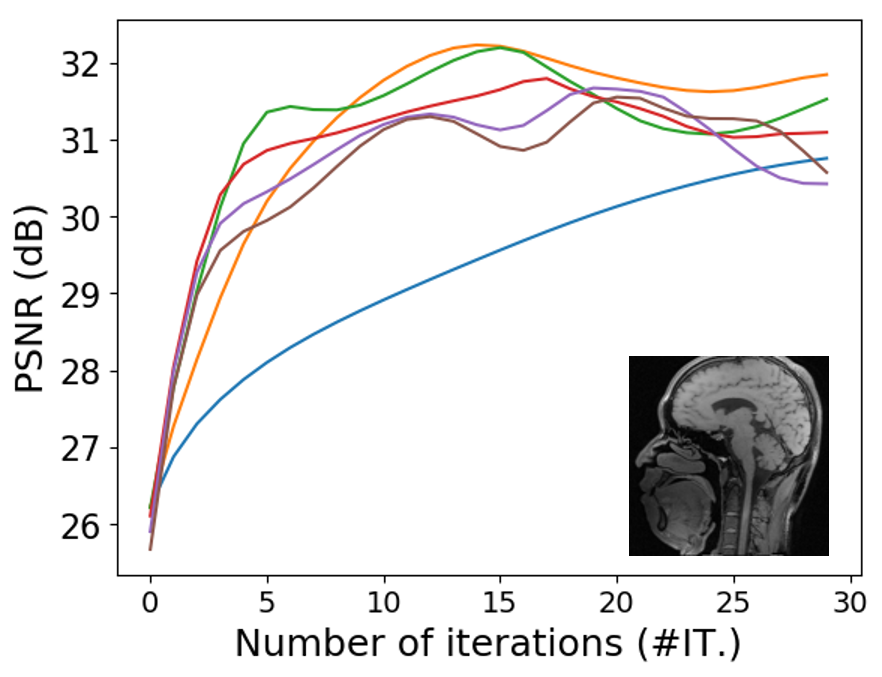} 
		& \includegraphics[align=c,width=0.235\linewidth]{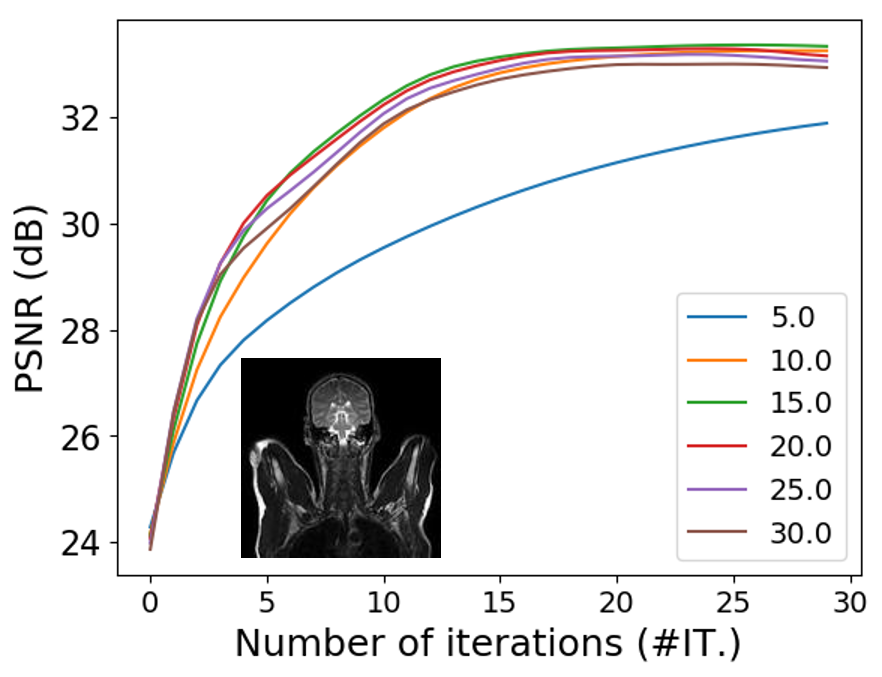}  \\
		\addlinespace[1.5pt]
		\rotatebox[origin=c]{90}{Sparse-View CT}  
		& \includegraphics[align=c,width=0.235\linewidth]{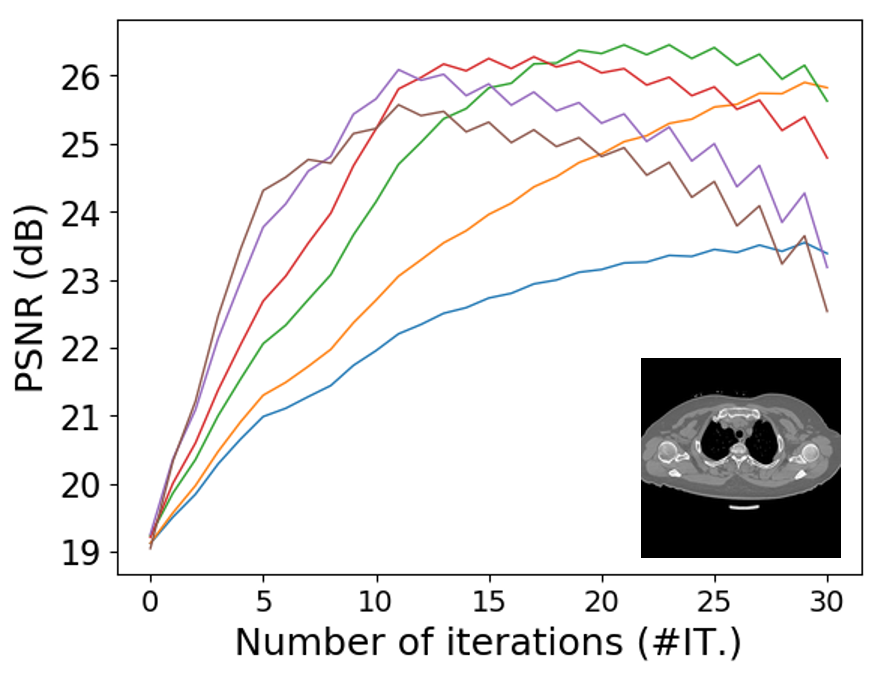}         	
		& \includegraphics[align=c,width=0.235\linewidth]{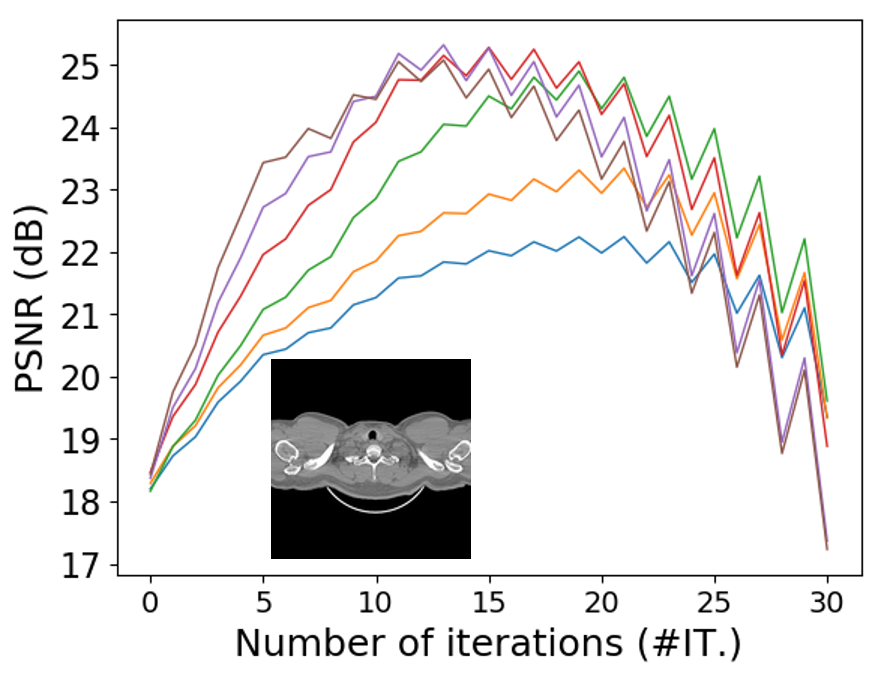} 
		& \includegraphics[align=c,width=0.235\linewidth]{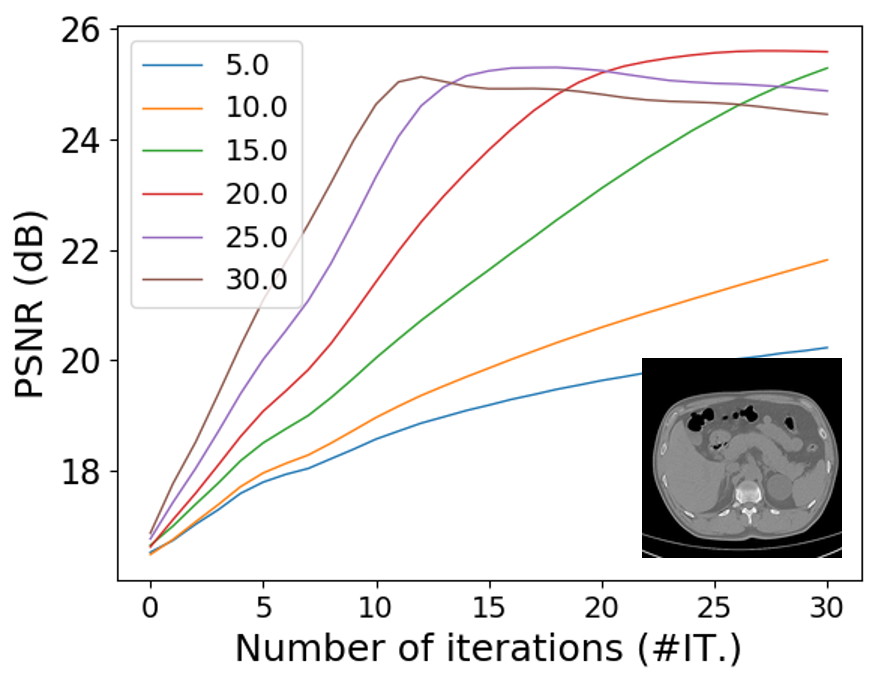} 
		& \includegraphics[align=c,width=0.235\linewidth]{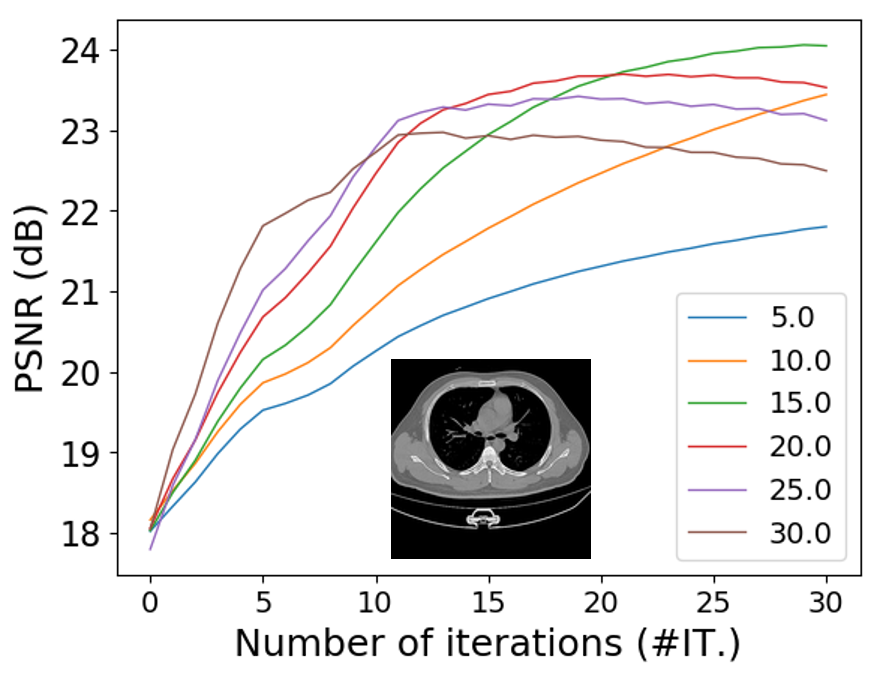}  \\
		\addlinespace[1.5pt]
	\end{tabular}
	\caption{ Performance comparison of PnP-ADMM using different denoising strengths. Plots display PSNR versus iterations on eight medical images on two applications (Compressed sensing MRI reconstruction with 20\% sampling rate and sparse-view CT reconstruction using 30 views). From the PSNR curves, one can observe that each image prefers different denoising strengths and sometimes even requires early stopping to reach the optimal performance.
	(Best viewed with zoom)
	}
	\label{fig:example}
\end{figure}

Whilst a PnP framework offers promising image recovery results, it suffers a major drawback that its performance is highly sensitive to the internal parameter selection, which generically includes 
the denoising strength (of the denoiser) $\sigma$ and the termination time $\tau$, and others specified by the given optimization scheme (\eg the penalty parameter $\mu$ in \eqref{eq:pnp-admm-2}). 
  The body of literature often utilizes  manual tweaking, \eg~\citep{Chang_2017_ICCV,Meinhardt_2017_ICCV}  or handcrafted criteria, \eg~\citep{chan2017plug,Zhang_2017_CVPR,eksioglu2016decoupled,tirer2018image} to select parameters for each specific problem setting. 
However, manual parameter tweaking requires several trials, which can be very cumbersome and time-consuming. Semi-automated handcrafted criteria (\eg monotonically decreasing the denoising strength) can, to some degree, ease the burden of exhaustive searches of large parameter spaces, but often lead to  sub-optimal local minima.  
Moreover, the optimal parameter setting differs from image to image, depending on the measurement model, the noise level, the noise type and the unknown image itself. These differences are illustrated
in  Figure~\ref{fig:example}, where peak signal-to-noise ratio (PSNR) curves are displayed for eight images on two applications (compressed sensing MRI and CT reconstruction) under varying denoising strengths.

\subsection{Contributions}
This work addresses the challenge of how to deal with the manual parameter tuning problem in a PnP framework.  To this end, we formulate the internal parameter selection as a sequential decision-making problem, where a policy is adopted to select a sequence of internal parameters to guide the optimization.
This formulation can be naturally fit into a reinforcement learning (RL) framework, where a policy agent seeks to map observations to actions,  with the aim of maximizing cumulative reward. The reward reflects the \textit{to do} or \textit{not to do} events for the agent, and a desirable high reward can be obtained if the policy leads to faster convergence and better restoration accuracy.

\begin{figure}[!t]
\centering
\includegraphics[width=1\linewidth,clip,keepaspectratio]{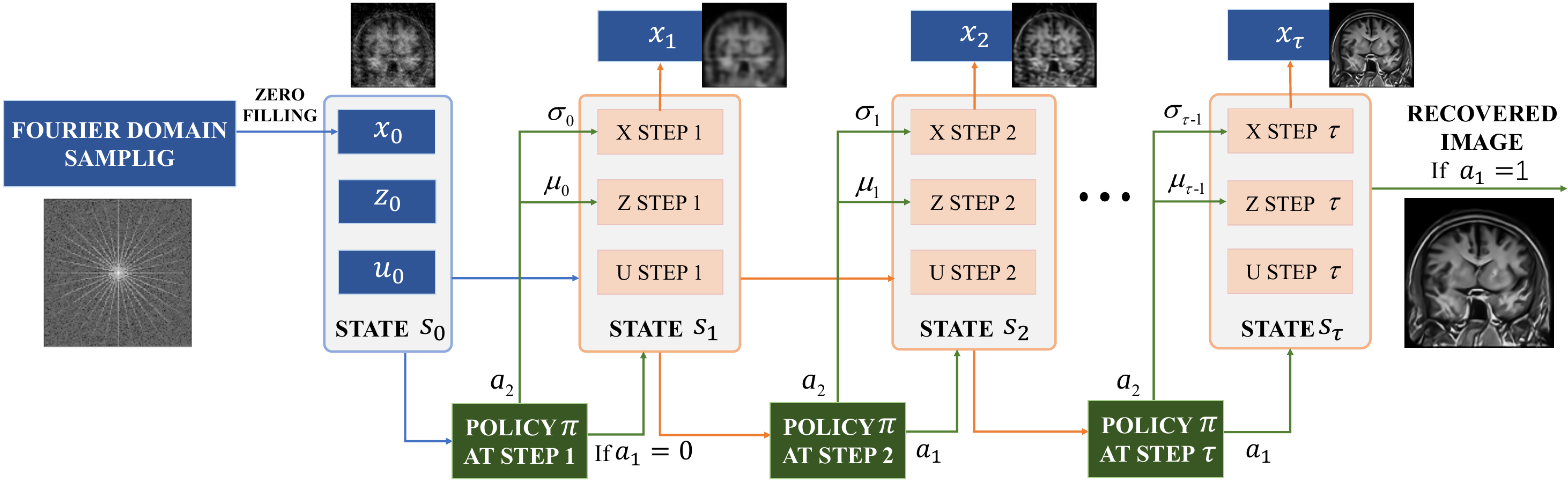}
\caption{Workflow of our proposed tuning-free plug-and-play framework instantiated by the PnP-ADMM algorithm, in which we formulate the internal parameter selection as a sequential decision-making problem. The visualization displays the example case for fast MRI reconstruction. 
}
\label{fig:framework}
\end{figure}

We demonstrate, through rigorous numerical and visual experiments, the advantage of our algorithmic approach on compressed sensing MRI, sparse-view CT, single-photon imaging, and phase retrieval problems.  We show that the policy allows to well approximate the intrinsic function that maps the input state to its optimal parameter setting. 
By using the learned policy, the guided optimization can reach comparable results to the ones using oracle parameters tuned via the inaccessible ground truth. 
An overview of our algorithm instantiated by PnP-ADMM is exhibited in Figure~\ref{fig:framework}. Though the majority of results in this paper are given by PnP-ADMM, we also show our generic approach can be extended to a wide range of PnP-type algorithms, including but not limited to the plug-and-play proximal gradient method (PGM) \citep{sun2019online}, half-quadratic splitting (HQS) \citep{Zhang_2017_CVPR}, regularization by denoising (RED) \citep{romano2017little} as well as denoising-based approximate message passing (D-AMP) \citep{metzler2016denoising}.
All the experiments collectively suggest the effectiveness of our algorithm in terms of both convergence speed and reconstruction accuracy. 

The main contributions in this work are as follows: 
\begin{enumerate}
\item We present a tuning-free plug-and-play (TFPnP) framework that can customize parameters towards a diverse range of imaging problems, and achieve  faster practical convergence and better empirical performance than handcrafted criteria. 
\item We introduce an efficient mixed model-free and model-based RL algorithm. It can optimize jointly the discrete termination time, and the continuous parameters \eg denoising strength, penalty parameters, and step size, etc.
\item We validate our approach with an extensive range of numerical and visual experiments and show our generic method is applicable to a wide spectrum of PnP-type algorithms. We also show that our well-designed approach leads to better results than state-of-the-art techniques on compressed sensing MRI, sparse-view CT, single-photon imaging, and phase retrieval.\footnote{Our code and pretrained models are made publicly available at https://github.com/Vandermode/TFPnP}
\end{enumerate}

The remainder of this paper is organized as follows.
In Section \ref{sec:related-work}, we provide an overview of PnP algorithmic techniques and other related philosophies. Section \ref{sec:TFPnP} introduces our TFPnP approach for automated parameter selection in a PnP framework.  Main experimental results on four inverse imaging problems are presented in Section \ref{sec:main-exp}, followed by further investigation of our TFPnP algorithm and its extensions to several PnP-type methods in Section \ref{sec:investigate}
and \ref{sec:extension}, respectively. Conclusion and future work are drawn in Section \ref{sec:conclusion}. A preliminary version of this work was presented as a conference paper in~\citep{wei2020tuning}.

\section{Related Work} \label{sec:related-work}
In this section, we first provide a brief overview of several important developments of PnP  techniques. Then we review related works of other categories, which include automated parameter selection, algorithm unrolling as well as reinforcement learning for image recovery.  

\medskip

\paragraph{Plug-and-play (PnP) Techniques}
The definitional concept of PnP was first introduced in  \citep{danielyan2010image,zoran2011learning,venkatakrishnan2013plug}. It has attracted great  attention owing to its effectiveness and flexibility to handle a wide range of inverse imaging problems.  Following this philosophy, the body of literature can be
roughly categorized in terms of six aspects including proximal algorithms, denoiser priors, imaging applications, theoretical (convergence) analysis, and other two PnP-type frameworks (regularization by denoising (RED) \citep{romano2017little} and denoising-based approximate message passing (D-AMP) \citep{metzler2016denoising}).
Note that the PnP with learning-based denoisers also belongs to a broader class of methods coined as
 \textit{learning to optimize}, whose overview is provided in~\citep{chen2021learning}. 

\begin{itemize}
\item \textbf{Proximal algorithms.~}
Initially, PnP was proposed to be combined with ADMM algorithm \citep{venkatakrishnan2013plug}. 
Later on, extensions to other first-order proximal splitting algorithms are studied
including  half-quadratic splitting \citep{Zhang_2017_CVPR}, primal-dual method \citep{ono2017primal}, accelerated proximal gradient method \citep{kamilov2017plug} as well as their stochastic variants \citep{sun2019online,sun2020scalable}.

\item \textbf{Denoising priors.~} A key building block of PnP framework is the denoising algorithm. Over the years 
a wide range of denoisers have been used for the PnP framework, 
such as BM3D \citep{dar2016postprocessing,rond2016poisson,sreehari2016plug,chan2017plug}, 
sparse representation \citep{brifman2016turning},  patch-based Wiener filtering \citep{venkatakrishnan2013plug}, 
nonlocal means \citep{venkatakrishnan2013plug,sreehari2016plug}, 
Gaussian mixture models \citep{teodoro2016image,teodoro2018convergent}, 
weighted nuclear norm minimization \citep{kamilov2017plug}, 
and deep learning-based denoisers \citep{Meinhardt_2017_ICCV,Zhang_2017_CVPR,Chang_2017_ICCV,zhang2020plug}.

\item \textbf{Imaging applications.~}  PnP methods reported empirical success on a large variety of imaging applications such as  bright field electronic tomography \citep{sreehari2016plug}, diffraction tomography \citep{sun2019online,wu2020simba},  nonlinear inverse scattering \citep{kamilov2017plug}, 
low-dose CT imaging \citep{he2018optimizing}, compressed sensing MRI \citep{eksioglu2016decoupled},  electron microscopy \citep{sreehari2017multi}, 
single-photon imaging \citep{chan2017plug}, phase retrieval \citep{metzler2018prdeep}, Fourier ptychography microscopy \citep{sun2019regularized}, light-field photography \citep{chun2019momentum}, 
hyperspectral sharpening \citep{teodoro2018convergent},  synthetic aperture radar image reconstruction \citep{alver2020plug}, 
snapshot compressive imaging \citep{liu2018rank,yuan2020plug}, image/video processing---\eg denoising, demosaicking, deblurring, super-resolution and inpainting \citep{heide2014flexisp,rond2016poisson,Meinhardt_2017_ICCV,tirer2018image,brifman2019unified,zhang2019deep,zhang2020plug}.  

\item  \textbf{Theoretical analysis.~} 
The empirical success of PnP spurs researchers to study its theoretical properties (\eg convergence), which is challenging as PnP priors are implicit and in general have no analytic mathematical expressions. Efforts along this direction were also made over recent years, to establish  convergence via the symmetric gradient \citep{sreehari2016plug}, the proper denoiser \citep{metzler2016denoising}, the bounded denoiser \citep{chan2017plug} or the linear (possibly non-symmetric) denoiser \citep{gavaskar2021plug,nair2021fixed},
 and/or the nonexpansiveness assumptions \citep{sreehari2016plug,teodoro2018convergent,sun2019online,chan2019performance}.
Remarkably, the authors of~\citep{ryu2019plug} show that PnP methods are contractive iterations
under a mild Lipschitz condition on the denoisers. They further propose a technique namely real spectral normalization 
to enforce the proposed Lipschitz condition on training deep learning-based denoisers. 
Additionally, the works of \citep{buzzard2018plug,chan2019performance} provide a consensus equilibrium view to PnP frameworks, 
shedding light on building connections among different PnP-type algorithms.

\item  \textbf{Regularization by denoising.~} Another framework closely related to PnP is so-called regularization by denoising (RED) \citep{romano2017little}, which uses the denoiser for defining the prior/regularizer explicitly. Under some conditions, the gradient of  the regularizer has a very simple expression, which can be readily utilized in first-order optimization solvers, \eg gradient descent and ADMM.  For general cases where the conditions do not hold, 
the RED algorithms can be explained via a score-matching framework or consensus equilibrium \citep{reehorst2018regularization}, 
or reformulated as a convex optimization problem via fixed-point projection \citep{cohen2020regularization}. 
A provably convergent block-coordinate RED is introduced in \citep{NEURIPS2019_9872ed9f} to tackle large-scale image recovery problems by a sequence of random partial updates. 
It is further extended into the asynchronous version \citep{sun2021async} for parallel and distributed implementations. 

\item  \textbf{Denoising-based approximate message passing.~}
Different from the original PnP based on certain proximal splitting algorithms, denoising-based approximate message passing (D-AMP) \citep{metzler2016denoising}, as another PnP-type framework, derives from the AMP algorithm that is built upon the loopy belief propagation in graphical models \citep{donoho2009message,donoho2010messageI,donoho2010messageII,rangan2010estimation,rangan2011generalized}. 
The main appealing feature of the AMP algorithm is that for
compressed sensing with i.i.d. sub-Gaussian measurement matrices, the method exhibits fast convergence with precise analytic characterizations on the algorithm behavior known as state evolution \citep{bayati2011dynamics}, which leads to near optimal tuning of internal parameters using Stein’s unbiased risk estimator (SURE) \citep{guo2015near}.  
 The AMP, however, is fragile in that even small deviations from the i.i.d. sub-Gaussian model can cause the algorithm to diverge. Although many extensions have been proposed to work with a far broader class of matrices \citep{cakmak2014s,manoel2015swept,ma2017orthogonal,rangan2019vector,millard2020approximate,metzler2021d}, the algorithm's behavior is still largely intractable under general measurement matrices \citep{rangan2016fixed,rangan2019convergence}, which seriously limits its application scope in practice.

\end{itemize}

\noindent
Differing from these aspects, in this work we focus on the challenge of parameter selection in PnP, where a bad choice of parameters often leads to severe degradation of the results \citep{romano2017little,chan2017plug}.
Unlike existing semi-automated parameter tuning criteria \citep{wang2017parameter,chan2017plug,Zhang_2017_CVPR,eksioglu2016decoupled,tirer2018image}, our method is fully automatic and is purely learned from the data, which significantly eases the burden of manual parameter tuning.

\paragraph{Automated Parameter Selection}
Choosing appropriate regularization parameters (or automated parameter selection in general) is a long-standing yet unresolved problem raised in a variety of domains such as signal/image processing, machine learning, and statistics communities. 
The existing methods for automatic parameter setting in literature can be broadly classified into four categories: 1) based on 
the discrepancy principle \citep{morozov1966solution,karl2005regularization}, 2) the L-curve method \citep{hansen1992analysis,hansen1993use,vogel1996non,reginska1996regularization,hansen1999curve}, 3) the generalized cross validation (GCV) \citep{golub1979generalized,wahba1990spline,reeves1994optimal,o1985cross,ramani2012regularization}, and 4) the Stein’s unbiased risk estimator (SURE) \citep{stein1981estimation,donoho1995adapting,blu2007sure,raphan2008optimal,giryes2008automatic,vonesch2008recursive,eldar2008generalized,ramani2008monte,giryes2011projected,ramani2012regularization,mousavi2013parameterless,bayati2013estimating,guo2015near,mousavi2018consistent}. 
The effectiveness of the former three approaches is often limited to linear reconstruction algorithms using \eg Tikhonov regularization \citep{hansen1993use,golub1979generalized}.  State-of-the-arts are therefore  attributed to SURE-based methods, which provide an unbiased estimate of mean-square error (MSE) for parameter tuning, considering certain analytical regularizers, \eg smoothed versions of the $\ell_p$ norm \citep{eldar2008generalized,giryes2011projected}, or general convex functions \citep{ramani2012regularization}.
Given more sophisticated non-convex and learned priors, the SURE is often restricted to AMP algorithms for compressed sensing applications due to the Gaussian noise assumption \citep{metzler2016denoising}. Our method, instead, is generally applicable to almost every PnP-type algorithm for a wide range of inverse imaging problems.

\paragraph{Algorithm Unrolling}
Perhaps the most confusable concept to  PnP in the deep learning era is  so-called algorithm unrolling methods (a.k.a. deep unfolding or unrolled optimization) \citep{gregor2010learning,hershey2014deep,wang2016proximal,NIPS2016_6406,Zhang_2018_CVPR,diamond2017unrolled,NIPS2017_6774,lefkimmiatis2018universal,yang2018proximal,adler2018learned,mardani2018neural,aggarwal2018modl,dong2018denoising,xie2019differentiable,gilton2019neumann,liu2019knowledge,zhang2020deep,aggarwal2020j,li2020end,lecouat2020fully,dong2021model,huang2021deep,liu2021stochastic}, which explicitly unroll/truncate iterative optimization algorithms into learnable deep architectures. 
In this way, the denoiser prior and other internal parameters are treated as trainable parameters, meanwhile, the number of iterations has to be fixed to enable end-to-end training.  By contrast, our TFPnP approach can adaptively select a stopping time and other internal parameters given varying input states. 
We refer interested readers to \citep{monga2021algorithm} for a thorough review of this topic. 

\paragraph{Reinforcement Learning for Image Recovery}
Although Reinforcement Learning (RL) has been applied in a range of domains, from game playing \citep{mnih2013playing,silver2016mastering} to robotic control \citep{schulman2015trust}, only few works have successfully employed RL to the image recovery tasks. The study in~\citep{yu2018crafting} learned an RL policy to select appropriate tools from a toolbox to progressively restore corrupted images. 
The work of~\citep{zhang2018dynamically} proposed a recurrent image restorer whose endpoint was dynamically controlled by a learned policy. In~\citep{furuta2019fully}, authors used RL to select a sequence of classic filters to process images gradually. 
The work of~\citep{yu2019path}  learned network path selection for image restoration in a multi-path CNN.  In contrast to these works, we apply a mixed model-free and model-based deep RL approach to automatically select the parameters for the PnP image recovery algorithm.

\section{Tuning-free Plug-and-Play Proximal Algorithm} \label{sec:TFPnP}
In this section, we elaborate on our tuning-free PnP proximal algorithm  described in ~\eqref{eq:pnp-admm-1}-\eqref{eq:pnp-admm-3}, which contains three main parts. Firstly, we describe how the automated parameter selection is driven. Secondly, we introduce our environment model. Finally, we introduce policy learning, which is guided by a mixed model-free and a model-based RL. 
We remark that, though our presentation on ~\eqref{eq:pnp-admm-1}-\eqref{eq:pnp-admm-3} is for ADMM, our methodology can be adapted to other proximal algorithms, see Section~\ref{sec:extension} for more examples. 


\subsection{RL Formulation for Automated Parameter Selection} 
Our aim is to automatically  select a sequence of parameters $(\sigma_0, \mu_0, \sigma_1, \mu_1 ,\cdots,\sigma_{\tau-1}, \mu_{\tau-1}$) to guide the optimization~\eqref{eq:pnp-admm-1}-\eqref{eq:pnp-admm-3} such that the recovered image $x^{\tau}$ is close to the underlying image $x$. 
We formulate this problem as a Markov decision process (MDP), which can be addressed via reinforcement learning (RL). 

We denote the MDP by the tuple $(\mathcal{S}, \mathcal{A}, p, r)$, where $\mathcal{S}$ is the state space, $\mathcal{A}$ is the action space,  $p$ is the transition function describing the environment dynamics, and $r$ is the reward function. 
Specifically, for our task, $\mathcal{S}$ is the space of optimization variable states, which includes the initialization  $(x_{0}, z_{0}, u_{0})$ and all intermediate results $(x_k, z_k, u_k)$ in the optimization process. $\mathcal{A}$ is the space of internal parameters, including both discrete termination time $\tau$ and the continuous denoising strength/penalty parameters ($\sigma_k$, $\mu_k$). The transition function $p: \mathcal{S} \times \mathcal{A} \to \mathcal{S}$ maps input state $s \in \mathcal{S}$ to its outcome state $s' \in \mathcal{S}$ after taking action $a \in \mathcal{A}$. The state transition can be expressed as $s_{t+1} = p(s_t, a_t)$, which is composed of one or several iterations in the optimization.  On each transition, the environment emits a reward in terms of the reward function $r: S \times \mathcal{A} \to \mathbb{R}$, which evaluates actions given the state. 
Applying a sequence of parameters to the initial state $s_0$ results in a trajectory $T$ of states, actions and rewards, \ie $T = \{s_0, a_0, r_0, \cdots, s_{N}, a_{N}, r_{N}\} $. 
Given a trajectory $T$, we define the return $R_{t}$ 
as the summation of discounted rewards after $s_t$, which reads:
\begin{align}
R_{t} = \sum_{t'=0}^{N-t} \rho^{t'} r (s_{t+t'}, a_{t+t'}),
\end{align}
where $\rho \in [0, 1]$ is a discount factor and prioritizes earlier rewards over later ones. 

Our goal is to learn a policy $\pi$, denoted as $\pi(a|s): \mathcal{S} \to \mathcal{A}$ for the decision-making agent, in order to maximize
the objective defined as 
 \begin{align}
J\left(\pi\right) = \mathbb{E}_{s_0 \sim S_0,  T \sim \pi} \left[ R_{0} \right],
\end{align}
where $\mathbb{E}$ represents expectation, $s_0$ is the initial state, and $S_0$ is the corresponding initial state distribution. Intuitively, the objective describes the expected return over all possible trajectories induced by the policy $\pi$. 
The expected return on states and state-action pairs under the policy $\pi$ are defined by state-value functions $V^{\pi}$ and action-value functions $Q^{\pi}$ respectively, \ie
\begin{align}
V^{\pi}\left(s\right)    &=  \mathbb{E}_{T\sim \pi} \left[ R_{0} | s_0 = s \right], \\
Q^{\pi}\left(s, a\right) &= \mathbb{E}_{T\sim \pi} \left[ R_{0} | s_0 = s, a_0 = a \right].
\label{eq: Q-function}
\end{align}
In our task, we decompose actions into two parts, \ie $a = (a_1, a_2)$\footnote{Strictly speaking, $a_t = (a_{t1}, a_{t2})$. Here we omit the notation $t$ (time step) for simplicity. The readers should not be confused as the meaning can be easily inferred from the context.  }, which includes a discrete decision $a_1$ on termination time $\tau$ and a continuous decision $a_2$ on denoising strength $\sigma$ and penalty parameter $\mu$. The policy also consists of two sub-policies: $\pi = (\pi_1, \pi_2)$, a stochastic policy and a deterministic policy that generate $a_1$ and $a_2$ respectively.  
The role of $\pi_1$ is to decide whether to terminate the iterative algorithm when the next state is reached. 
It samples a boolean-valued outcome $a_1$ from a two-class categorical distribution $\pi_1(\cdot | s)$, whose probability mass function is calculated from the current state $s$. 
We move forward to the next iteration if $a_1 = 0$, otherwise, the optimization would be terminated to output the final state.  Compared to the stochastic policy $\pi_1$, we treat $\pi_2$ deterministically, \ie $a_2 = \pi_2(s)$. Since $\pi_2$ is differentiable with respect to the environment,  its gradient can be precisely estimated. %

\subsection{Environment Model}\label{sec: env-denoiser}

In RL the environment is characterized by two components, \ie the environment dynamics and  reward function. In our task, the environment dynamics is described by the transition function $p$ related to the PnP-ADMM. Here, we detail the setting of the PnP-ADMM as well as the reward function used for training policy. 


\paragraph{Denoiser Prior}
A differentiable environment makes policy learning more efficient. 
To make the environment differentiable with respect to $\pi_2$\footnote{$\pi_1$ is non-differentiable towards environment regardless of the formulation of the environment.}, 
we take a convolutional neural network (CNN) denoiser as the image prior. 
In practice, we use a residual U-Net~\citep{ronneberger2015u}  architecture, which was originally designed for medical image segmentation and was recently found to be useful in image denoising~\citep{brooks2019unprocessing}.  Besides, 
we incorporate an additional tunable noise level map into the input as  \citep{zhang2018ffdnet}, enabling us to provide continuous noise level control (\ie different denoising strength) within a single network. 

\paragraph{Proximal Operator of Data-fidelity Term}
Enforcing consistency with measured data requires evaluating the proximal operator of $\mathcal{D}$ in~\eqref{eq:pnp-admm-2}. For  inverse problems, there might exist fast solutions due to the special structure of the observation model. We adopt the fast solution if it is available (\eg closed-form solution using fast Fourier transform, rather than the general matrix inversion), 
otherwise a single step of gradient descent is performed.

\paragraph{Transition Function}
To reduce the computation cost, we define the transition function $p$ to involve $m$ iterations of the optimization. 
At each time step, the agent  needs to decide the internal parameters for $m$ iterates.  
Larger $m$ would lead to coarser control of the termination time, but make the decision-making process more efficient. 
Its value is set to 5 empirically. 
To avoid the optimization loops infinitely, 
 the maximum time step $N$ is set to 6 in our algorithm, leading to 30 iterations of the optimization at most. 

\paragraph{Reward Function}
To take both image recovery performance and runtime efficiency into account,
we define the reward function as: 
\begin{align}
r(s_t, a_t) = \left[ \zeta(p(s_t, a_t)) - \zeta(s_t) \right]  - \eta, 
\end{align}
The first term, $\zeta(p(s_t, a_t)) - \zeta(s_t)$,  
denotes the PSNR increment made by the policy, where $\zeta(s_t)$ denotes the PSNR of the recovered image at step $t$.  
A higher reward is acquired if the policy leads to higher performance gain in terms of PSNR. 
The second term, $\eta$, implies penalizing the policy as it does not select to terminate at step $t$, where $\eta$ sets the degree of penalty.
A negative reward is given if the PSNR gain does not exceed the degree of penalty, thereby encouraging the policy to early stop the iteration with diminishing returns.  We set $\eta=0.05$ in our algorithm\footnote{The choice of the hyperparameters  $m,N$ and $\eta$ is discussed in Section \ref{sec: hyperparameter}.}.

\begin{table}[!t]
	\centering
	\setlength{\tabcolsep}{1mm}{
	\renewcommand{\arraystretch}{1.5}
	\begin{tabular}{c|c|c}
	 \textsc{Layer name}   & \textsc{Feature extractor} & \textsc{Output size} \\ \hline
	 conv1  & $3\times3, 64$   & $64\times64$ \\ \hline
	 layer1   & $\left[ \substack{3\times3,\, 64\\3\times3, \, 64} \right] \times 2$ & $32\times32$ \\ \hline
	 layer2   & $\left[ \substack{3\times3,\, 128\\3\times3,\, 128} \right] \times 2$ & $16\times16$ \\ \hline
	 layer3   & $\left[ \substack{3\times3,\, 256\\3\times3,\, 256} \right] \times 2$ & $8\times8$\\ \hline
	 layer4   & $\left[ \substack{3\times3,\, 512\\3\times3,\, 512} \right] \times 2$ & $4\times4$\\ \hline
 	avgpool & $4\times4$ average pooling & $1\times1$ \\ \hline
	\end{tabular}}
	\caption{Network configuration of the feature extractor of our policy and value networks. 
	We refer to  `$3\times3,\, 64$' as  $3\times 3$ kernel size and 64 output feature maps.
	Each layer is composed of two building blocks of the residual network.
	The building block consists of two convolutional layers with batch normalization (BN) \citep{ioffe2015batch}, ReLU activation, and skip connection.  The BN and ReLU are replaced by weight normalization \citep{salimans2016weight} and TReLU \citep{xiang2017trelu} respectively in the value network. 
	}	
	\label{tb:network}
\end{table}

\subsection{RL-based Policy Learning}
In this section, we present a mixed model-free and model-based RL algorithm to learn the policy.
Specifically, model-free RL (agnostic to the environment dynamics) is used to train $\pi_1$, while model-based RL is utilized to optimize $\pi_2$ to make full use of the environment model\footnote{$\pi_2$ can also be optimized in a model-free manner. The comparison can be found in the Section \ref{sec:csmri}.}. 
We employ the actor-critic framework \citep{sutton2000policy}, 
which uses a policy network $\pi_\theta(a_t| s_t)$ (actor) and a value network $V_\phi^{\pi}(s_t)$ (critic) to formulate the policy and the state-value function respectively.  

The design principle of a policy/value network is to make it simple yet effective. For convenience, we follow \citep{Huang_2019_ICCV} that uses residual structures similar to ResNet-18 \citep{He_2016_CVPR} as the feature extractor in the policy and value networks, followed by fully-connected layers and activation functions to produce desired outputs. The network configuration of the feature extractor of the policy and value networks is listed in Table \ref{tb:network}.
To make the policy aware of the problem settings \citep{silver2016mastering}, we encode the auxiliary setting-related  information (\ie the measurement noise level and the number of steps that have been taken so far in the optimization process) into the input of the policy network. Notice the input of our policy network is the optimization variables ($x_k, z_k, u_k$) with spatial size $H \times W$.  To encode the extra scalar-type information into the policy network, we augment the input with the same spatial-sized feature maps with all entries equal to the same value representing the auxiliary information. These values are normalized into [0, 1] range to avoid inconsistent magnitude.
It is worth noting that the extra computation cost of the policy network is marginal, compared with the iteration cost in the PnP-ADMM~\eqref{eq:pnp-admm-1}-\eqref{eq:pnp-admm-3}.


The policy and the value networks are learned in an interleaved manner.  
For each gradient step, we optimize the value network parameters $\phi$ by minimizing
\begin{align}
    L_{\phi} = \mathbb{E}_{s \sim B, a \sim \pi_{\theta}(s)} \left[ \frac{1}{2} (r(s, a) + \gamma V^{\pi}_{\hat{\phi}}(p(s, a)) - V^{\pi}_{\phi}(s) )^2 \right],
\end{align}
where $B$ is the distribution of previously sampled states, practically implemented by a state buffer. This partly serves as a role of the experience replay mechanism \citep{lin1992self-improving}, which is observed to ``smooth" the training data distribution \citep{mnih2013playing}. The update makes use of a target value network $V^{\pi}_{\hat{\phi}}$, where $\hat{\phi}$ is the exponentially moving average of the value network weights and has been shown to stabilize training \citep{mnih2015human-level}. 

The policy network has two sub-policies, which employs shared convolutional layers to extract image features, followed by two separated groups of fully-connected layers to produce termination probability $\pi_1(\cdot|s)$ (after softmax) or denoising strength/penalty parameters $\pi_2(s)$ (after sigmoid). We denote the parameters of the sub-policies as $\theta_1$ and $\theta_2$ respectively, and we seek to optimize $\theta=(\theta_1, \theta_2)$ so that the objective $J(\pi_\theta)$ is maximized. 
The policy network is trained using policy gradient methods \citep{peters2006policy}. 
The gradient of $\theta_1$ is estimated by a likelihood estimator in a model-free manner, while the gradient of $\theta_2$ is estimated relying on backpropagation via environment dynamics in a model-based manner. 
Specifically, for discrete termination time decision $\pi_1$, 
we apply the policy gradient theorem \citep{sutton2000policy} to obtain unbiased Monte Carlo estimate of $\triangledown_{\theta_1} J(\pi_\theta)$ using the advantage function $A^\pi(s, a) = Q^\pi(s, a) - V^\pi(s)$ as target, which is computed as:
\begin{align}
\triangledown_{\theta_1} J(\pi_\theta)= &\mathbb{E}_{s \sim B, a \sim \pi_{\theta}(s)}
\left[\triangledown_{\theta_1} \mathrm{log}\,\pi_1(a_1|s)\,A^{\pi}(s, a)\right]. 
\label{eq:pi_1}
\end{align}
  For continuous denoising strength and penalty parameter selection $\pi_2$,
we utilize the deterministic policy gradient theorem \citep{silver2014deterministic} to formulate its gradient which reads:
\begin{align}
\triangledown_{\theta_2} J(\pi_\theta)= &\mathbb{E}_{s \sim B, a \sim \pi_{\theta}(s)}
\left[\triangledown_{a_2} \,Q^{\pi}(s, a)\triangledown_{\theta_2} \pi_2 (s) \right],
\label{eq:pi_2}
\end{align}
where we approximate the action-value function $Q^{\pi}(s, a)$ by $r(s, a) + \gamma V_\phi^{\pi}(p(s, a))$ given its unfolded definition \citep{sutton2018reinforcement}.

Using the chain rule, we can directly obtain the gradient of $\theta_2$ by backpropagation via the reward function, the value network and the transition function, in contrast to relying on the gradient backpropagated from only the learned action-value function in the model-free DDPG algorithm \citep{lillicrap2015continuous}. 

The detailed training algorithm for our policy learning is summarized in Algorithm \ref{alg:training}. It requires an image dataset $D$, a degradation operator $g(\cdot)$, learning rates $l_{\theta}$, $l_{\phi}$, and a weight parameter $\beta$. 
To sample initial states $s_0$, we define the degradation operator $g(\cdot)$ as a composition of a forward model and an initialization function.
The forward model maps the underlying image $x$ to its observation $y$, while the initialization function generates the initial estimate $x_0$ from the observation $y$. 
For linear inverse problems, $g(\cdot)$ is typically defined by the composition of the forward operator and the adjoint operator of the problem (for example, $g(\cdot)$ is a composition of the partially-sampled Fourier transform and the inverse Fourier transform in CS-MRI).
For nonlinear inverse problem---phase retrieval, we employ the HIO algorithm \citep{Fienup1982Phase} as the initialization function.

\begin{algorithm}[!t]
  \caption{Training Scheme} 
  \label{alg:training}
  \small
  \begin{algorithmic}[1] 
  \REQUIRE 
  Image dataset $D$, degradation operator $g(\cdot)$,  learning rates $l_{\theta}$, $l_{\phi}$, weight parameter $\beta$. 
  \STATE Initialize network parameters $\theta$, $\phi$, $\hat{\phi}$ and state buffer $B$. 
   \FOR{each training iteration}
    \STATE sample initial state $s_0$ from $D$ via $g(\cdot)$
     \FOR{environment step $t \in [0, N)$}
  	    \STATE $a_t \sim \pi_{\theta} (a_t | s_t) $
  	    \STATE $s_{t+1} \sim p(s_{t+1} | s_t, a_t) $
  	    \STATE $B \leftarrow B \cup \{s_{t+1}\}$
  	    \STATE break if the boolean outcome of $a_t$ equals to 1 
     \ENDFOR
     \FOR{each gradient step}
        \STATE sample states from the state buffer $B$
        \STATE $\theta_1 \leftarrow \theta_1 + l_{\theta} \triangledown_{\theta_1} J(\pi_\theta)  $
        \STATE $\theta_2 \leftarrow \theta_2 + l_{\theta} \triangledown_{\theta_2} J(\pi_\theta)  $
        \STATE $\phi \leftarrow \phi - l_{\phi} \triangledown_\phi L_{\phi} $
        \STATE $\hat{\phi} \leftarrow \beta \phi + (1-\beta) \hat{\phi}$
     \ENDFOR
    \ENDFOR
      \ENSURE
      Learned policy network $\pi_{\theta}$
  \end{algorithmic}
\end{algorithm}

\section{Experimental Results} \label{sec:main-exp}
 
In this section, we first detail the implementation of our TFPnP algorithm, then present  experimental results on three inverse imaging applications. 
Both linear (\ie  Compressed Sensing MRI and Sparse-view CT) and nonlinear (\ie Phase Retrieval) inverse imaging problems are explored to demonstrate the robustness and  applicability of our algorithm.

\subsection{Implementation Details}
Our TFPnP algorithm requires two training stages sequentially, 
\ie  the denoiser (denoising network) training stage and the policy learning stage. 
To train the denoising network, we follow the common practice that uses 87,000 overlapping patches (with size $128\times128$) drawn from 400 images from the BSD dataset \citep{MartinFTM01}. For each patch, we add white Gaussian noise with noise level sampled from $[1, 50]$. 
The denoising networks are trained with 50 epochs using $L_1$ loss and Adam optimizer \citep{kingma2014adam} with batch size 32. The base learning rate is set to $10^{-4}$ and halved at epoch 30, then reduced to $10^{-5}$ at epoch 40.

To train the policy network (and an auxiliary value network only used  during training), we use the 17,125 resized images with size $128\times128$ from the PASCAL VOC dataset \citep{everingham2014pascal}. 
We use Adam optimizer with batch size 48 and 2500 training iterations. 
We start by setting the learning rates $l_{\theta}$, $l_{\phi}$ for updating the policy network $\pi_\theta$ and the value network $V_\phi^\pi$ to $1\times10^{-4}$ and $5\times 10^{-5}$ respectively. Then we reduced these values to  $5\times 10^{-5}$ and $1\times10^{-5}$ respectively at training iteration 1600. The value network learning makes use of a target value network, which is a soft copy of the value network itself. The weight parameter $\beta$ for softly updating the target value network is set to $10^{-3}$ (see line 15 in Algorithm \ref{alg:training}). 
In each training iteration, we alternate between collecting states (in a state buffer) from the environment with the current policy and updating the network parameters using policy gradients from batches sampled from the state buffer $B$. Ten gradient steps are performed at each training iteration. 

For each inverse imaging application, only a single policy network is trained to tackle different imaging settings, for instance, 
 one policy network is learned to handle multiple sampling ratios (with 2$\times$/4$\times$/8$\times$ acceleration) and noise levels (5/10/15) simultaneously for the  CS-MRI application. 
  
\begin{table}[t]
    \centering
\begin{tabular}{C{.2\linewidth}|C{.15\linewidth}|C{.15\linewidth}|C{.15\linewidth}|}
\hline
\multicolumn{1}{|c|}{\textsc{Method}}                 & \cellcolor[HTML]{EFEFEF}DnCNN & \cellcolor[HTML]{EFEFEF}MemNet & \cellcolor[HTML]{EFEFEF}U-net \\ \hline
\multicolumn{1}{|l|}{Denoising Performance} & 27.18                         & 27.32                          & 27.40                       \\ \hline
\multicolumn{1}{|l|}{PnP Performance}       & 25.43                         & 25.67                          & 25.76                 \\ \hline
\multicolumn{1}{|l|}{Run Times (ms)}           & 8.09                          & 64.65                          & 5.65                \\ \hline
\end{tabular}
    \caption{Numerical comparison of different CNN-based denoisers. We show the results of  Gaussian denoising performance (PSNR) under noise level $\sigma=50$,  the CS-MRI performance (PSNR) when plugged into the PnP-ADMM, and  the GPU runtime (millisecond) of denoisers when processing an image with size $256\times256$.}
    \label{tb:denoiser-profile}
\end{table}

\begin{table}[t]
\centering
{
\begin{tabular}{lcccccc} 
		\toprule
		 & \multicolumn{2}{c}{$2\times$} & \multicolumn{2}{c}{$4\times$} & \multicolumn{2}{c}{$8\times$} \\ \cline{2-7} 
		\textsc{Policies} & PSNR & \#IT.  & PSNR & \#IT.  & PSNR & \#IT. \\  \midrule
		handcrafted & 30.05 & 30.0 & 27.90 & 30.0 & 25.76 & 30.0\\
		handcrafted$^{*}$ & 30.06 & 29.1 & 28.20 & 18.4 & 26.06 & 19.4\\
		fixed  & 23.94 & 30.0 & 24.26 & 30.0 & 22.78 & 30.0 \\
		fixed$^{*}$  & 28.45 & 1.6 & 26.67 & 3.4 & 24.19 & 7.3 \\
		fixed optimal & 30.02 & 30.0 & 28.27 & 30.0 & 26.08 & 30.0  \\
		fixed optimal$^{*}$ & 30.03 & 6.7 & 28.34 & 12.6 & 26.16 & 16.7 \\
		greedy & 29.51 & 30.0 & 28.39 & 30.0 & 26.20 & 30.0 \\
		greedy$^{*}$ & 30.07 & 8.0 & 28.47 & 17.6 & 26.33 & 19.0 \\ 
		oracle & 30.25 & 30.0 & 28.60 & 30.0 & 26.41 & 30.0 \\
		oracle$^{*}$ & \textcolor{blue}{30.26} & 8.6 & \color{orange}{28.61} & 13.9 & \color{blue}{26.45} & 21.6  \\ \midrule
		model-free & 28.79 & 30.0 & 27.95 & 30.0 & 26.15 & 30.0 \\
	    Ours  & \color{orange}{30.33} & 5.0 & \textcolor{blue}{28.58} & 10.0 & \color{orange}{26.52} & 15.0 \\
		\bottomrule
\end{tabular}}
\caption{Comparisons of different policies used in PnP-ADMM algorithm for CS-MRI on seven widely used medical images under various acceleration factors (2$\times$/4$\times$/8$\times$) and noise level 15. 
We show both PSNR and the number of iterations (\#IT.) used to induce the results. * denotes to report the best PSNR over all iterations (\ie with optimal early stopping). 
The best results are indicated by \textcolor{orange}{orange} color and the second best results are denoted by \textcolor{blue}{blue} color. 
}
\label{tb:policy-comparison}
\end{table}

\begin{table}[t]
	\centering
	\footnotesize
	\setlength{\tabcolsep}{0.7mm}{
	\begin{tabular}{|C{.095\linewidth}|C{.05\linewidth}|C{.05\linewidth}|C{.095\linewidth}C{.095\linewidth}C{.105\linewidth}C{.105\linewidth}C{.115\linewidth}C{.085\linewidth}C{.085\linewidth}|}
		\hline
		\multirow{2}{*}{\textsc{Dataset}} & \multirow{2}{*}{$f$} & \multirow{2}{*}{$\sigma_n$}  & \multicolumn{2}{c|}{ \cellcolor[HTML]{EFEFEF}\textsc{Traditional}} & \multicolumn{2}{c|}{ \cellcolor[HTML]{EFEFEF}\textsc{Algorithm Unrolling}} & \multicolumn{3}{c|}{ \cellcolor[HTML]{EFEFEF}\textsc{PnP}} \\ \cline{4-10}
		& & &  RecPF & FCSA & ADMMNet  & ISTANet & BM3D-MRI & IRCNN & Ours \\\hline 
\multirow{9}{*}{Medical7}  
&\multirow{3}{*}{$2\times$} &$5$ &$32.46$ &$31.70$ &$33.10$ &$34.58$ &$33.33$ &\textcolor{blue}{$34.67$} &\textcolor{orange}{$35.10$}\\
& &$10$ &$29.48$ &$28.33$ &$31.37$ &\textcolor{blue}{$31.81$} &$29.44$ &$31.80$ &\textcolor{orange}{$32.04$}\\
& &$15$ &$27.08$ &$25.52$ &$29.16$ &\textcolor{blue}{$29.99$} &$26.90$ &$29.96$ &\textcolor{orange}{$30.33$}\\\cline{2-10}
&\multirow{3}{*}{$4\times$} &$5$ &$28.67$ &$28.21$ &$30.24$ &$31.34$ &$30.33$ &\textcolor{blue}{$31.36$} &\textcolor{orange}{$31.81$}\\
& &$10$ &$26.98$ &$26.67$ &$29.20$ &\textcolor{blue}{$29.71$} &$28.30$ &$29.52$ &\textcolor{orange}{$29.86$}\\
& &$15$ &$25.58$ &$24.93$ &$27.87$ &\textcolor{blue}{$28.38$} &$26.66$ &$27.94$ &\textcolor{orange}{$28.58$}\\\cline{2-10}
&\multirow{3}{*}{$8\times$} &$5$ &$24.72$ &$24.62$ &$26.57$ &\textcolor{blue}{$27.65$} &$26.53$ &$27.32$ &\textcolor{orange}{$28.31$}\\
& &$10$ &$23.94$ &$24.04$ &$26.21$ &\textcolor{blue}{$26.90$} &$25.81$ &$26.44$ &\textcolor{orange}{$27.28$}\\
& &$15$ &$23.18$ &$23.36$ &$25.49$ &\textcolor{blue}{$26.23$} &$25.09$ &$25.53$ &\textcolor{orange}{$26.52$}\\\hline
\multirow{9}{*}{MICCAI}  
&\multirow{3}{*}{$2\times$} &$5$ &$36.39$ &$34.90$ &$36.74$ &$38.17$ &$36.00$ &\textcolor{blue}{$38.42$} &\textcolor{orange}{$38.61$}\\
& &$10$ &$31.95$ &$30.12$ &$34.20$ &$34.81$ &$31.39$ &\textcolor{blue}{$34.93$} &\textcolor{orange}{$35.06$}\\
& &$15$ &$28.91$ &$26.68$ &$31.42$ &$32.65$ &$28.46$ &\textcolor{blue}{$32.81$} &\textcolor{orange}{$33.12$}\\\cline{2-10}
&\multirow{3}{*}{$4\times$} &$5$ &$33.05$ &$32.30$ &$34.15$ &$35.46$ &$34.79$ &\textcolor{blue}{$35.80$} &\textcolor{orange}{$36.17$}\\
& &$10$ &$30.21$ &$29.56$ &$32.58$ &\textcolor{blue}{$33.13$} &$31.63$ &$32.99$ &\textcolor{orange}{$33.38$}\\
& &$15$ &$28.13$ &$26.93$ &$30.55$ &\textcolor{blue}{$31.48$} &$29.35$ &$30.98$ &\textcolor{orange}{$31.75$}\\\cline{2-10}
&\multirow{3}{*}{$8\times$} &$5$ &$28.35$ &$28.71$ &$30.36$ &$31.62$ &$31.34$ &\textcolor{blue}{$31.66$} &\textcolor{orange}{$32.69$}\\
& &$10$ &$26.86$ &$27.68$ &$29.78$ &\textcolor{blue}{$30.54$} &$29.86$ &$30.16$ &\textcolor{orange}{$30.93$}\\
& &$15$ &$25.70$ &$26.35$ &$28.83$ &\textcolor{blue}{$29.50$} &$28.53$ &$28.72$ &\textcolor{orange}{$29.75$}\\\hline
	\end{tabular}}
	\caption{Quantitative results (PSNR) of different CS-MRI methods under various acceleration factors $f$ and noise levels $\sigma_n$ on two datasets.
	The best results are indicated in \textcolor{orange}{orange} color and the second best results are denoted in \textcolor{blue}{blue} color.  
	}	
	\label{tb:CS-MRI}
\end{table}

\begin{figure}[t]
	\centering
	\setlength\tabcolsep{1.5pt}
	\footnotesize
	\begin{tabular}{cccccccc}		
	 \fontsize{9pt}{9pt}\selectfont RecPF &  \fontsize{9pt}{9pt}\selectfont FCSA & \fontsize{9pt}{9pt}\selectfont ADMMNet & \fontsize{9pt}{9pt}\selectfont ISTANet & \fontsize{9pt}{9pt}\selectfont BM3D-MRI & \fontsize{9pt}{9pt}\selectfont IRCNN  & \fontsize{9pt}{9pt}\selectfont Ours & \fontsize{8pt}{8pt}\selectfont Ground Truth  \\
	\includegraphics[width=0.116\linewidth]{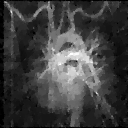}
	& \includegraphics[width=0.116\linewidth]{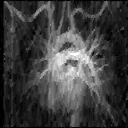}
	& \includegraphics[width=0.116\linewidth]{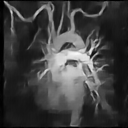}
	& \includegraphics[width=0.116\linewidth]{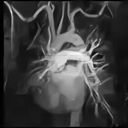}
	& \includegraphics[width=0.116\linewidth]{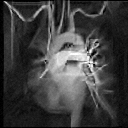}
	& \includegraphics[width=0.116\linewidth]{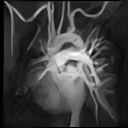}
	& \includegraphics[width=0.116\linewidth]{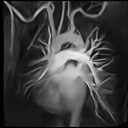}
	& \includegraphics[width=0.116\linewidth]{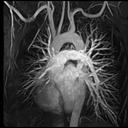} \\
	 22.57 & 22.27 & 24.15 & 24.61 & 23.64 & 24.16 & 25.28 & PSNR \\
	\includegraphics[width=0.116\linewidth]{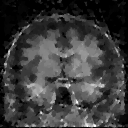}
	& \includegraphics[width=0.116\linewidth]{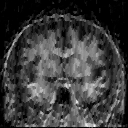}
	& \includegraphics[width=0.116\linewidth]{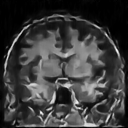}
	& \includegraphics[width=0.116\linewidth]{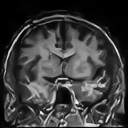}
	& \includegraphics[width=0.116\linewidth]{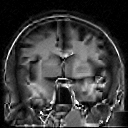}
	& \includegraphics[width=0.116\linewidth]{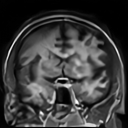}
	& \includegraphics[width=0.116\linewidth]{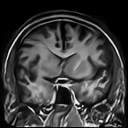}
	& \includegraphics[width=0.116\linewidth]{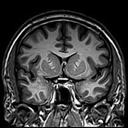} \\
	18.74 & 19.23 & 20.48 & 21.37 & 20.62 & 20.91 & 22.02 & PSNR \\
	\includegraphics[width=0.116\linewidth]{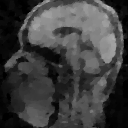}
	& \includegraphics[width=0.116\linewidth]{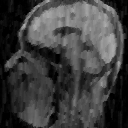}
	& \includegraphics[width=0.116\linewidth]{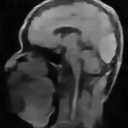}
	& \includegraphics[width=0.116\linewidth]{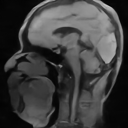}
	& \includegraphics[width=0.116\linewidth]{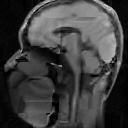}
	& \includegraphics[width=0.116\linewidth]{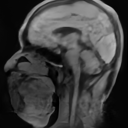}
	& \includegraphics[width=0.116\linewidth]{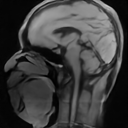}
	& \includegraphics[width=0.116\linewidth]{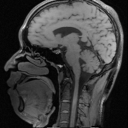} \\
	24.89 & 24.47 & 26.85 & 27.90 & 26.72 & 27.74 & 28.65 & PSNR \\
	\end{tabular} 
	\caption{Visual and numerical CS-MRI reconstruction comparison against  the state-of-the-art techniques on medical images. The numerical values denote the PSNR obtained by each technique. (Details are better appreciated on screen).
	}
	\label{fig:comparison-cs-mri}
\end{figure}

\subsection{Compressed Sensing MRI} \label{sec:csmri}
Magnetic resonance imaging (MRI) \citep{560324} is an essential imaging modality for diagnosing and evaluating a host of conditions and diseases, but suffers from slow data acquisition. Compressed sensing MRI (CS-MRI) \citep{lustig2008compressed} accelerates MRI by acquiring less data through subsampling. 
Recovering high-quality images from undersampled MRI data is  an ill-posed inverse imaging problem. 
The forward model of CS-MRI can be mathematically described as: $ y = \mathcal{F}_p x + \omega$, 
where $x \in \mathbb{C}^N$ is the underlying image,  the operator $\mathcal{F}_p : \mathbb{C}^N \rightarrow \mathbb{C}^M$, with $M < N$, denotes the partially-sampled Fourier transform, and $\omega \sim \mathcal{N} \left(0, \sigma_n I_M \right)$ is the additive white Gaussian noise. The data-fidelity term, for the MRI reconstruction, is $\mathcal{D} (x) = \frac{1}{2} \| y - \mathcal{F}_p x \|^2$
whose proximal operator is described in \citep{eksioglu2016decoupled}.

\paragraph{Denoiser Priors}
To show how denoiser priors affect the performance of the PnP, we train three state-of-the-art CNN-based denoisers, \ie DnCNN \citep{zhang2017beyond}, MemNet \citep{Tai_2017_ICCV} and residual U-net \citep{ronneberger2015u}, with a tunable noise level map. We compare both the Gaussian denoising performance and the PnP performance\footnote{We exhaustively search the best denoising strength/penalty parameters to exclude the impact of internal parameters.} using these denoisers. 
As shown in Table \ref{tb:denoiser-profile}, the residual U-net and MemNet consistently outperform DnCNN in terms of denoising and CS-MRI. 
It seems to imply a better Gaussian denoiser is also a better denoiser prior for the PnP framework\footnote{Further investigation of this argument can be found in Section \ref{sec: gaussian-denoiser}.}. 
Since U-net is significantly faster than MemNet and yields slightly better performance, we choose U-net as our denoiser prior. 

\paragraph{Comparisons of Different Policies} \label{sec: policy-comparison}
We start by providing some insights of our learned policy  by comparing the performance
of PnP-ADMM with different policies: i) the handcrafted policy
used in IRCNN~\citep{Zhang_2017_CVPR}; ii) the fixed policy that uses fixed parameters ($\sigma=15$, $\mu=0.1$); iii) the fixed optimal policy that adopts fixed parameters searched to maximize the average PSNR across all testing images;
iv) the greedy policy that selects parameters in a greedy manner to maximize the PSNR at each iteration during optimization; 
v) the oracle policy that uses different parameters for different images such that the PSNR of each image is maximized and  vi) our learned policy network to optimize parameters for each image. We remark that all compared policies are run for 30 iterations whilst ours automatically chooses the termination time.

To understand the usefulness of the early stopping
mechanism, we also report the results of these policies with optimal early stopping\footnote{It should be noted that some policies (\eg ``fixed optimal"
and ``oracle") require to access the ground truth to determine parameters, which is generally impractical in real testing scenarios. }.
Moreover, we analyze whether the model-based RL benefits our algorithm by comparing it with the learned policy by model-free RL whose $\pi_2$ is optimized using the model-free DDPG algorithm \citep{lillicrap2015continuous}.

The results of all aforementioned policies are provided in 
Table \ref{tb:policy-comparison}. We can see that  bad choice of parameters (see ``fixed") induces poor results, in which the early stopping is  required to rescue performance (see ``fixed$^*$"). 
When the parameters are properly assigned, the early stopping would be helpful to reduce computation cost. The greedy policy serves as an upper bound of SURE-based parameter tuning methods if applicable, which however often yields suboptimal results. Our learned policy leads to fast practical convergence as well as excellent performance, sometimes even outperforms the oracle policy tuned via inaccessible ground truth (in $2\times$ and $8\times$ case). We note this is owing to the varying parameters across iterations  generated automatically in our algorithm, which yield  extra flexibility than constant parameters over iterations.
Besides, we find that the learned model-free policy produces suboptimal denoising strength/penalty parameters compared with our mixed model-free and model-based policy, and it also fails to learn early stopping.


\paragraph{Comparisons with State-of-the-arts}
 
We compare our method against six state-of-the-art methods for CS-MRI, including the traditional optimization-based approaches (RecPF \citep{yang2010fast} and FCSA \citep{huang2010mri}), the PnP approaches (BM3D-MRI \citep{eksioglu2016decoupled} and IRCNN \citep{Zhang_2017_CVPR}), and the  unrolling approaches (ADMMNet \citep{NIPS2016_6406} and ISTANet \citep{Zhang_2018_CVPR}).
To keep comparison fair, for each algorithm unrolling method, only a single network is trained to tackle all the cases using the same dataset as ours.  All parameters involved in the competing methods are manually tuned optimally or automatically chosen as described in
the reference papers. 
Table \ref{tb:CS-MRI} shows the method performance on two medical image datasets:  7 widely used medical images (Medical7) \citep{huang2010mri} and 50 medical images from MICCAI 2013 grand challenge dataset\footnote{https://my.vanderbilt.edu/masi/}. 
The visual comparison can be found in Figure~\ref{fig:comparison-cs-mri}. It can be seen that our approach significantly outperforms the state-of-the-art PnP method (IRCNN) by a large margin, especially under the difficult $8\times$ 
case. In the low acceleration cases (\eg $2\times$), our algorithm only runs 5 iterations to arrive at the desired performance, in contrast with 30 or 70 iterations required in IRCNN and BM3D-MRI respectively.  

\begin{figure}[t]
	\centering
	\setlength\tabcolsep{1.5pt}
	\footnotesize
	\begin{tabular}{cccccccc}		
	 \fontsize{9pt}{9pt}\selectfont FBP &  \fontsize{9pt}{9pt}\selectfont TV & \fontsize{9pt}{9pt}\selectfont FBPconv & \fontsize{9pt}{9pt}\selectfont RED-CNN & \fontsize{9pt}{9pt}\selectfont LPD & \fontsize{9pt}{9pt}\selectfont RPGD  & \fontsize{9pt}{9pt}\selectfont Ours & \fontsize{8pt}{8pt}\selectfont Ground Truth  \\
	\includegraphics[width=0.116\linewidth]{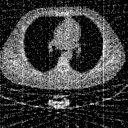}
	& \includegraphics[width=0.116\linewidth]{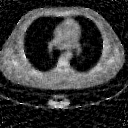}
	& \includegraphics[width=0.116\linewidth]{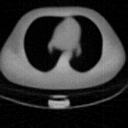}
	& \includegraphics[width=0.116\linewidth]{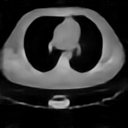}
	& \includegraphics[width=0.116\linewidth]{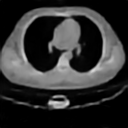}
	& \includegraphics[width=0.116\linewidth]{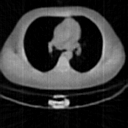}
	& \includegraphics[width=0.116\linewidth]{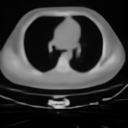}
	& \includegraphics[width=0.116\linewidth]{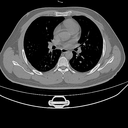} \\
	18.30 & 20.91 & 22.16 & 22.34 &  22.72 & 23.19  & 23.21 & PSNR \\
	\includegraphics[width=0.116\linewidth]{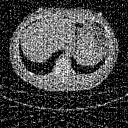}
	& \includegraphics[width=0.116\linewidth]{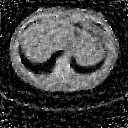}
	& \includegraphics[width=0.116\linewidth]{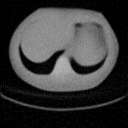}
	& \includegraphics[width=0.116\linewidth]{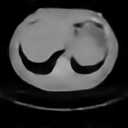}
	& \includegraphics[width=0.116\linewidth]{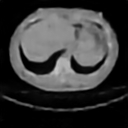}
	& \includegraphics[width=0.116\linewidth]{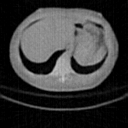}
	& \includegraphics[width=0.116\linewidth]{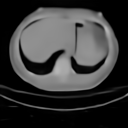}
	& \includegraphics[width=0.116\linewidth]{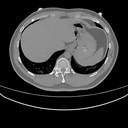} \\
	15.54 & 18.16 & 22.55 & 22.87 & 22.95 & 23.62 & 24.43 & PSNR \\
	\includegraphics[width=0.116\linewidth]{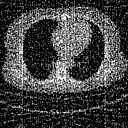}
	& \includegraphics[width=0.116\linewidth]{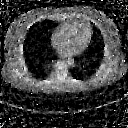}
	& \includegraphics[width=0.116\linewidth]{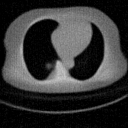}
	& \includegraphics[width=0.116\linewidth]{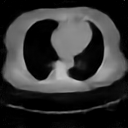}
	& \includegraphics[width=0.116\linewidth]{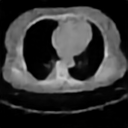}
	& \includegraphics[width=0.116\linewidth]{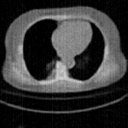}
	& \includegraphics[width=0.116\linewidth]{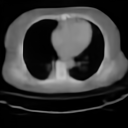}
	& \includegraphics[width=0.116\linewidth]{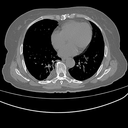} \\
	14.33 & 17.12 & 22.71 & 22.18 &  22.50 & 23.13 & 23.48 & PSNR \\
	\end{tabular} 
	\caption{Sparse-view CT reconstruction results and their corresponding PSNR values for all compared methods. The results are produced on several medical images using 30 views. (Best view on screen with zoom)
	}
	\label{fig:comparison-ct}
\end{figure}

\subsection{Sparse-View Computed Tomography}
Computed tomography (CT) \citep{herman2009fundamentals} utilizes X-ray measurements taken from different angles to produce tomographic images of a body, which is also widely used in clinical applications as MRI. 
Given the potential risk of X-ray radiation to the patient, low-dose CT has been intensively studied in response to concerns of patient safety in full-dose CT scans.  Typically, low-dose CT scan can be achieved either by reducing the number of projection views (sparse-view CT) or by lowering the operating current of an X-ray tube. 
Here, we focus on the sparse-view CT reconstruction with the goal to reduce the number of views (or dose) for CT imaging while retaining the quality of full-dose reconstruction.  The forward operator is given by the Radon transform \citep{natterer2001mathematics} that integrates the signal over a set of lines given by a sparse acquisition geometry. 
We implement Radon transform by the \textit{TorchRadon} library \citep{ronchetti2020torchradon}, which can be seamlessly integrated into PyTorch with full differentiability. 
We validate our method on a challenging case of sparse-view CT reconstruction, using a sparse 30-view\footnote{Standard full-view reconstruction is often performed by 720-view projections \citep{jin2017deep,gupta2018cnn}. The sparse 30-view reconstruction corresponds to $24\times$ dosage reduction.  } parallel beam geometry with 182 detector pixels to generate projections, blended with varying amounts (5\%/7.5\%/10\%) of Gaussian noise. 

The competing methods for CT reconstruction include filtered backprojection (FBP) \citep{herman2009fundamentals}, total variation (TV) reconstruction \citep{sidky2008image}, RED-CNN  \citep{chen2017low}, FBPconv \citep{jin2017deep}, learned primal dual (LPD) reconstruction \citep{adler2018learned} and RPGD \citep{gupta2018cnn}. The former two are the traditional methods, while the latter four are the state-of-the-art deep learning methods. For TV reconstruction, we adopt the PDHG solver \citep{chambolle2011first} with optimally tuned parameters. 
For the learning-based methods,  we train their models using the same dataset as ours, and manually tune configurations to achieve  performance as higher as possible. The numerical and visual results are evaluated on 50 randomly selected lung CT images from the  COVID-19 CT lung and infection segmentation dataset \citep{ma2020towards}, shown in Table~\ref{tb:sparse-view-CT} and Figure~\ref{fig:comparison-ct} respectively.  Our TFPnP algorithm yields considerably better results than state-of-the-arts (\eg it outperforms the second-best method RPGD 0.4dB in terms of PSNR under 7.5\% Gaussian noise setting), suggesting the effectiveness of our algorithm.

\begin{table}[!t]
\centering
\begin{tabular}{lccc} 
		\toprule
		 & $5\%$ & $7.5\%$ & $10\%$ \\ \cline{2-4} 
		\textsc{Algorithms} & PSNR  & PSNR  & PSNR  \\  \midrule
		FBP & 18.37 & 16.16 & 14.32 \\
		TV  & 21.47 &  19.17 & 16.95 \\
	    FBPconv & 23.00 & 22.94 & 22.86 \\
		RED-CNN & 23.45 & 23.11 & 22.71 \\
		LPD & 23.86 & 23.44 & 23.00 \\
		RPGD & \textcolor{blue}{24.29} & \textcolor{blue}{23.82} & \textcolor{blue}{23.38} \\
		Ours  & \textcolor{orange}{24.50} & \textcolor{orange}{24.23} & \textcolor{orange}{23.72} \\ 
		\bottomrule
\end{tabular}
\caption{Quantitative results (PSNR) of Sparse-view (30 views) CT reconstruction with varying degrees (5\%, 7.5\%, 10\%) of Gaussian noise added into the projections.}
\label{tb:sparse-view-CT}
\end{table}

\begin{table}[!t]
\centering
\begin{tabular}{lccc} 
		\toprule
		& $K = 4$ & $K = 6$ & $K = 8$ \\ \cline{2-4} 
		\textsc{Algorithms} & PSNR  & PSNR  & PSNR  \\  \midrule
		MLE  & 13.48 &  17.51 & 20.12 \\
		Chan \etal & 25.11 & 27.26 & 28.67 \\
		Ryu \etal & \textcolor{blue}{25.26} & \textcolor{blue}{27.80} & \textcolor{blue}{29.18} \\		
		Ours  & \textcolor{orange}{25.55} & \textcolor{orange}{28.47} & \textcolor{orange}{30.32} \\ 
		\bottomrule
\end{tabular}
\caption{Quantitative results (PSNR) of single-photon imaging with different oversampling ratios ($K=4, 6, 8$)}
\label{tb:single-photon-imaging}
\end{table}

\subsection{Single-photon Imaging}
Quanta imaging sensor (QIS) is an emerging class of solid-state imaging sensors that 
are capable of detecting individual photons in space and time \citep{fossum2011quanta,fossum2016quanta}. 
The principle of QIS is analogous to photographic film:
photon flux hitting a pixel area of the single-photon detector, triggers a binary response (1-bit signal) 
revealing the intensity of light during exposure.
The 1-bit measurements acquired by the QIS follow a quantized
Poisson process, which must be “decoded” to recover the underlying image \citep{yang2011bits,chan2014efficient}. 
QIS is a spatial oversampling device composed of many tiny single-photon detectors called jots. In each unit space, $K$ jots are used to acquire light corresponding to a pixel in the usual sense  (\eg a pixel in a CMOS sensor). 
By assuming homogeneous distribution of the light within each pixel, we consider single-photon imaging using a simplified QIS imaging model which relates the underlying image $x \in \mathbb{R}^N$ and the 1-bit measurements $y \in \mathbb{R}^{NK}$ as
$$ y = \textbf{1}\left( z \ge 1 \right), \quad  z \sim \mathcal{P} \left(\alpha_s G x\right) $$
where $G: \mathbb{R}^N \to \mathbb{R}^{NK}$ is an oversampling operator that duplicates each pixel to $K$ pixels, $\alpha_s \in \mathbb{R}$ is sensor gain, $\mathcal{P}$ denotes the Poisson distribution, $\textbf{1} (\cdot \ge 1)$ is an indicator function that executes binary quantization on threshold 1. 

The goal is to reconstruct the unknown image of interest $x$ from the observed binary photons $y$.
Given the log likelihood function of the quantized Poisson process \citep{yang2011bits}, 
the corresponding data-fidelity term $\mathcal{D}$ can be defined as
\begin{equation} \label{eq:data-term-spi}
\mathcal{D}(x) = \sum_{j=1}^{N} - K_j^0 \mathop{\mathrm{log}} ( e^{-\frac{\alpha_s x_j}{K}} ) - K_j^1 \mathop{\mathrm{log}} (1 - e^{-\frac{\alpha_s x_j}{K}} ), 
\end{equation}  
where $K_j^1 = \sum_{i=1}^K y_{(j-1)K+i}$ and $K_j^0 = \sum_{i=1}^K (1 - y_{(j-1)K+i})$ denote the numbers of ones and zeros in the $j$-th unit pixel (containing $K$ jots) respectively. 
The proximal operator of \eqref{eq:data-term-spi} is given in \citep{chan2014efficient}, which requires a numerical scheme to address a 
root finding problem of a transcendental equation.  In practice, we follow \citep{ryu2019plug} to use a vectorized bisection algorithm 
for root finding, and transplant their method into Pytorch with differentiable programming. The iteration times of bisection are empirically set to 10, as a good balance between computation precision and cost.

\begin{figure}[!t]
	\centering
	\setlength\tabcolsep{1.5pt}
	\footnotesize
	\begin{tabular}{cccccc}		
	 \fontsize{9pt}{9pt}\selectfont Binary input &  \fontsize{9pt}{9pt}\selectfont MLE  & \fontsize{9pt}{9pt}\selectfont Chan \etal & \fontsize{9pt}{9pt}\selectfont Ryu \etal  & \fontsize{9pt}{9pt}\selectfont Ours & \fontsize{8pt}{8pt}\selectfont Ground Truth  \\
	\includegraphics[width=0.133\linewidth]{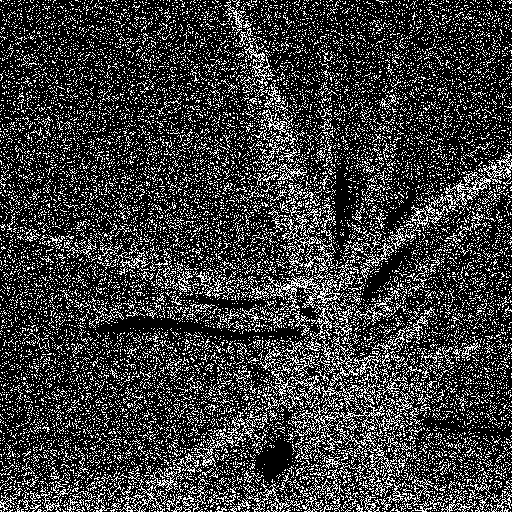}
	& \includegraphics[width=0.133\linewidth]{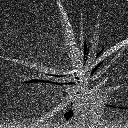}
	& \includegraphics[width=0.133\linewidth]{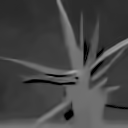}
	& \includegraphics[width=0.133\linewidth]{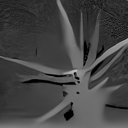}
	& \includegraphics[width=0.133\linewidth]{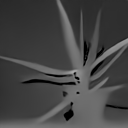}
	& \includegraphics[width=0.133\linewidth]{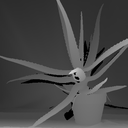} \\
	-  & 16.34 &  29.09 & 29.40  & 30.32 & PSNR \\
	\includegraphics[width=0.133\linewidth]{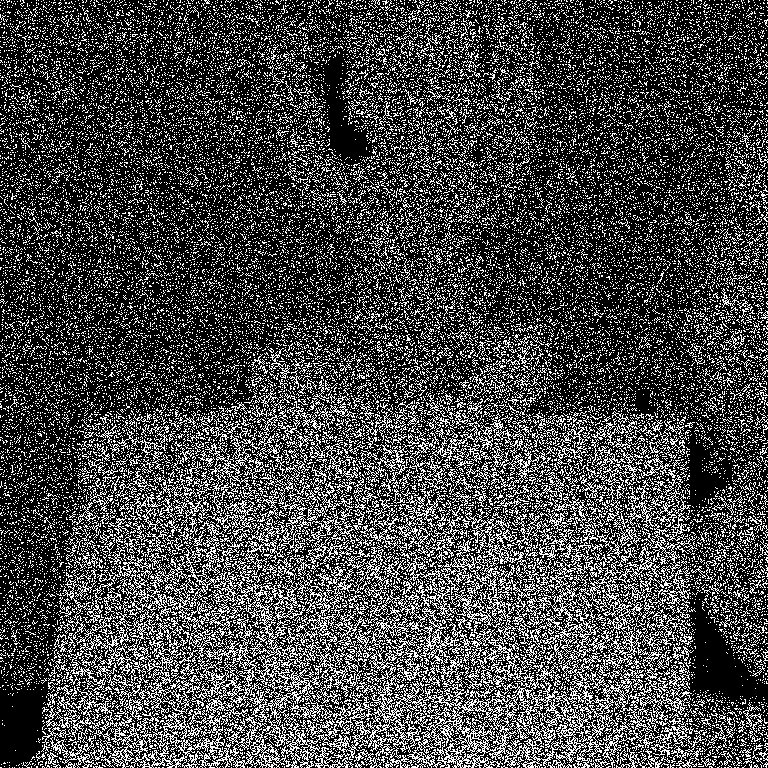}
	& \includegraphics[width=0.133\linewidth]{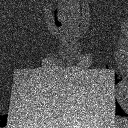}
	& \includegraphics[width=0.133\linewidth]{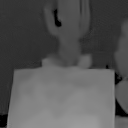}
	& \includegraphics[width=0.133\linewidth]{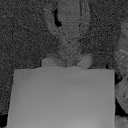}
	& \includegraphics[width=0.133\linewidth]{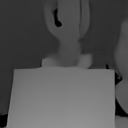}
	& \includegraphics[width=0.133\linewidth]{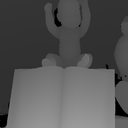} \\
	-  & 19.20 &  35.42 & 29.42  & 35.81 & PSNR \\
	\includegraphics[width=0.133\linewidth]{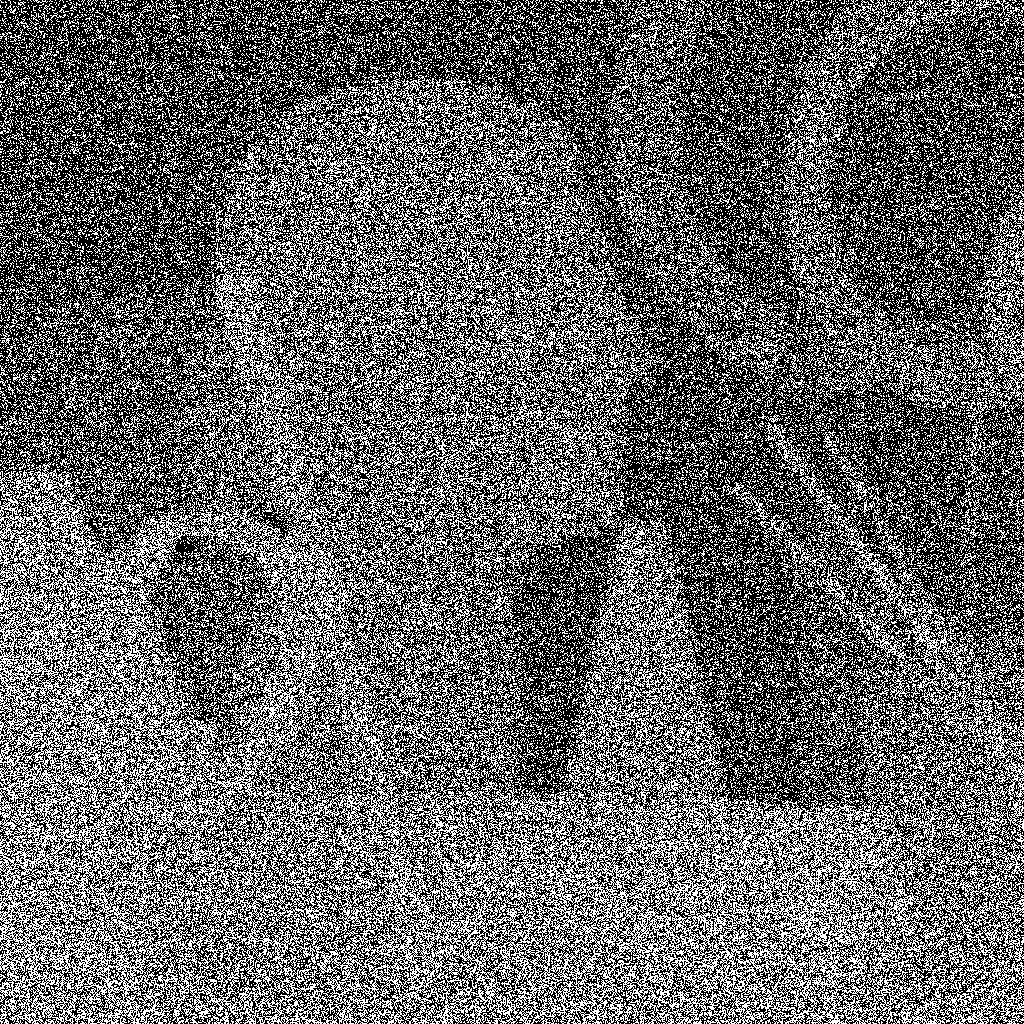}
	& \includegraphics[width=0.133\linewidth]{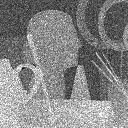}
	& \includegraphics[width=0.133\linewidth]{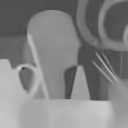}
	& \includegraphics[width=0.133\linewidth]{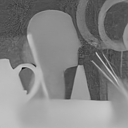}
	& \includegraphics[width=0.133\linewidth]{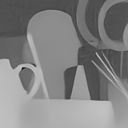}
	& \includegraphics[width=0.133\linewidth]{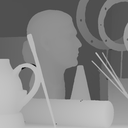} \\
	-  & 19.44 &  31.08 & 32.48  & 33.32 & PSNR \\
	\end{tabular} 
	\caption{Single-photon imaging results and their associated PSNR values for our technique and the compared methods. (Best view on screen with zoom)}
	\label{fig:comparison-spi}
\end{figure}

We consider three sets of experiments with different oversampling ratios 
($K = 4, 6, 8$ along both horizontal and vertical directions). 
The sensor gain is set as $\alpha_s = K^2$. 
For comparison, we choose the three existing algorithms, including
the maximum likelihood estimation (MLE) method \citep{yang2011bits}, 
and two PnP methods in \citep{chan2017plug} and \citep{ryu2019plug}. 
The visual and numerical results are shown in Figure~\ref{fig:comparison-spi} and Table~\ref{tb:single-photon-imaging} respectively. 
From these results, one can observe that our method reconstructed the highest-quality images that closely resemble the ground truth, conforming with the quantitative assessment via PSNR. 

\begin{figure}[!t]
	\centering
	\setlength\tabcolsep{1.5pt}
	\footnotesize
	\begin{tabular}{cccccccc}		
	\fontsize{9pt}{9pt}\selectfont HIO & \fontsize{9pt}{9pt}\selectfont WF & \fontsize{9pt}{9pt}\selectfont DOLPHIn & \fontsize{9pt}{9pt}\selectfont SPAR & \fontsize{5.5pt}{5.5pt}\selectfont BM3D-prGAMP & \fontsize{9pt}{9pt}\selectfont prDeep & \fontsize{9pt}{9pt}\selectfont Ours & \fontsize{7.5pt}{7.5pt}\selectfont Ground Truth   \\
	\includegraphics[width=0.116\linewidth]{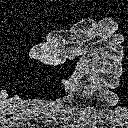}
	& \includegraphics[width=0.116\linewidth]{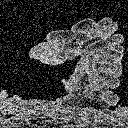}
	& \includegraphics[width=0.116\linewidth]{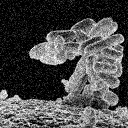}
	& \includegraphics[width=0.116\linewidth]{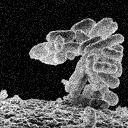}
	& \includegraphics[width=0.116\linewidth]{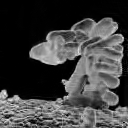}
	& \includegraphics[width=0.116\linewidth]{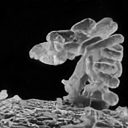}
	& \includegraphics[width=0.116\linewidth]{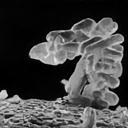}
	& \includegraphics[width=0.116\linewidth]{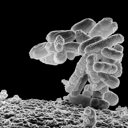} \\
	14.36 & 15.37 & 21.13 & 23.86 & 23.40 & 24.47 & 25.02 & PSNR \\
	\includegraphics[width=0.116\linewidth]{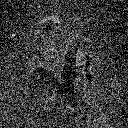}
	& \includegraphics[width=0.116\linewidth]{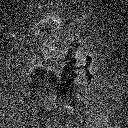}
	& \includegraphics[width=0.116\linewidth]{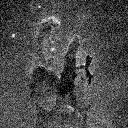}
	& \includegraphics[width=0.116\linewidth]{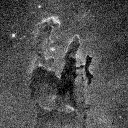}
	& \includegraphics[width=0.116\linewidth]{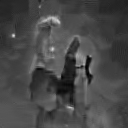}
	& \includegraphics[width=0.116\linewidth]{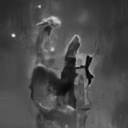}
	& \includegraphics[width=0.116\linewidth]{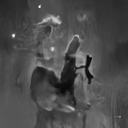}
	& \includegraphics[width=0.116\linewidth]{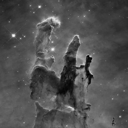} \\
    14.40 & 15.52 & 19.35 & 22.48 & 25.66 & 27.72 & 28.01 & PSNR \\
	\includegraphics[width=0.116\linewidth]{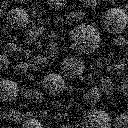}
	& \includegraphics[width=0.116\linewidth]{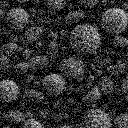}
	& \includegraphics[width=0.116\linewidth]{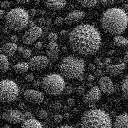}
	& \includegraphics[width=0.116\linewidth]{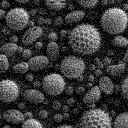}
	& \includegraphics[width=0.116\linewidth]{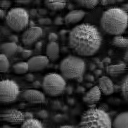}
	& \includegraphics[width=0.116\linewidth]{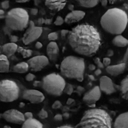}
	& \includegraphics[width=0.116\linewidth]{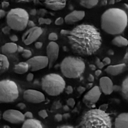}
	& \includegraphics[width=0.116\linewidth]{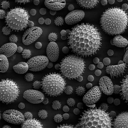} \\
	15.10 & 16.27 & 19.62 & 22.51 & 23.61 & 24.59 & 25.12 & PSNR \\
	\end{tabular} 
	\caption{Visual and numerical comparison of the recovered images, from noisy intensity-only coded diffraction pattern (CDP). The values denote the PSNR for each corresponding image. (Details are better appreciated on screen)}
	\label{fig:comparison-phase-retrieval}
\end{figure}

\begin{table}[!htbp]
\centering
\begin{tabular}{lccc} 
		\toprule
		 & $\alpha=9$ & $\alpha=27$ & $\alpha=81$ \\ \cline{2-4} 
		\textsc{Algorithms} & PSNR  & PSNR  & PSNR  \\  \midrule
		HIO & 35.96 & 25.76 & 14.82 \\
		WF  & 34.46 &  24.96 & 15.76 \\
		DOLPHIn & 29.93 & 27.45 & 19.35 \\
		SPAR & 35.20 & 31.82 & 22.44 \\
		BM3D-prGAMP & \textcolor{blue}{40.25} & 32.84 & 25.43 \\
		prDeep  & 39.70 & \textcolor{blue}{33.54} & \textcolor{blue}{26.82} \\
		Ours  & \textcolor{orange}{40.33} & \textcolor{orange}{33.90} & \textcolor{orange}{27.23} \\ 
		\bottomrule
\end{tabular}
\caption{Quantitative comparison of different phase retrieval algorithms. The results are computed on four coded diffraction pattern measurements and  varying the amount of Possion noise (a large $\alpha$ indicates low sigma-to-noise ratio). The best results are displayed in \textcolor{orange}{orange} colour and the second best in \textcolor{blue}{blue} colour.
}
\label{tb:phase-retrieval}
\end{table}

\subsection{Phase Retrieval}
The goal of phase retrieval (PR) \citep{gerchberg1972practical} is to recover the underlying image from only the amplitude of the output of a complex linear system. It appears in many optical imaging applications, \eg X-ray crystallography \citep{millane1990phase} and Fourier ptychography \citep{zheng2013wide-field}.   Mathematically, PR can be defined as the problem of recovering a signal $x\in \mathbb{R}^{N}$ or $\mathbb{C}^{N}$ from measurement $y$ of the form: 
   $$ y = \sqrt{\left| Ax \right|^2 + \omega}, $$
where the measurement matrix $A$ represents the forward operator of the system, and $\omega$ represents shot noise. We approximate it with $\omega \sim \mathcal{N} (0, \alpha^2|Ax|^2)$. The term $\alpha$ controls the sigma-to-noise ratio in this problem. 

We test algorithms for phase retrieval with coded diffraction pattern (CDP) measurements \citep{CandPhase2015pr}. Multiple measurements with different random spatial modulator (SLM) patterns are recorded. We model the capture of four measurements using a phase-only SLM as \citep{metzler2018prdeep}. 
Each measurement operator can be mathematically described as $A_i = \mathcal{F}D_i, \quad i \in [1,2,3,4]$,
where $\mathcal{F}$ can be represented by the 2D Fourier transform 
and $D_i$ is diagonal matrices with nonzero elements drawn uniformly from the unit circle in the complex planes. 

We compare our method with three classic approaches (HIO \citep{Fienup1982Phase}, WF \citep{candes2014phase}, and DOLPHIn \citep{MairalDOLPHIn}) and three PnP approaches (SPAR \citep{katkovnik2017phase}, BM3D-prGAMP \citep{Metzler2016BM3D} and prDeep \citep{metzler2018prdeep}). 
Like CS-MRI and sparse-view CT applications, the parameters of these algorithms are carefully selected. 
Table \ref{tb:phase-retrieval} and Figure~\ref{fig:comparison-phase-retrieval} summarize the results of all competing methods on twelve images used in \citep{metzler2018prdeep}. It can be seen that our method still leads to state-of-the-art performance in this nonlinear inverse problem, and produces cleaner and clearer results than other competing methods.

\section{Algorithm Investigation} \label{sec:investigate}
In Section \ref{sec:main-exp}, we have shown the applications of our algorithms on both linear and nonlinear inverse imaging problems. 
We conduct more experiments to further investigate our TFPnP algorithm. We focus on four aspects of the algorithm: i) the choice of hyperparameters $m$, $N$ and $\eta$, ii) the relationship between the Gaussian denoising performance and the PnP performance, iii) the impact of auxiliary setting-related information feed into the policy, and iv) the behavior of the learned policy---how do the generated internal parameters (the denoising strength $\sigma_k$ and the penalty parameter $\mu_k$) look like? 
All experiments are carried out for the application of CS-MRI.

\begin{table}[!t]
\centering
{
\begin{tabular}{lcccccc} 
		\toprule
		 & \multicolumn{2}{c}{$2\times$} & \multicolumn{2}{c}{$4\times$} & \multicolumn{2}{c}{$8\times$} \\ \cline{2-7} 
		($m$, $N$, $\eta$) & PSNR & \#IT.  & PSNR & \#IT.  & PSNR & \#IT. \\  \midrule
		(3, 10, 0.05) & 30.26 & 3.0 & 28.59 & 9.9 & 26.27 & 16.7 \\
		(10, 3, 0.05) & 30.32 & 10.0 & 28.57 & 10.0 & 26.39 & 10.0 \\
		(5, 6, 0.05) & 30.33 & 5.0 & 28.58 & 10.0 & 26.52 & 15.0 \\
		\midrule
		(5, 6, 0) & 30.30 & 27.1 & 28.60 & 29.3 & 26.40 & 30.0 \\
	    (5, 6, 0.1)  & 30.34 & 5.0 & 28.44 & 5.0 & 26.29 & 10.7 \\
	    (5, 6, 0.25)  & 30.34 & 5.0 & 28.37 & 5.0 & 25.53 & 5.0 \\
		\bottomrule
\end{tabular}}
\caption{Numerical comparison of the learned policies trained with different hyperparameters. The results are computed for the application of CS-MRI on seven medical images using undersampling factors of \{2$\times$, 4$\times$, 8$\times$\} and noise level of 15. The numerical values refer to the PSNR and the  number of iterations (\#IT.) 
}
\label{tb:hyperparameter}
\end{table}

\subsection{Hyperparameter Analysis} \label{sec: hyperparameter}
We discuss the choice of hyperparameters $m$, $N$ and $\eta$ in our TFPnP algorithm, where $m$ denotes the number of iterations of the optimization involved in the transition function $p$, $N$ is the maximum time step to run the policy, and $\eta$ sets the degree of penalty defined in the reward function. 
Table \ref{tb:hyperparameter} shows the results of learned policies trained with different hyperparameter settings ($m$, $N$ and $\eta$) for CS-MRI. 
These results are divided into two groups (separated by the middle line) to analyze the effects of $(m, N)$ and $\eta$ respectively. 
Note that $N$ is varied with $m$ such that the maximum number of iterations $m\times N$ is fixed to 30 to keep comparisons fair. 
In the first group, we fix the value of $\eta$ and change the value of $m$.  We observe all these settings yield similar results. It is up to users' preference to choose $m$, where larger $m$ leads to coarser control of the termination time, but only requires fewer rounds of the decision making. In the second group, we keep $(m, N)$ constant while manipulating the value of $\eta$. We find $\eta$ serves as a key parameter to encourage early stopping. With $\eta=0$, the policy would not learn to early stop the optimization process, whilst large $\eta$ would make the policy stop too early.

\begin{figure}[!t]
\centering
\includegraphics[width=.5\linewidth,clip,keepaspectratio]{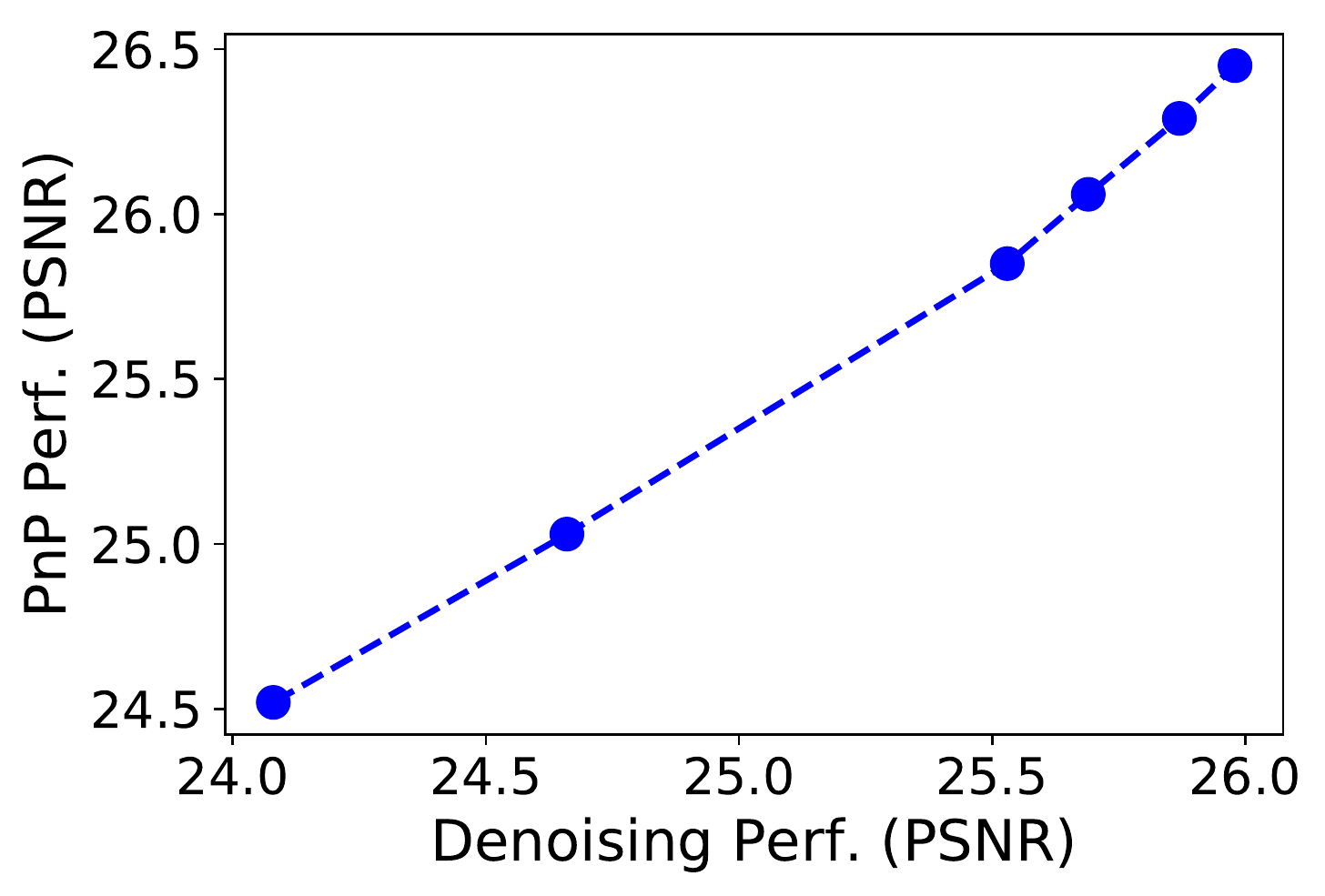}
\caption{Performance (Perf.) behavior between Gaussian denoising and PnP-ADMM (with oracle policy) in terms of PSNR. The better the Gaussian denoiser the better the denoiser prior.}
\label{fig:denoising-PnP}
\end{figure}

\subsection{The Effect from the Gaussian Denoiser} \label{sec: gaussian-denoiser}
To reveal the relationship between the Gaussian denoiser and the PnP performance, 
we train a set of denoising networks (U-net) with incremental Gaussian denoising performance (by adjusting the number of filters of networks). Then we incorporate these denoisers into the PnP-ADMM framework to evaluate their PnP performance on the CS-MRI application. The denoising strength/penalty parameters and the termination time are exhaustively searched to maximize PSNR to exclude the impact of internal parameters. 
Figure~\ref{fig:denoising-PnP} clearly illustrates the positive correlation between the Gaussian denoising and PnP performance, indicating that a better Gaussian denoiser could serve as a better image prior when plugging into the PnP framework. 

\begin{table}[!t]
\centering
{
\begin{tabular}{lcccccc} 
		\toprule
		 & \multicolumn{2}{c}{$2\times$} & \multicolumn{2}{c}{$4\times$} & \multicolumn{2}{c}{$8\times$} \\ \cline{2-7} 
		 & PSNR & \#IT.  & PSNR & \#IT.  & PSNR & \#IT. \\  \midrule
		w/o extra info & 30.15 & 27.4 & 28.36 & 29.1 & 26.27 & 30.0 \\
		+ ``noise level" & 30.31 & 9.1 & 28.53 & 15.0 & 26.34 & 21.4 \\
		+ ``step" & 30.29 & 10.0 & 28.48 & 17.5 & 26.37 & 23.6 \\
	    Ours (full setting) & 30.33 & 5.0 & 28.58 & 10.0 & 26.52 & 15.0 \\
		\bottomrule
\end{tabular}}
    \caption{Investigation of auxiliary setting-related information, \ie the measurement noise level (denoted by "noise level") and the number of steps that have been taken so far (denoted by "step") encoded into the policy network. The results are evaluated on CS-MRI with speed-up of \{2$\times$, 4$\times$, 8$\times$\} and noise level of 15.}
\label{tb:setting-related-info}
\end{table}


\subsection{The Effect from Auxiliary Setting-related Information}
Our policy network makes use of auxiliary setting-related information (\ie the measurement noise level, and the number of steps that has been taken) encoded as additional input feature planes to learn a problem-aware policy. 
To analyze its impact, we train policy networks without or with partial auxiliary information. As shown in Table~\ref{tb:setting-related-info}, by encoding these setting-related quantities, the policy network learns to be aware of the problem settings, thus deriving a better parameter-tuning policy with faster convergence and higher accuracy.

\begin{figure}[!t]
	\centering
	\setlength\tabcolsep{1.5pt}
	\begin{tabular}{ccccccc}		
	\includegraphics[width=0.134\linewidth]{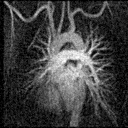}
	& \includegraphics[width=0.134\linewidth]{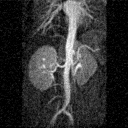}
	& \includegraphics[width=0.134\linewidth]{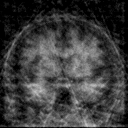}
	& \includegraphics[width=0.134\linewidth]{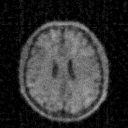}
	& \includegraphics[width=0.134\linewidth]{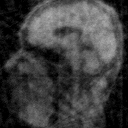}
	& \includegraphics[width=0.134\linewidth]{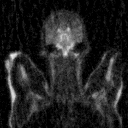}
	& \includegraphics[width=0.134\linewidth]{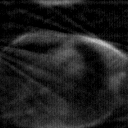} \\
	\includegraphics[width=0.134\linewidth]{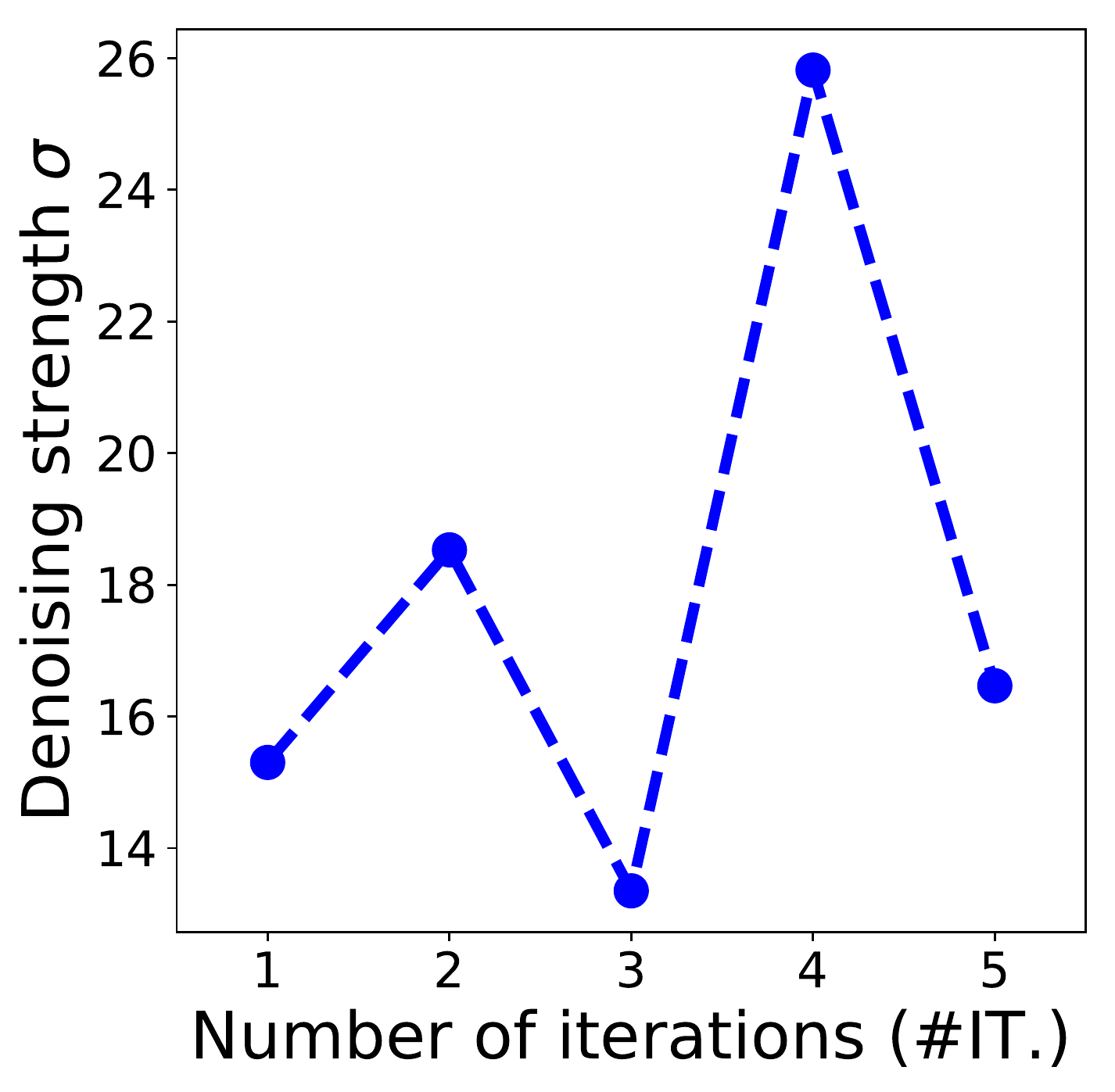}
	& \includegraphics[width=0.134\linewidth]{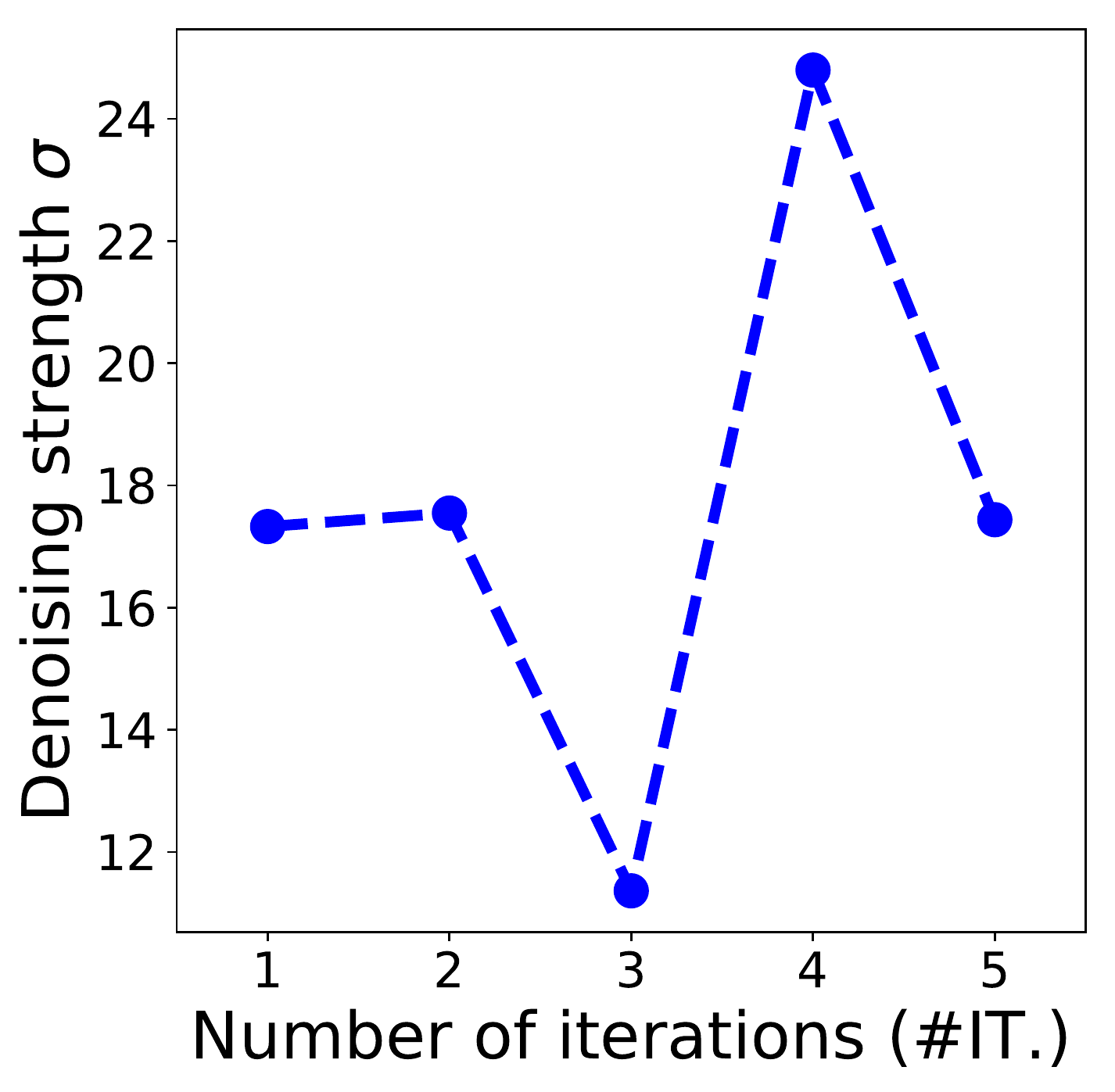}
	& \includegraphics[width=0.134\linewidth]{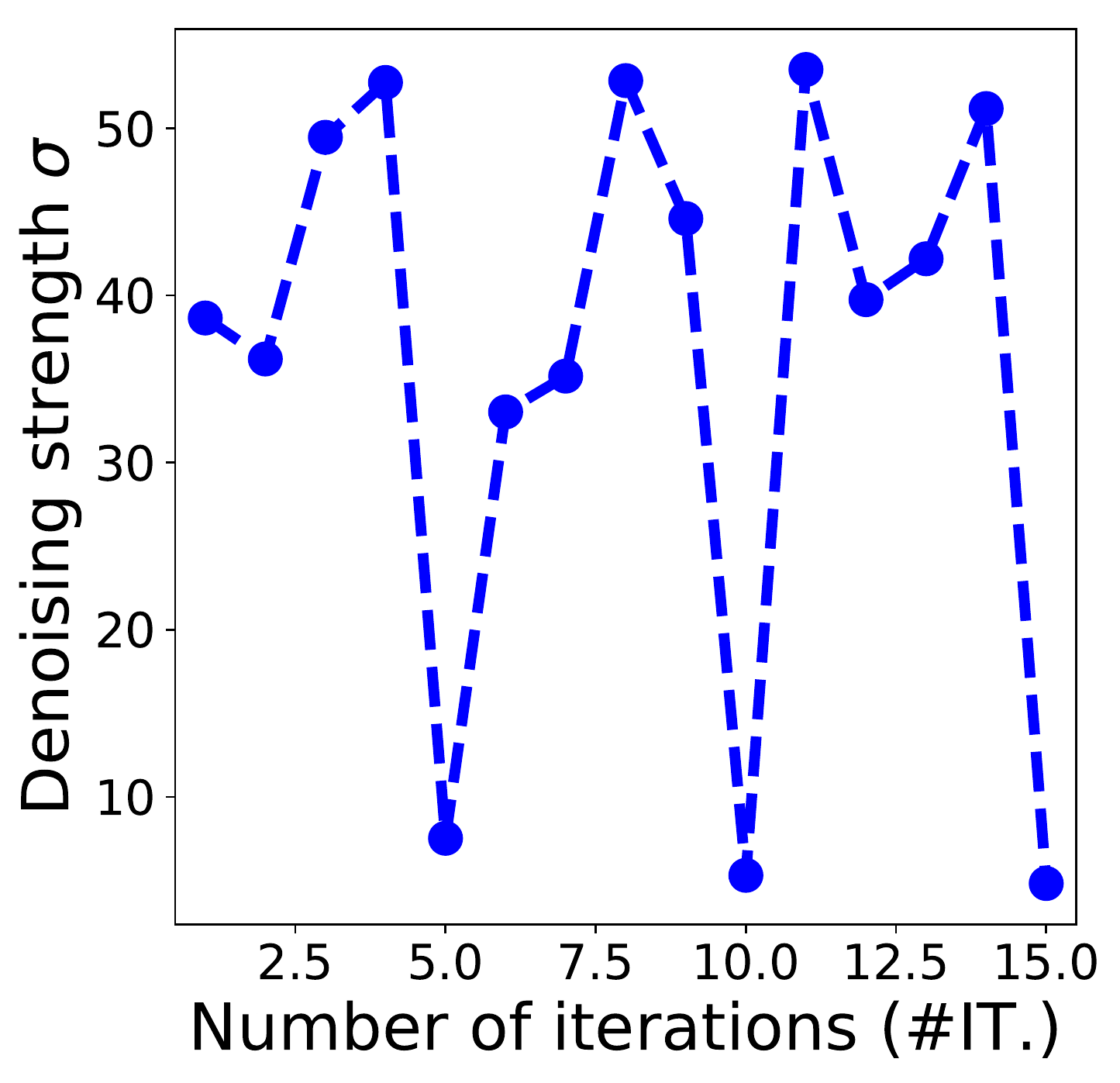}
	& \includegraphics[width=0.134\linewidth]{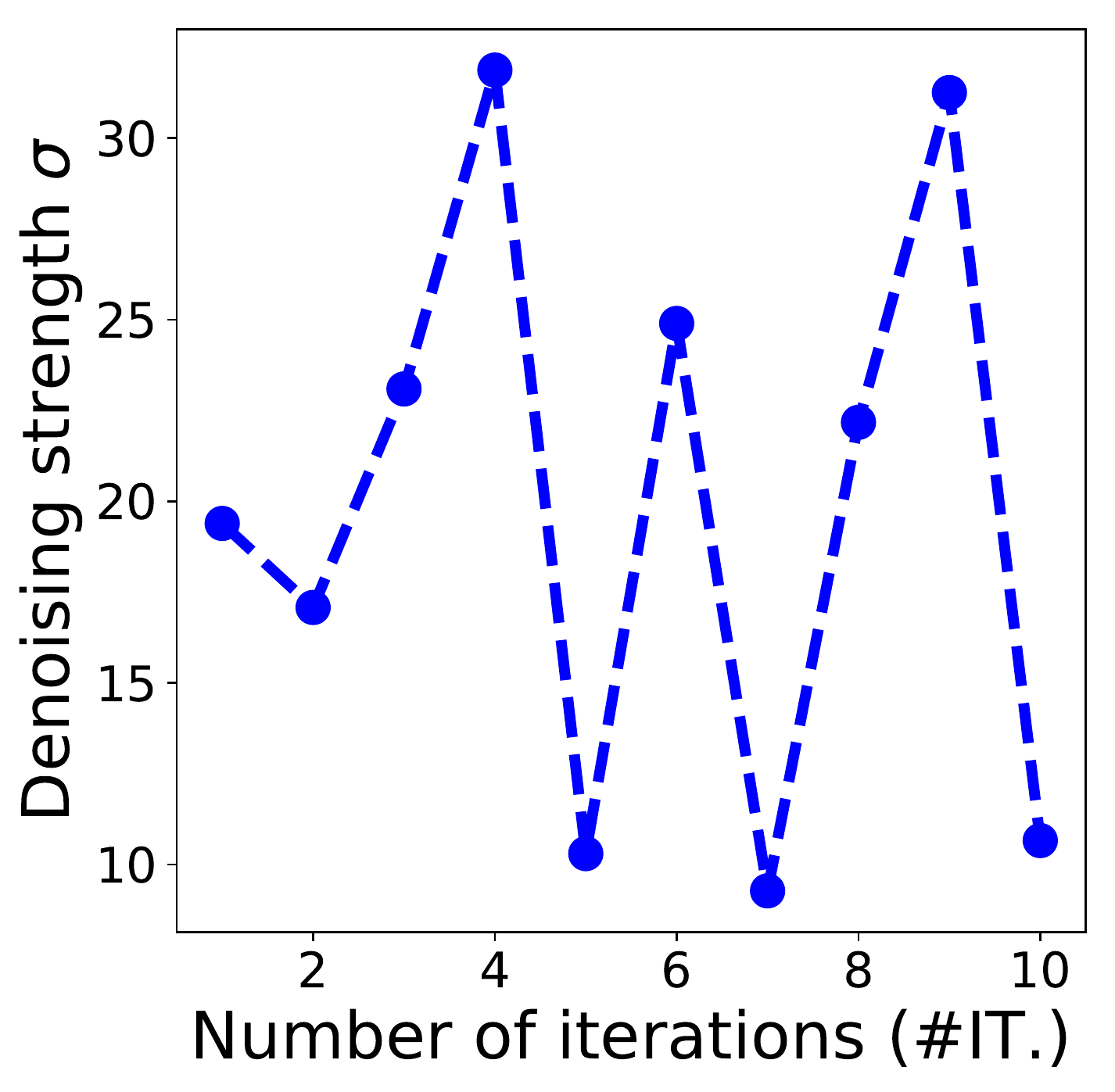}
	& \includegraphics[width=0.134\linewidth]{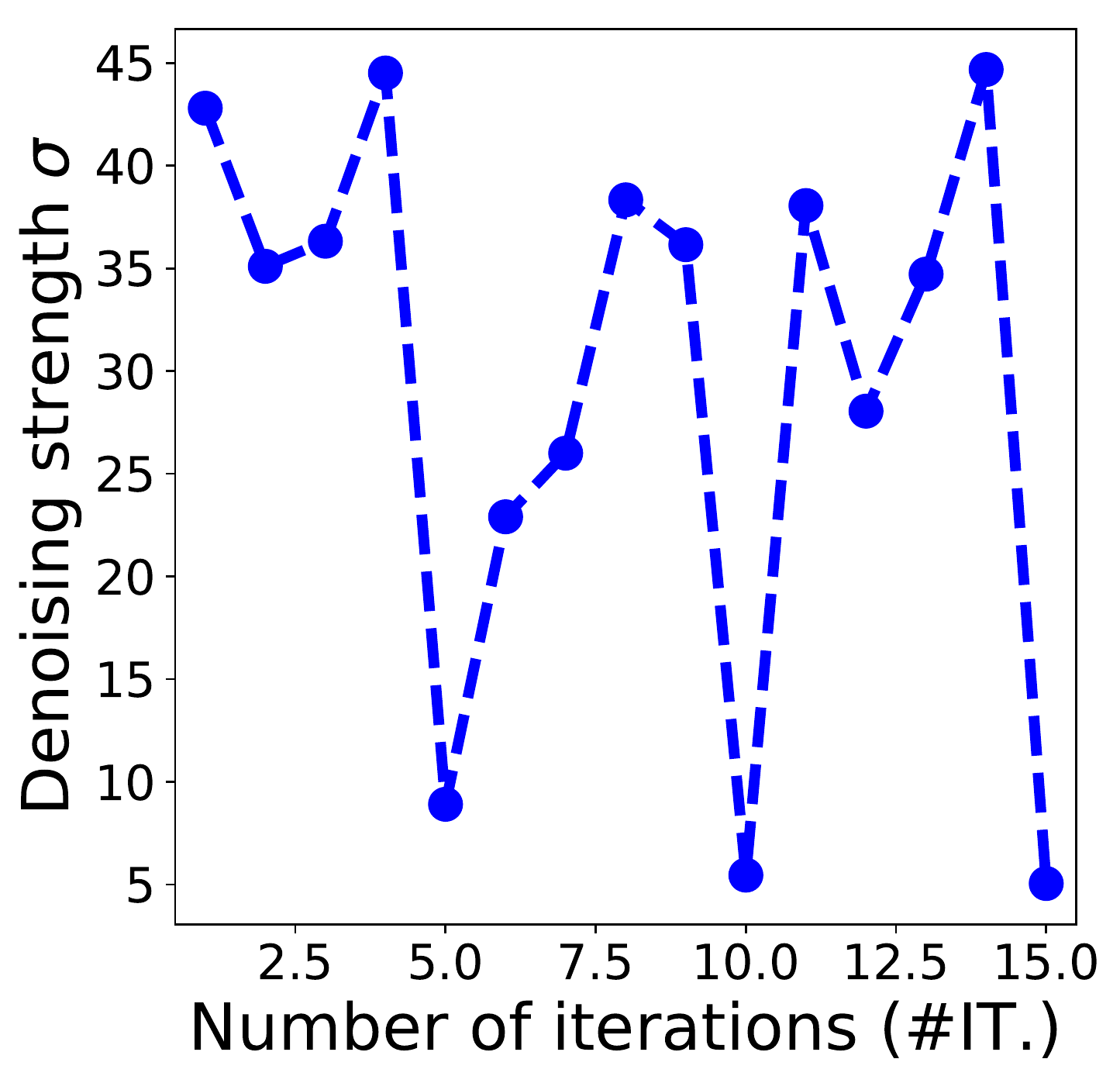}
	& \includegraphics[width=0.134\linewidth]{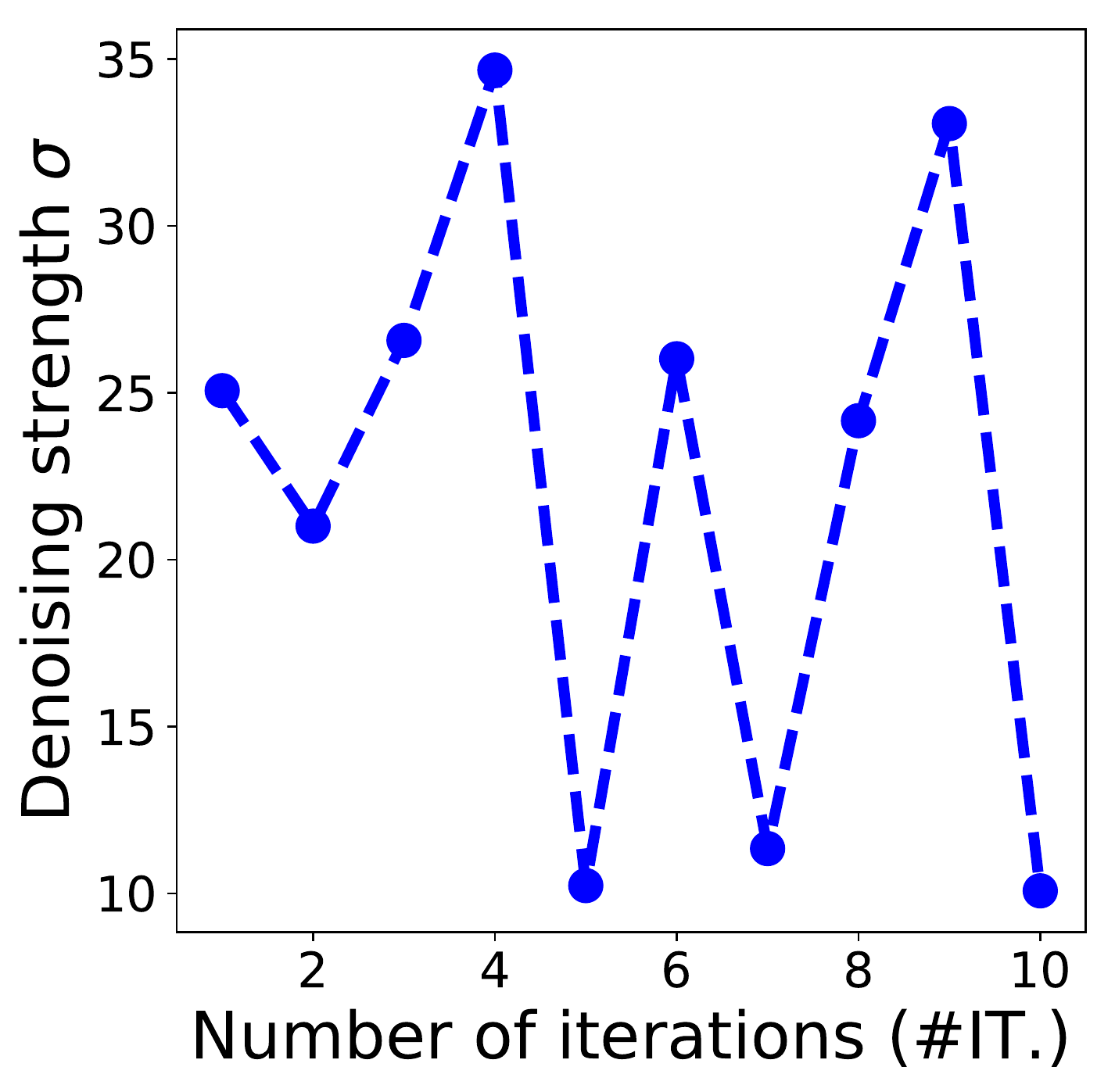}
	& \includegraphics[width=0.134\linewidth]{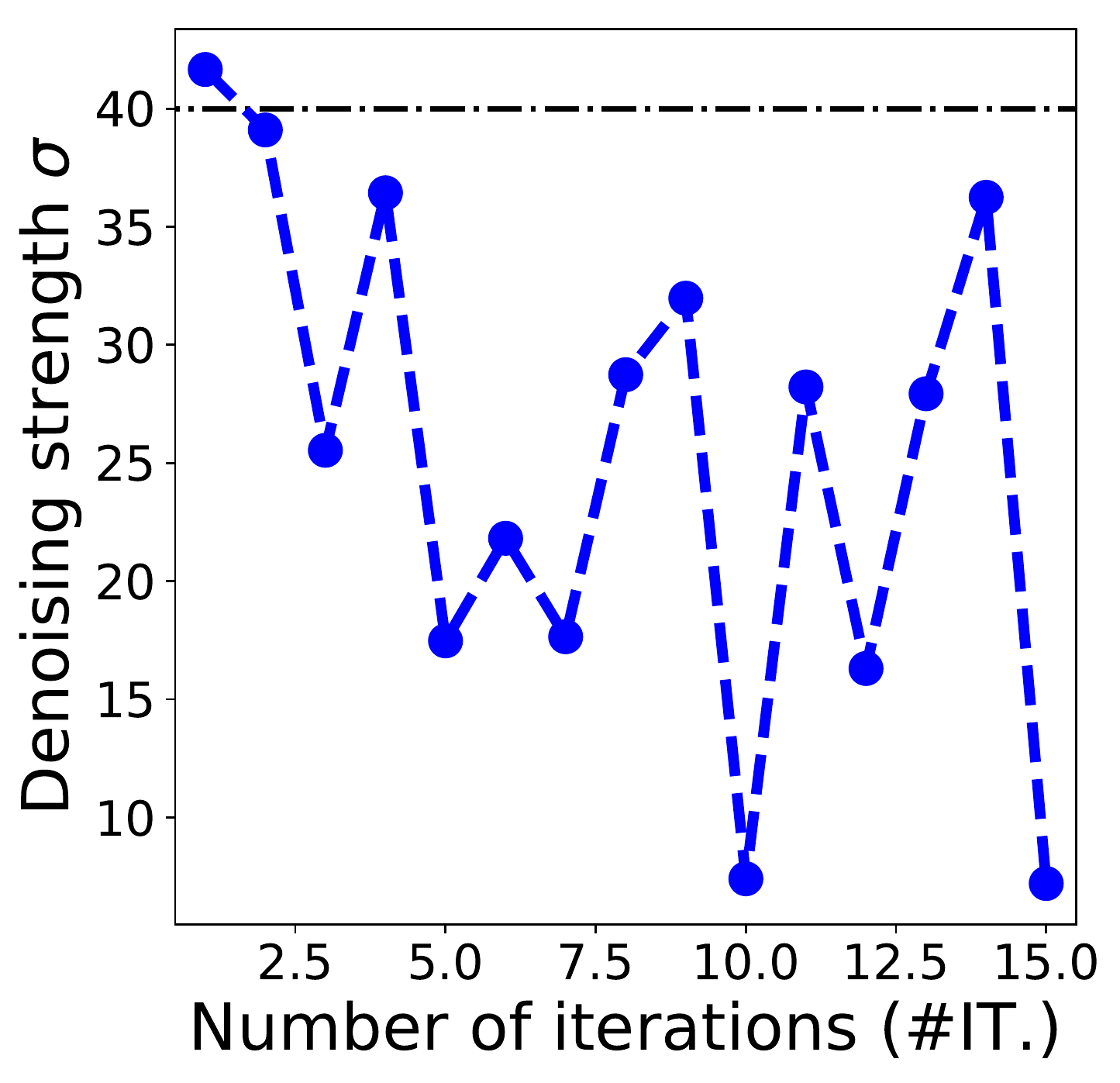} \\
	\includegraphics[width=0.134\linewidth]{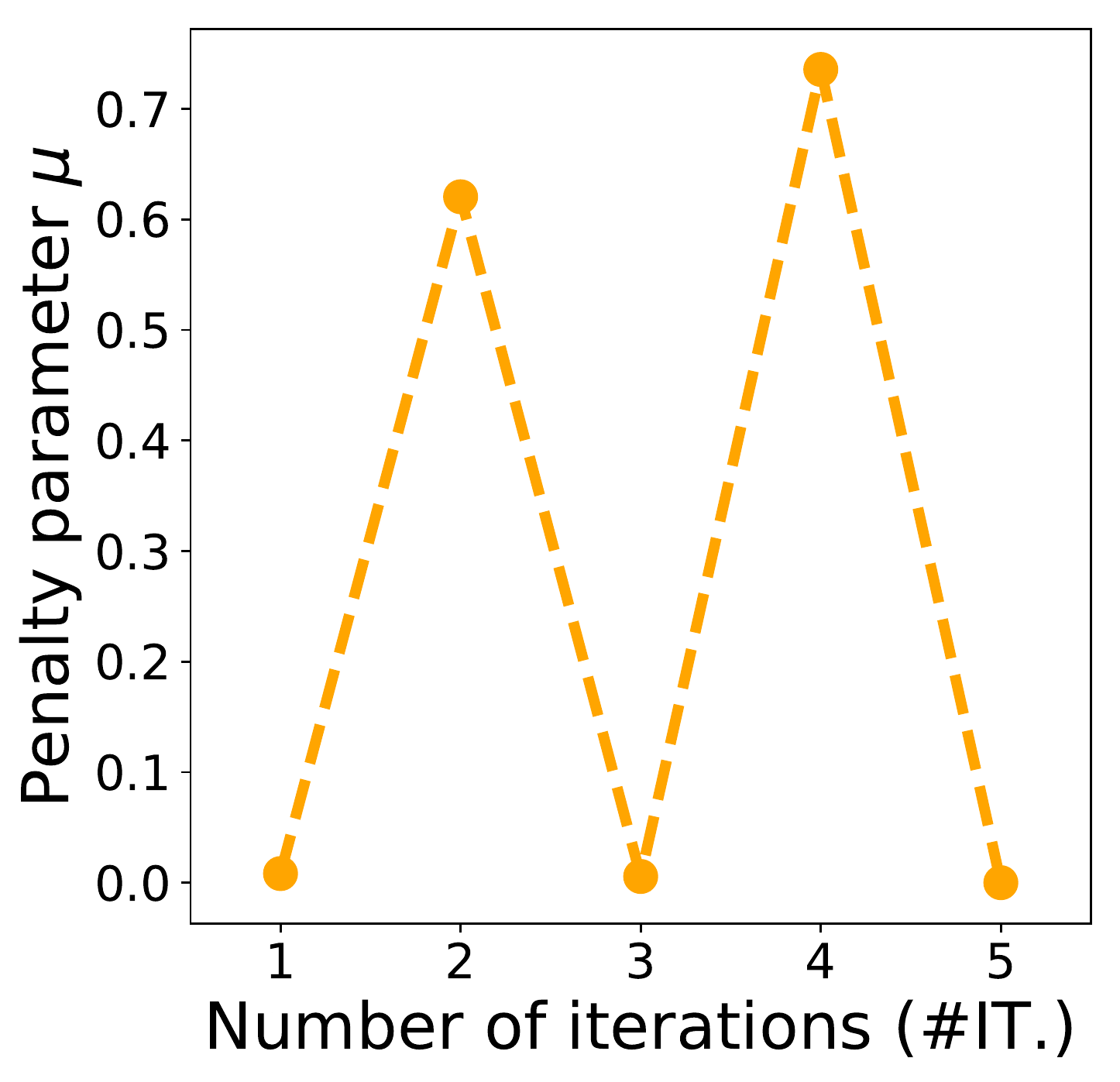}
	& \includegraphics[width=0.134\linewidth]{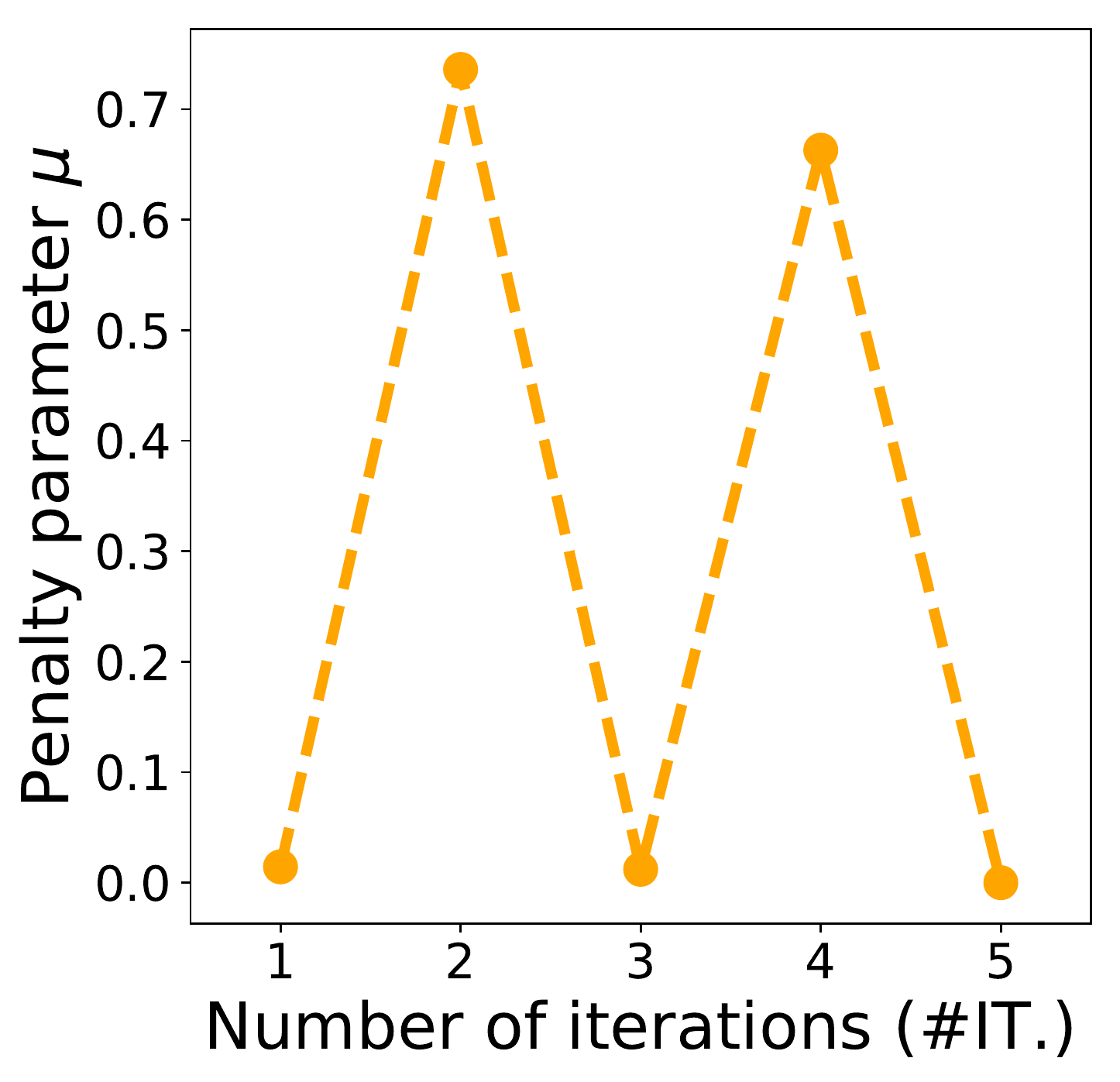}
	& \includegraphics[width=0.134\linewidth]{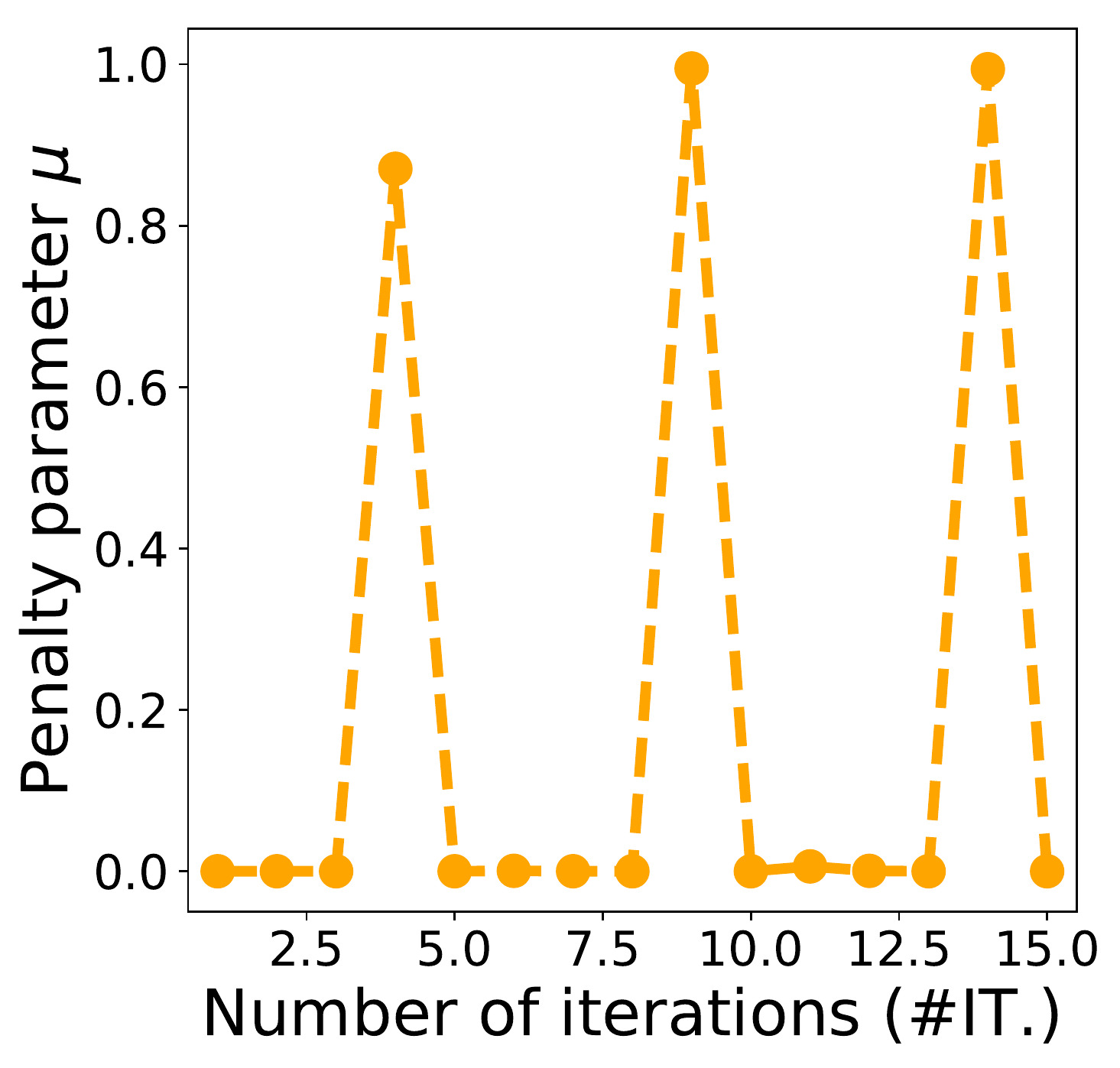}
	& \includegraphics[width=0.134\linewidth]{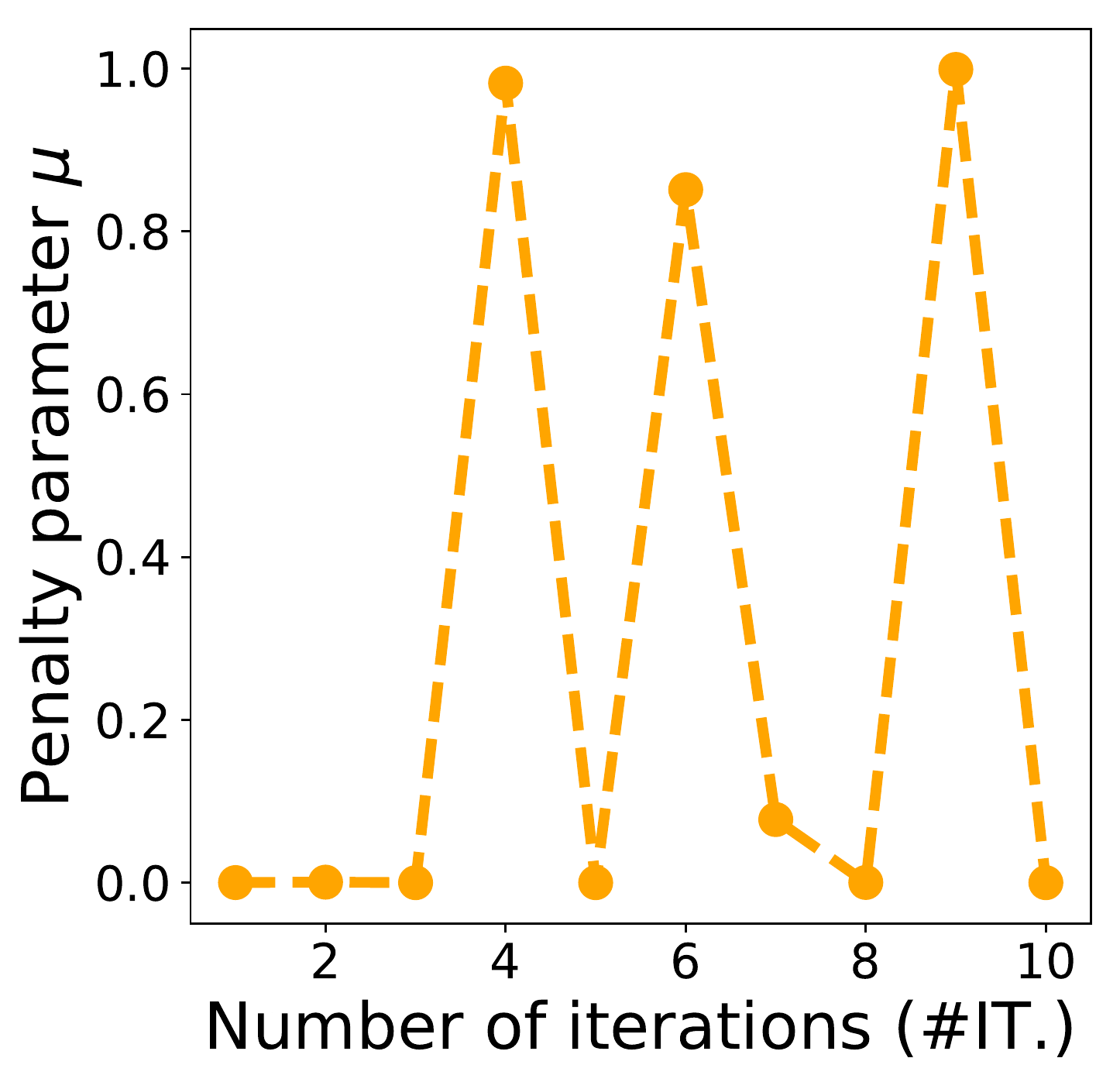}
	& \includegraphics[width=0.134\linewidth]{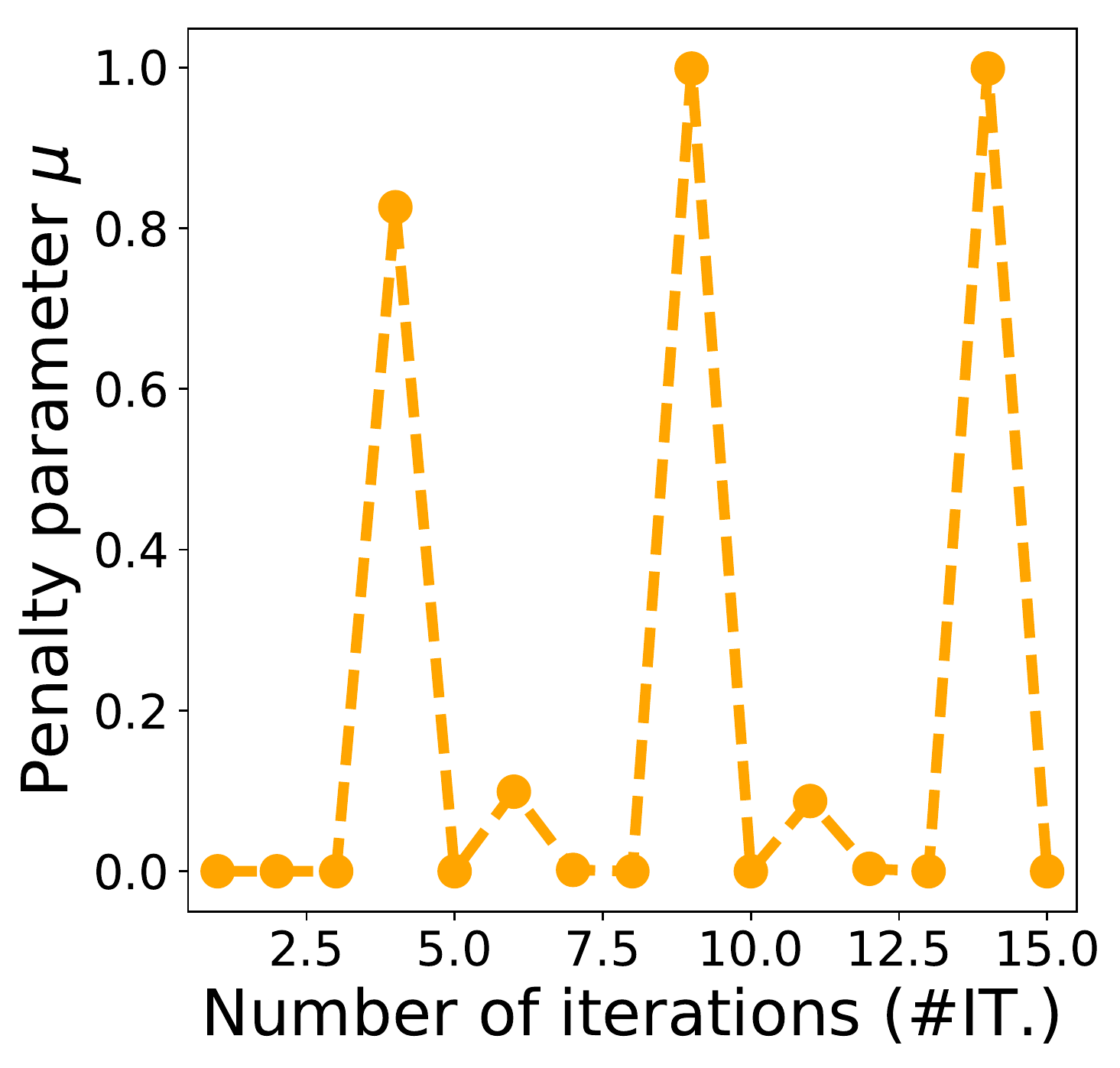}
	& \includegraphics[width=0.134\linewidth]{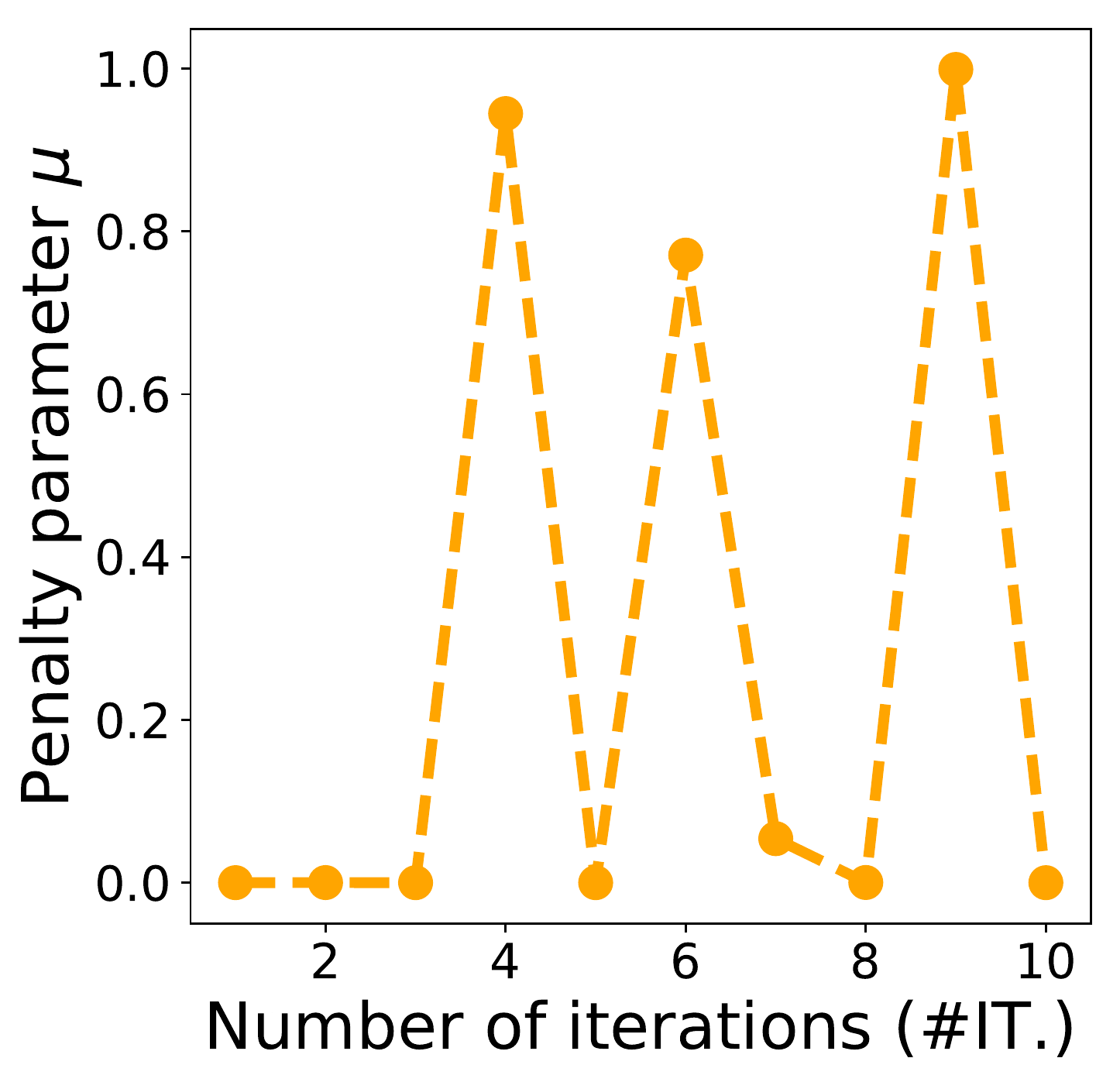}
	& \includegraphics[width=0.134\linewidth]{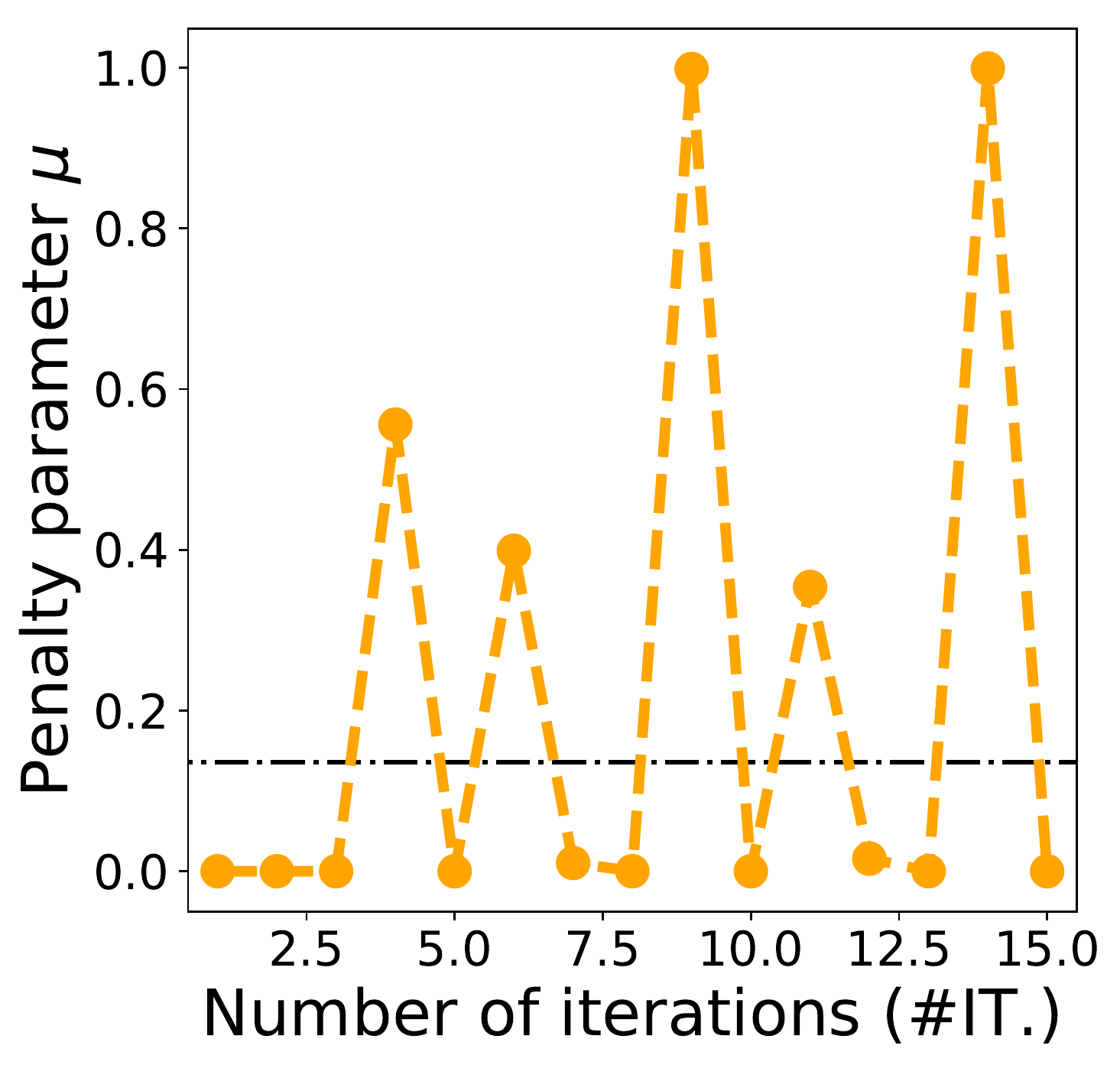} \\
	\end{tabular} 
	\caption{Behavior curves of our learned policy for the application of CS-MRI reconstruction on medical images. The first row displays the initial estimate $x_0$ of the underlying image. The second row shows the predicted denoising strength, $\sigma$, computed by our learned policy for each corresponding image. The third row displays the predicted penalty parameters $\mu$ vs the number of iterations.}
	\label{fig:policy-behavior}
\end{figure}

\subsection{Behaviors of Learned Policy}
To visualize the behavior of our learned policy towards different images, 
we illustrate its generated parameters ($\sigma_k$ and $\mu_k$) along iterations in Figure~\ref{fig:policy-behavior}.
It can be observed that our learned policy customizes different internal parameters for each different image, whilst the generated parameters change adaptively across iterations of the optimization process. 
The adaptive penalization has been justified to result in faster convergence in the convex setting both empirically and theoretically \citep{xu2017adaptive,xu2017admm,xu2017adaptive_relaxed,NIPS2015_5723,he2000alternating}. 
We conjecture similar result holds for the nonconvex setting as well, where the underlying adaptive parameter scheme 
can be automatically mined by our algorithm from the data, leading to a more efficient and effective policy than the oracle one. 

\begin{table}[!t]
\centering
\footnotesize
\begin{tabular}{lcccccccccc} 
		\toprule
		 &  \multicolumn{2}{c}{PnP-PGM} & \multicolumn{2}{c}{PnP-APGM} & \multicolumn{2}{c}{PnP-HQS} & \multicolumn{2}{c}{RED-ADMM} &
		 \multicolumn{2}{c}{PnP-ADMM} 
		 \\ \cline{2-11} 
		\textsc{Policies} & PSNR & \#IT.  & PSNR & \#IT.  & PSNR & \#IT.  & PSNR & \#IT.  & PSNR & \#IT. \\  \midrule
		fixed  & \underline{24.98} & 30.0 & 24.39 & 30.0 & 24.82 & 30.0 & 24.57 & 30.0 & 22.78 & 30.0  \\
		fixed$^{*}$  & 24.99 & 20.9 & \underline{25.07} & 13.6 & 24.83 & 20.6 & 24.58 & 27.3 & 24.19 & 7.3  \\
		fixed optimal & 25.66 & 30.0 & 24.73 & 30.0 & 25.27 & 30.0  & 25.14  & 30.0 & \underline{26.08} &  30.0  \\
		fixed optimal$^{*}$ & 25.68 & 21.1 & 25.69 & 11.4 & 25.35 & 27.6 & 25.15 & 29.1 & \underline{26.16} & 16.7  \\
		greedy & 24.83 & 30.0 & 24.46 & 30.0 & 24.34 & 30.0  & 24.95  & 30.0 & \underline{26.20} &  30.0  \\
		greedy$^{*}$ & 25.21 & 15.8 & 25.04 & 9.3 & 24.35 & 22.9 & 25.44 & 19.0 & \underline{26.33} & 19.0  \\		
		oracle & 25.85 & 30.0 & 25.21 & 30.0 & 25.50 & 30.0 & 25.36 & 30.0 & \underline{26.41} & 30.0  \\
		oracle$^{*}$ & \textcolor{blue}{25.90} & 21.4 & \color{blue}{25.87} & 11.9 &  \color{blue}{25.55} & 24.6  & \color{blue}{25.45} & 27.8 & \color{blue}{\underline{26.45}} & 21.6  \\ \midrule
	    Ours  & \color{orange}{26.28} & 20.7 &  \textcolor{orange}{26.36} & 10.0 & \textcolor{orange}{25.96} & 22.2 & \textcolor{orange}{26.11} & 22.1 & \color{orange}{\underline{26.52}} & 15.0   \\
		\bottomrule
\end{tabular}
\caption{Comparisons of different PnP-type algorithms using different policies for CS-MRI reconstruction (* denotes the policies with optimal early stopping). 
The maximum number of iteration is set as 30 for policies without early stopping.  
The results are computed on seven medical images using an acceleration factor of 8$\times$ and a noise level of 15. The numerical values reflect the PSNR and number of iterations  (\#IT.).  
The best and the second best results among policies within the same category of the PnP-type algorithm (the best results in each column) are displayed in \textcolor{orange}{orange} and \textcolor{blue}{blue} colors respectively, and the best results among all PnP-type algorithms using the same policy (the best result in each row) are \underline{underlined}. 
}
\label{tb:proximal-comparison}
\end{table}

\begin{table}[!t]
\centering
{
\begin{tabular}{lcccccc} 
		\toprule
		 & \multicolumn{2}{c}{$2\times$} & \multicolumn{2}{c}{$4\times$} &  \multicolumn{2}{c}{$8\times$} \\ \cline{2-7} 
		D-AMP & PSNR & \#IT.  & PSNR & \#IT.  & PSNR & \#IT. \\  \midrule
		original & 29.49 & 30.0 & 28.12 & 30.0 & 26.02 & 30.0 \\
		original$^{*}$ & 29.51 & 6.3 & 28.13 & 9.7 & 26.09 & 17.1 \\
		Ours & 30.31 & 5.0 & 28.55 & 10.7 & 26.49 & 14.3 \\
		\bottomrule
\end{tabular}}
    \caption{Numerical comparison (PSNR / \#IT.) of original D-AMP (* denotes optimal early stopping) and D-AMP with our learned policy for CS-MRI under various acceleration factors ($2\times$,$4\times$,$8\times$) with noise level 15.}
    \label{tb:D-AMP}
\end{table}

\section{Extensions to Other PnP-type Algorithms} \label{sec:extension}
In previous sections, we have provided extensive experiments to demonstrate the efficiency of our proposed algorithmic approach, instantiated by PnP-ADMM, on several inverse imaging problems. In this section, we show that the advantages of our approach hold for other PnP-type\footnote{We use the term ``PnP-type" to refer to all PnP techniques including PnP,  RED, and D-AMP algorithms.} algorithms as well.
We remark that the extensions of our approach to other proximal algorithms are straightforward and flexible, \ie one only needs to adjust the desired outputs of the policy network (the number of internal parameters) to match the corresponding PnP algorithms. 
In what follows, we detail the extensions on five representative PnP-type algorithms---PnP-PGM/APGM \citep{sun2019online}, PnP-HQS \citep{Zhang_2017_CVPR}, RED-ADMM \citep{romano2017little},  D-AMP \citep{metzler2016denoising},
and comprehensively study their performance on the CS-MRI application.

\subsection{Other PnP-type Algorithms}

\paragraph{PnP-PGM/APGM}  We first introduce the formulation of PnP proximal gradient method (PGM) and its accelerated variant (APGM).
Both of them can be expressed in an unified manner as follows:
\begin{equation} \label{eq:pnp-apg}
  \begin{split}
&z_{k+1}  =  s_k - \gamma_k \nabla \mathcal{D} ( s_k ),   \\
&x_{k+1} = \mathrm{Prox}_{\sigma_k^2 \mathcal{R}} ( z_{k+1} ) =  \mathcal{H}_{\sigma_k} ( z_{k+1} ), \\
&s_{k+1} = x_{k+1} + \frac{q_{k} - 1}{ q_{k+1} } ( x_{k+1} - x_k ),
  \end{split}
\end{equation}
where $\gamma_k$ denotes the step size for the gradient descent of the data-fidelity term. $q_k$ is an intermediate parameter for computing inertial parameter. 
If $q_k=1$ and its update  follows:
$$ q_{k+1} = \frac{1}{2} \big( 1 + \sqrt{1+4q_{k}^2} \big),  $$
 the algorithm \eqref{eq:pnp-apg} corresponds to APGM (or FISTA) \citep{beck2009fast}. 
When $q_k \equiv 1$, then $s_k = x_k$ and \eqref{eq:pnp-apg} recovers the standard PGM.

The fundamental conceptual differences between PGM and ADMM lie at their treatment of the data-fidelity $\mathcal{D}$. To ensure 
 data consistency, PGM adopts a simple gradient descent with step size $\gamma_k$, while ADMM relies on the proximal operator $\mathrm{Prox}_{\frac{1}{\mu_k} \mathcal{D}}$ with penalty parameter $\mu_k$. This leads to the different forms of internal parameters\footnote{Note we do not set $\mu_k =1 / \gamma_k$ to unify the formulations, as the proximal operator can also be solved inexactly using gradient descent, which would introduce an additional step size parameter.}.  Therefore, to adapt our algorithm into PnP-PGM, we modify the policy network to determine the step size $\gamma_k$, in place of the penalty parameter $\mu_k$. 
 For PnP-APGM, we treat $\bar{q_k} := \frac{q_{k} - 1}{ q_{k+1} }$ as an additional parameter and learn to predict $\bar{q_k}$  using the policy network.

\paragraph{PnP-HQS} Next, we introduce the PnP half quadratic splitting (HQS) algorithm, formulated as: 
\begin{equation} \label{eq:pnp-hqs}
  \begin{split}
&x_{k+1} = \mathrm{Prox}_{\sigma_k^2 \mathcal{R}} \left( z_k  \right) = \mathcal{H}_{\sigma_k} \left( z_k \right), \\
&z_{k+1} = \mathrm{Prox}_{\frac{1}{\mu_k} \mathcal{D}} \left( x_{k+1}  \right), \\
  \end{split}
\end{equation}
Roughly speaking,  HQS  can be viewed as a simplified version of ADMM. The only difference is that HQS does not have the dual variable $u$ (or $u$ is set to zero constantly), and the internal parameters of PnP-HQS conform  with those of  PnP-ADMM. As a result, no modification is required to learn a policy customized for PnP-HQS. 

\paragraph{RED-ADMM} Then, we show our algorithm can be used in another PnP-type framework, namely regularization by denoising (RED) as well. Unlike the PnP with implicit prior characterized by the PnP denoiser, the RED formulates an explicit prior term involving an denoising engine:
\begin{equation} \label{eq:red-prior}
    \mathcal{R}\left( x \right) = \frac{1}{2} x^T \left( x - \mathcal{H}_{\sigma} \left(x\right)  \right),
\end{equation}
in which the denoising engine itself is applied on the candidate image $x$, and the penalty induced is proportional to the inner product between this image and its denoising residual, $x - \mathcal{H}_{\sigma}\left(x\right)$. Under some assumptions on $\mathcal{H} \left(\cdot \right)$, the gradient of the regularization term is manageable \citep{romano2017little}. 
As a result, the resulted objective function can be solved by proper numerical schemes, \eg gradient descent, ADMM or fixed-point strategy. 
For simplicity, we choose the inexact RED-ADMM with one fixed-point iteration as a proof-of-concept, which is given by
\begin{equation} \label{eq:red-admm}
  \begin{split}
&v_{k+1} = \mathcal{H}_{\sigma_k} \left( x_k  \right), \\
&x_{k+1} = \tfrac{1}{\mu_k + \lambda_k} \big( \lambda_k v_{k+1}  + \mu_k  ( z_k - u_k )  \big), \\
&z_{k+1} = \mathrm{Prox}_{\frac{1}{\mu_k} \mathcal{D}} \left(x_{k+1} + u_k \right), \\
&u_{k+1} = u_k + x_{k+1} - z_{k+1}.
  \end{split}
\end{equation}
Compared to PnP-ADMM, 
 we modify our policy network and incorporate $\lambda_k$ into our parameter prediction.

\paragraph{D-AMP} Finally, we present our algorithmic approach on the denoising-based approximate message passing (D-AMP) framework. 
Different from proximal algorithms developed in optimization community, 
the AMP algorithm is initially derived from  the loopy belief propagation in graphical models from Bayesian statistics community \citep{donoho2009message}. One form of AMP algorithm reads
\begin{equation} \label{eq:d-amp}
  \begin{split}
  &v_{k+1} =  x_k + \mathcal{A}^H z_k, \\
  &x_{k+1} = f \left( v_{k+1}  \right), \\
  &o_{k+1} = \frac{1}{M} z_{k} \mathop{\mathrm{Div}} \{ f \left( v_{k+1}  \right) \},  \\
  &z_{k+1} = y - \mathcal{A} x_{k+1} + o_{k+1}, 
    \end{split}
\end{equation}
where $\mathcal{A}^H: \mathbb{R}^M \to \mathbb{R}^N$ or $\mathbb{C}^M \to \mathbb{C}^N$ is the adjoint operator for the forward operator $\mathcal{A}$, and $\mathop{\mathrm{Div}} \{ f (v) \}$ denotes 
the divergence (\ie the summation of derivatives) of an operator $f$ with respect to its input $v$. 
In the original form of AMP, $f$ is a shrinkage/thresholding non-linearity, which can be replaced by a proper denoiser $ \mathcal{H}_{\sigma_k} $ demonstrated in D-AMP framework. For complicated denoisers without analytical form of divergence, 
the divergence is approximated by the Monte Carlo method \citep{ramani2008monte}. 
One distinguishing feature of AMP algorithms is the Onsager correction term $o_k$, which renders the effective noise define by $v_k - x$  approximately Gaussian, 
 and thus the noise level of effective noise at each iteration can be estimated via
\begin{equation} \label{eq:amp-sigma}
\hat{\sigma}_{k+1} = \frac{\|  z_k \|_2 }{\sqrt{M}},
\end{equation}
This quantity can be used to adaptively set the denoising strength $\sigma_k$ of the employed denoiser in an automated manner. 
However, the above property of AMP only holds for compressed sensing applications with i.i.d. sub-Gaussian measurement matrices. 
Even though it has been extended to handle larger classes of random matrices,
\citep{cakmak2014s,manoel2015swept,ma2017orthogonal,rangan2019vector, rangan2019convergence,millard2020approximate,metzler2021d}, 
the behavior of AMP under general $\mathcal{A}$ is largely unpredictable and may even diverge in many situations where the internal parameters are not properly chosen. 
Here, we show our automated parameter tuning scheme is also effective to tackle this dilemma
by predicting the denoising strength at D-AMP iteration using our learned policy. 
Instead of directly inferring $\sigma_k$,  we make full use of  the initial estimate of \eqref{eq:amp-sigma}\footnote{It turns out \eqref{eq:amp-sigma} yields over-estimated noise level for CS-MRI task, consequently we follow \citep{eksioglu2018denoising} to adopt $\hat{\sigma}_{k+1} = \frac{\|  z_k \|_2 }{\sqrt{N}}$ instead. 
} and introduce a multiplicative factor $\delta$ 
 into it, thereby the final $\sigma_k = \delta \hat{\sigma}_k$. The policy network is thus responsible for predicting $\delta$ from data. 


\paragraph{Result Analysis}
With these simple modifications at hand,  our algorithm can be readily applied to these PnP-type algorithms to learn their tailored policies.  Moreover, we also evaluate the other policies defined in Section~\ref{sec: policy-comparison}.  Table~\ref{tb:proximal-comparison} and Table~\ref{tb:D-AMP} summarize the numerical results of different PnP-type algorithms equipped with different policies, 
 from which we observed that our learned policy can produce extremely effective parameters for these PnP-type algorithms and consistently outperforms the oracle policies. 
Note that, PnP-ADMM is still the top-performing PnP-type algorithm, while PnP-APGM with our learned policy can reach a comparable performance using fewer iterations.
These results collectively show the adaptive parameter scheme (yielded by our learned policy) is of great benefit to PnP-type algorithms, 
with regard to both practical convergence and restoration accuracy.

\begin{figure}[!t]
	\centering
	\setlength\tabcolsep{1.5pt}
	\begin{tabular}{cccccc}		
	& PnP-PGM & PnP-APGM & PnP-HQS & PnP-ADMM\\
	\rotatebox[origin=c]{90}{\textsc{IT\#1}}
	& \includegraphics[align=c,width=0.23\linewidth]{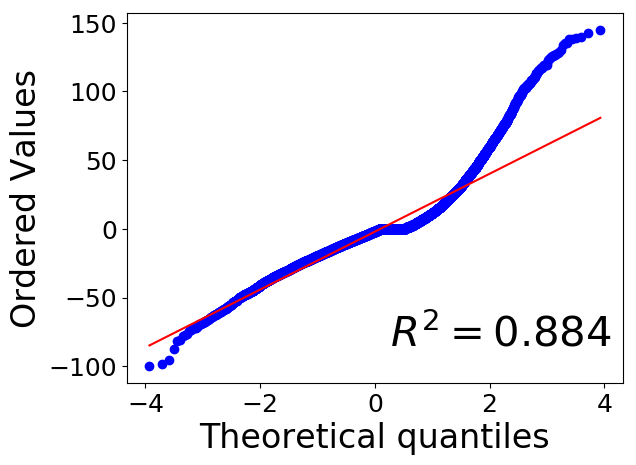}
	& \includegraphics[align=c,width=0.23\linewidth]{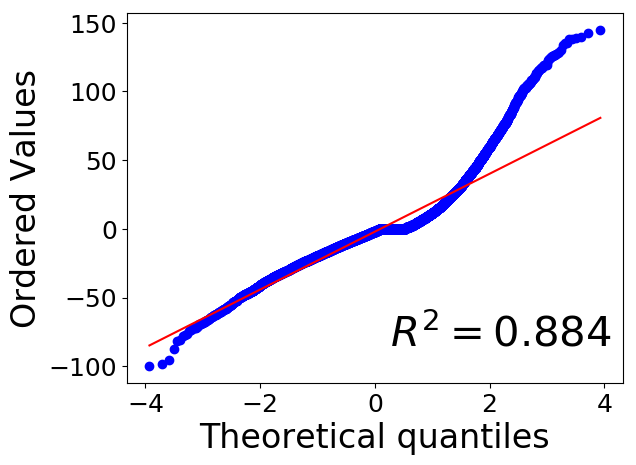}
	& \includegraphics[align=c,width=0.23\linewidth]{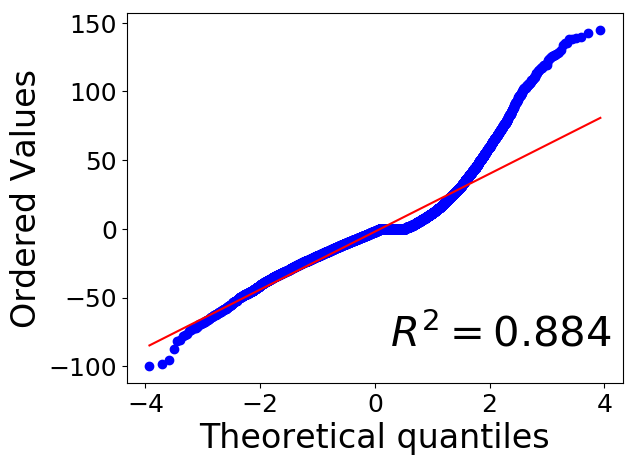}
	& \includegraphics[align=c,width=0.23\linewidth]{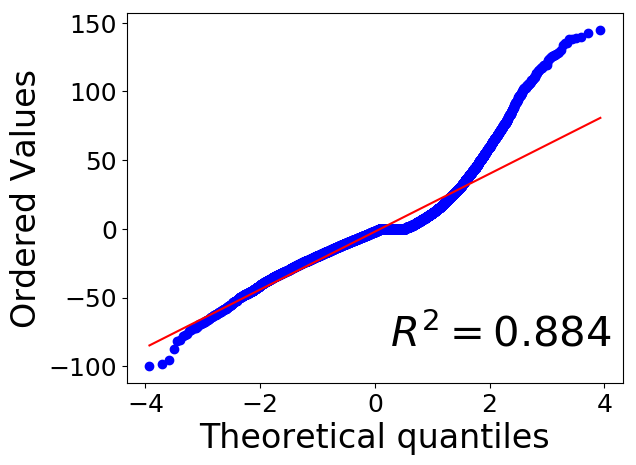}	\\	
	\rotatebox[origin=c]{90}{\textsc{IT\#2}}
	& \includegraphics[align=c,width=0.23\linewidth]{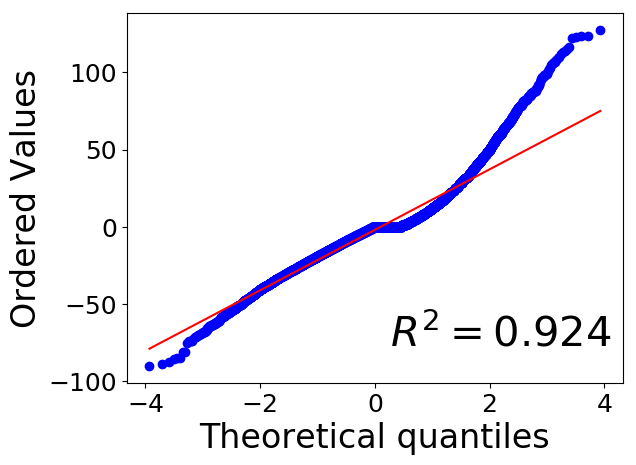}
	& \includegraphics[align=c,width=0.23\linewidth]{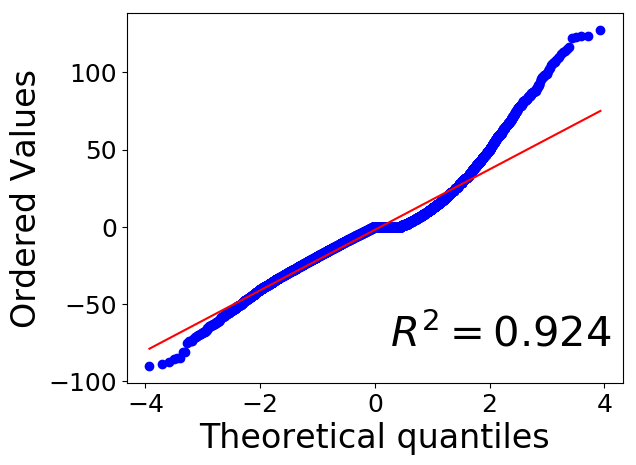}
	& \includegraphics[align=c,width=0.23\linewidth]{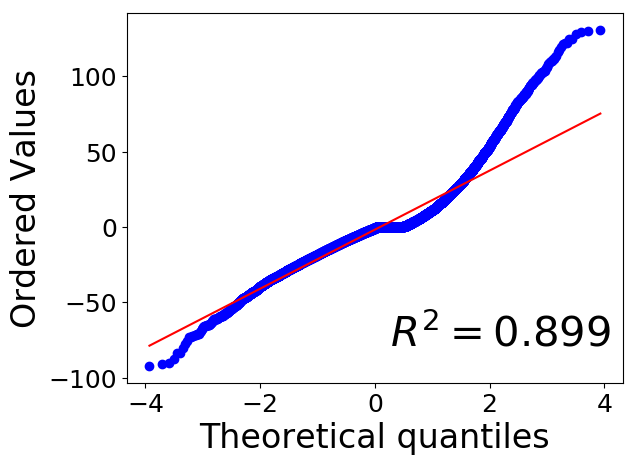}
	& \includegraphics[align=c,width=0.23\linewidth]{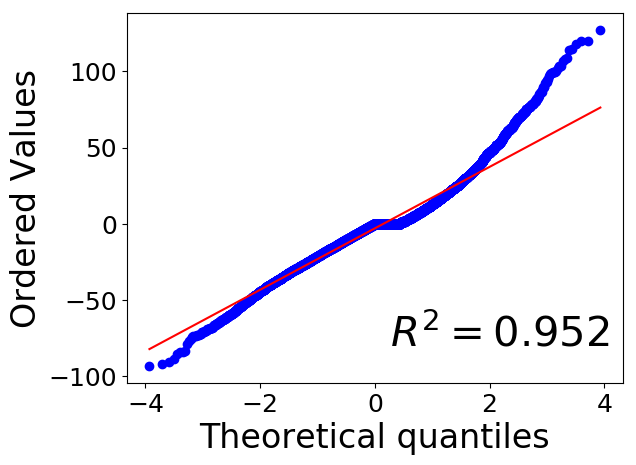}	\\
	\rotatebox[origin=c]{90}{\textsc{IT\#3}}
	& \includegraphics[align=c,width=0.23\linewidth]{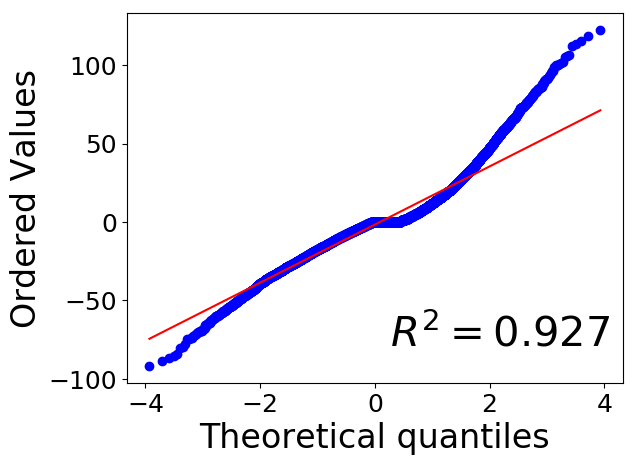}
	& \includegraphics[align=c,width=0.23\linewidth]{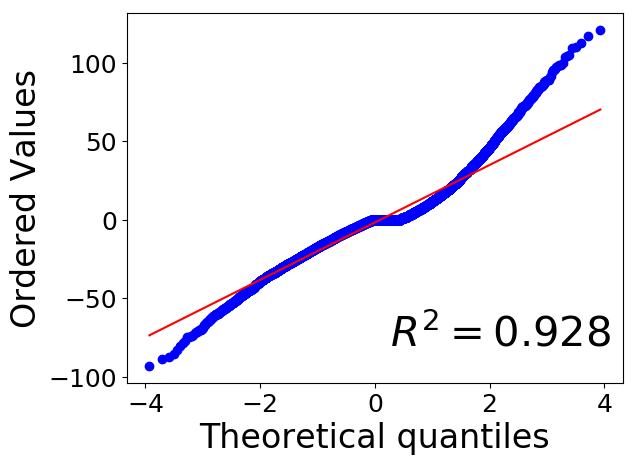}
	& \includegraphics[align=c,width=0.23\linewidth]{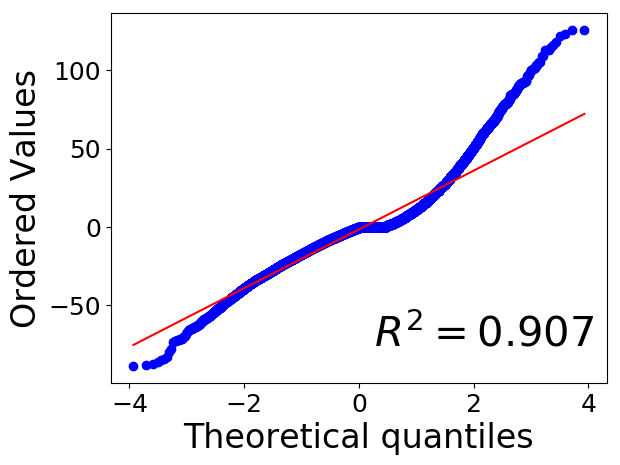}
	& \includegraphics[align=c,width=0.23\linewidth]{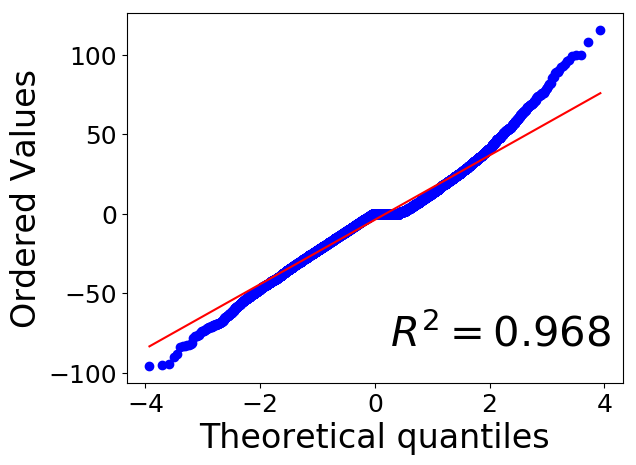}	\\
	\rotatebox[origin=c]{90}{\textsc{IT\#30}}
	& \includegraphics[align=c,width=0.23\linewidth]{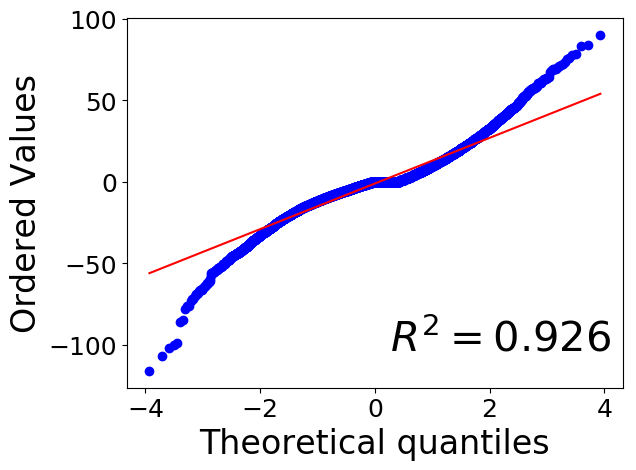}
	& \includegraphics[align=c,width=0.23\linewidth]{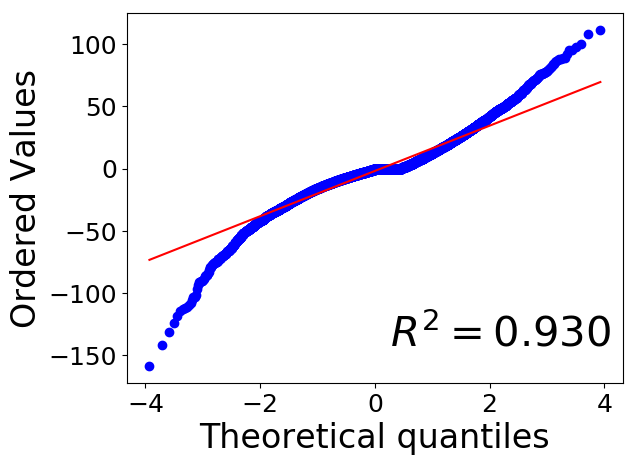}
	& \includegraphics[align=c,width=0.23\linewidth]{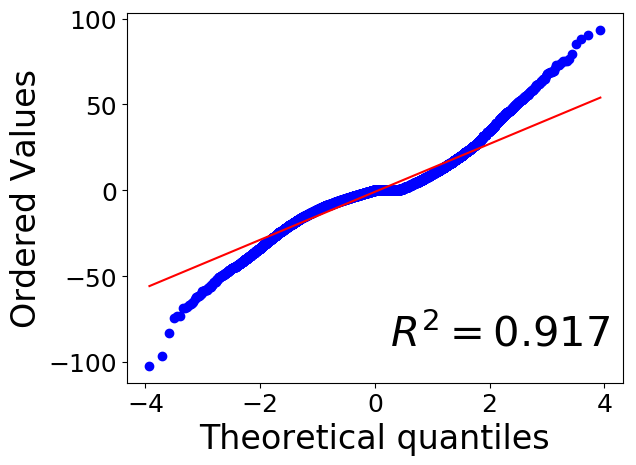}
	& \includegraphics[align=c,width=0.23\linewidth]{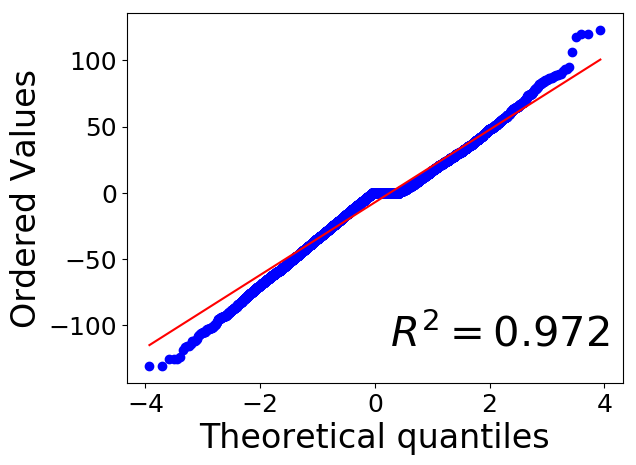}	\\
	\end{tabular} 
	\caption{Gaussian probability plots of iteration noise at four specific iterations (\#1, \#2, \#3 and \#30) for different PnP algorithms with oracle policies. The resulting image looks close to a straight line if the data are approximately Gaussian distributed. 
	A higher $R^2$ indicates a better fit.  
	}
	\label{fig:PnP-QQPlot}
\end{figure}

\begin{figure}[!t]
	\centering
	\setlength\tabcolsep{1.5pt}
	\begin{tabular}{ccc}	
		\rotatebox[origin=c]{90}{\textsc{Fixed Policy}} &
		\includegraphics[align=c,width=0.45\linewidth]{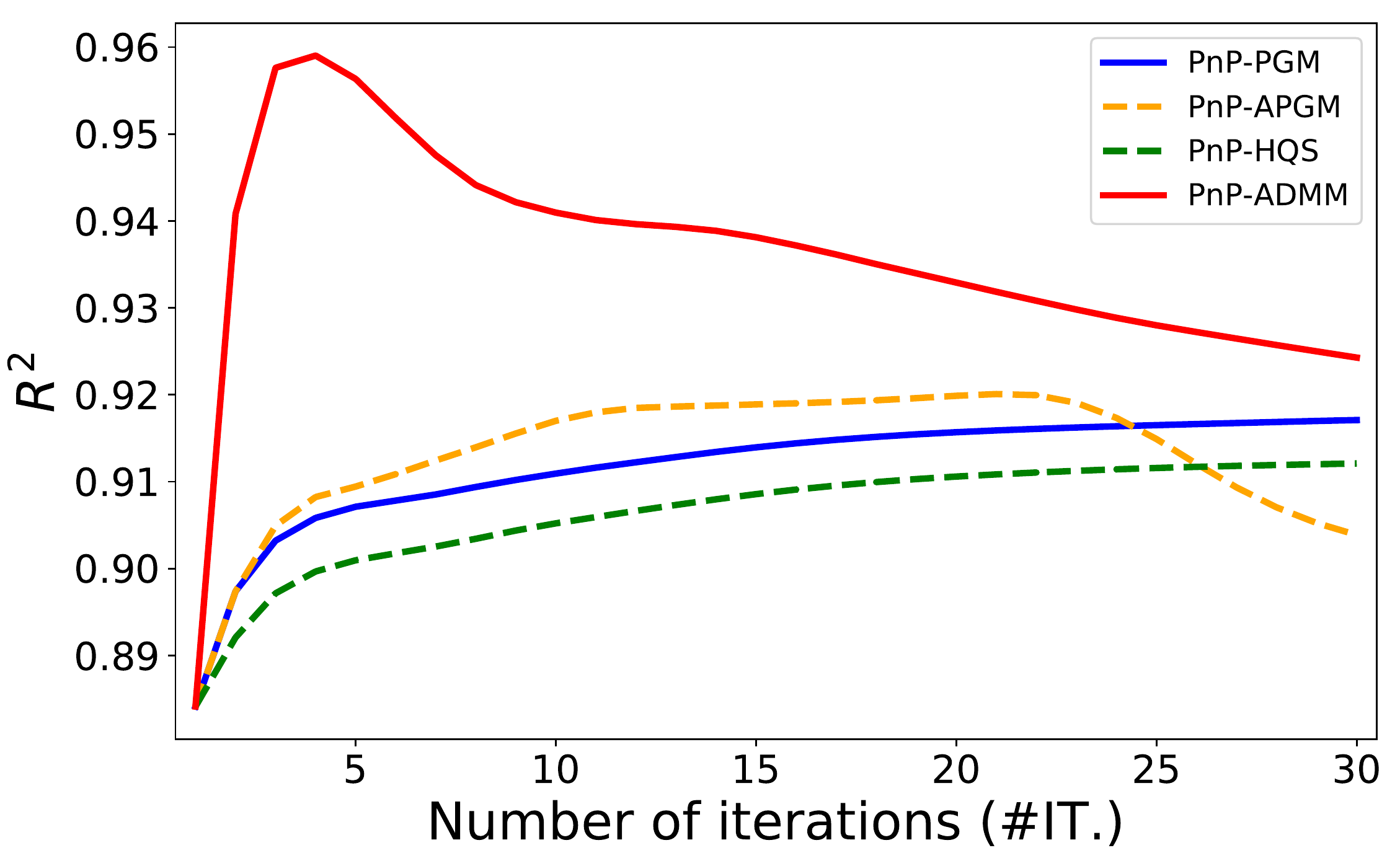} &
		\includegraphics[align=c,width=0.45\linewidth]{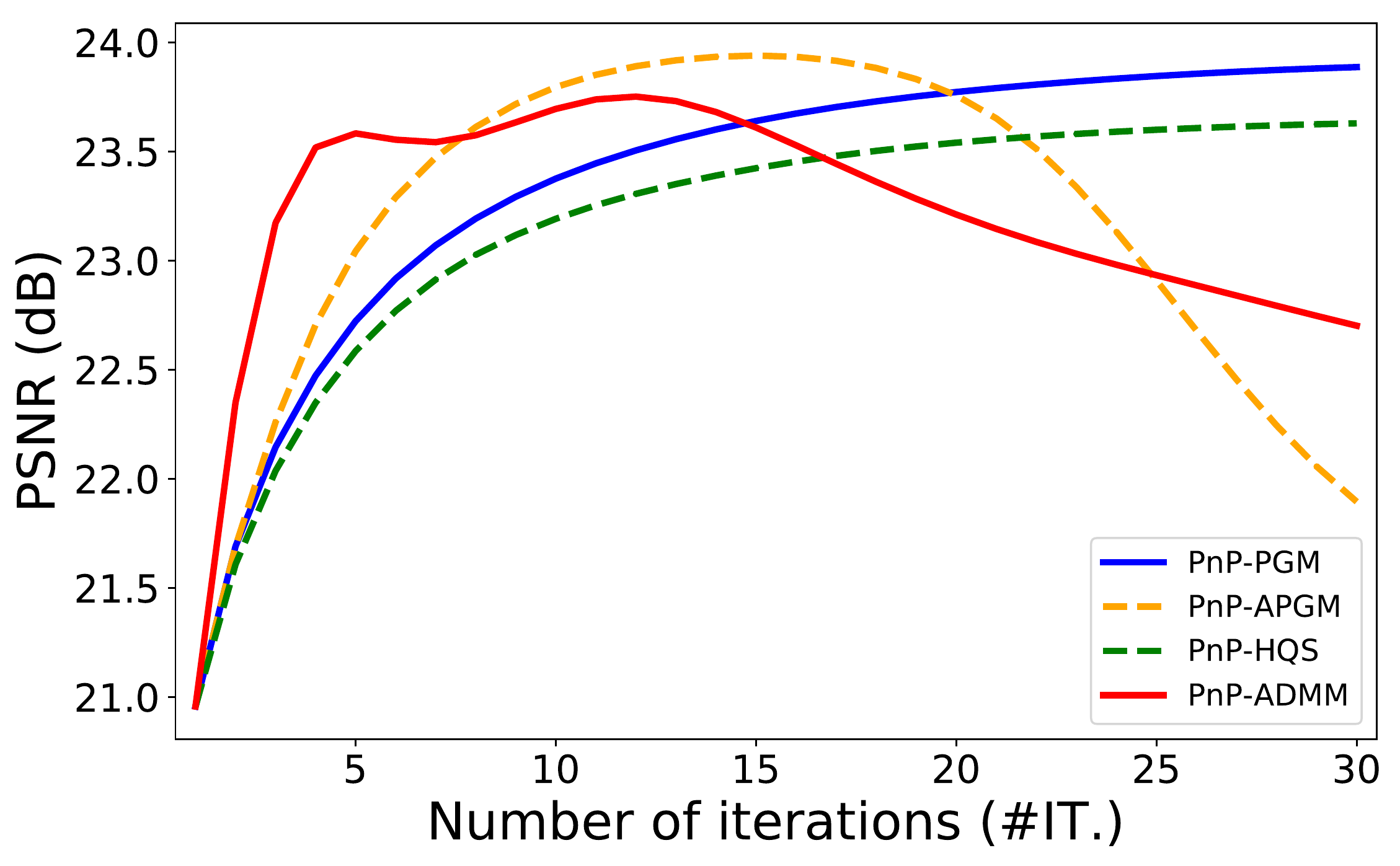} \\

		\rotatebox[origin=c]{90}{\textsc{Oracle Policy}} &
		\includegraphics[align=c,width=0.45\linewidth]{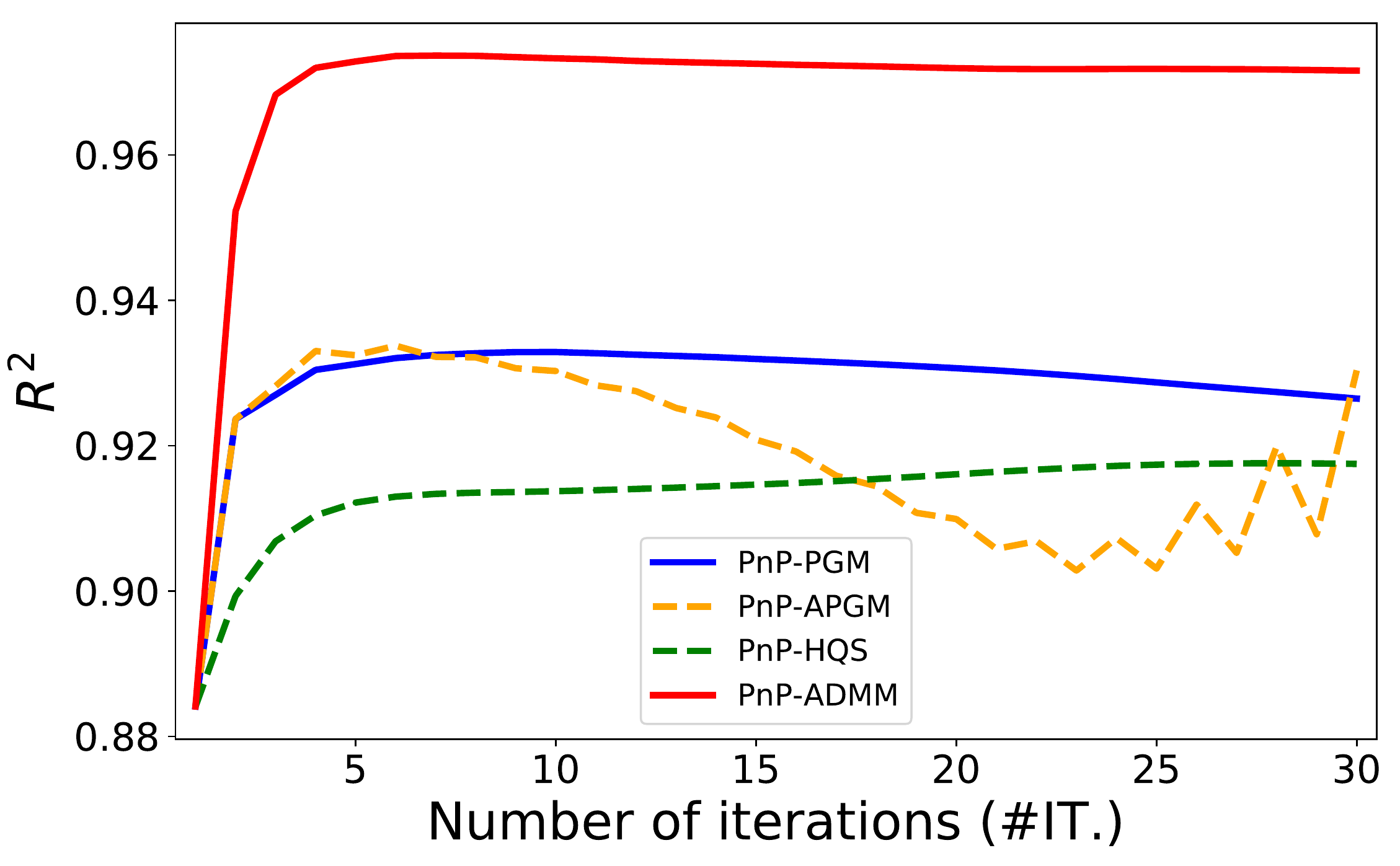} &
		\includegraphics[align=c,width=0.45\linewidth]{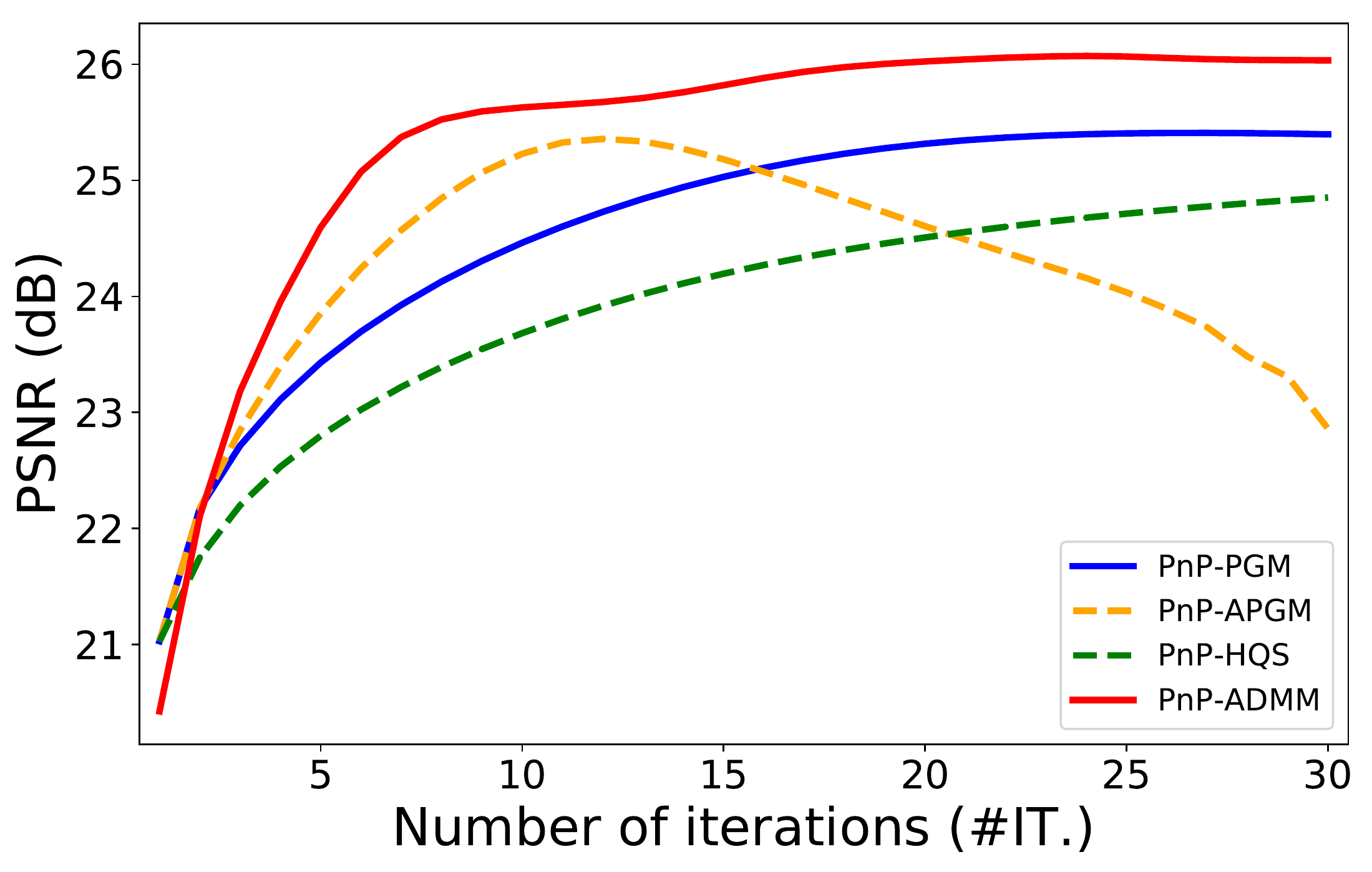} \\

		\rotatebox[origin=c]{90}{\textsc{Learned Policy}} &
		\includegraphics[align=c,width=0.45\linewidth]{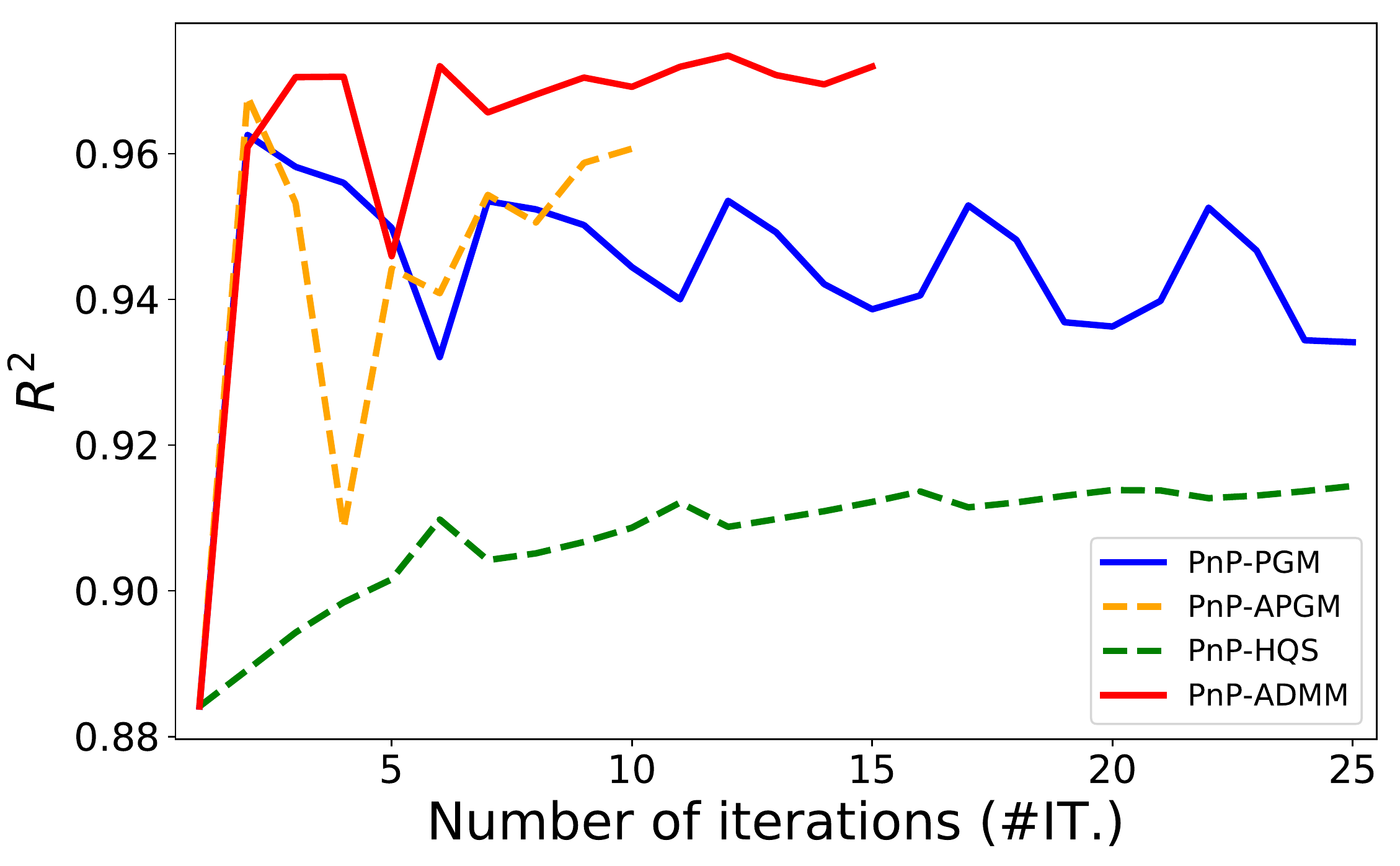} &
		\includegraphics[align=c,width=0.45\linewidth]{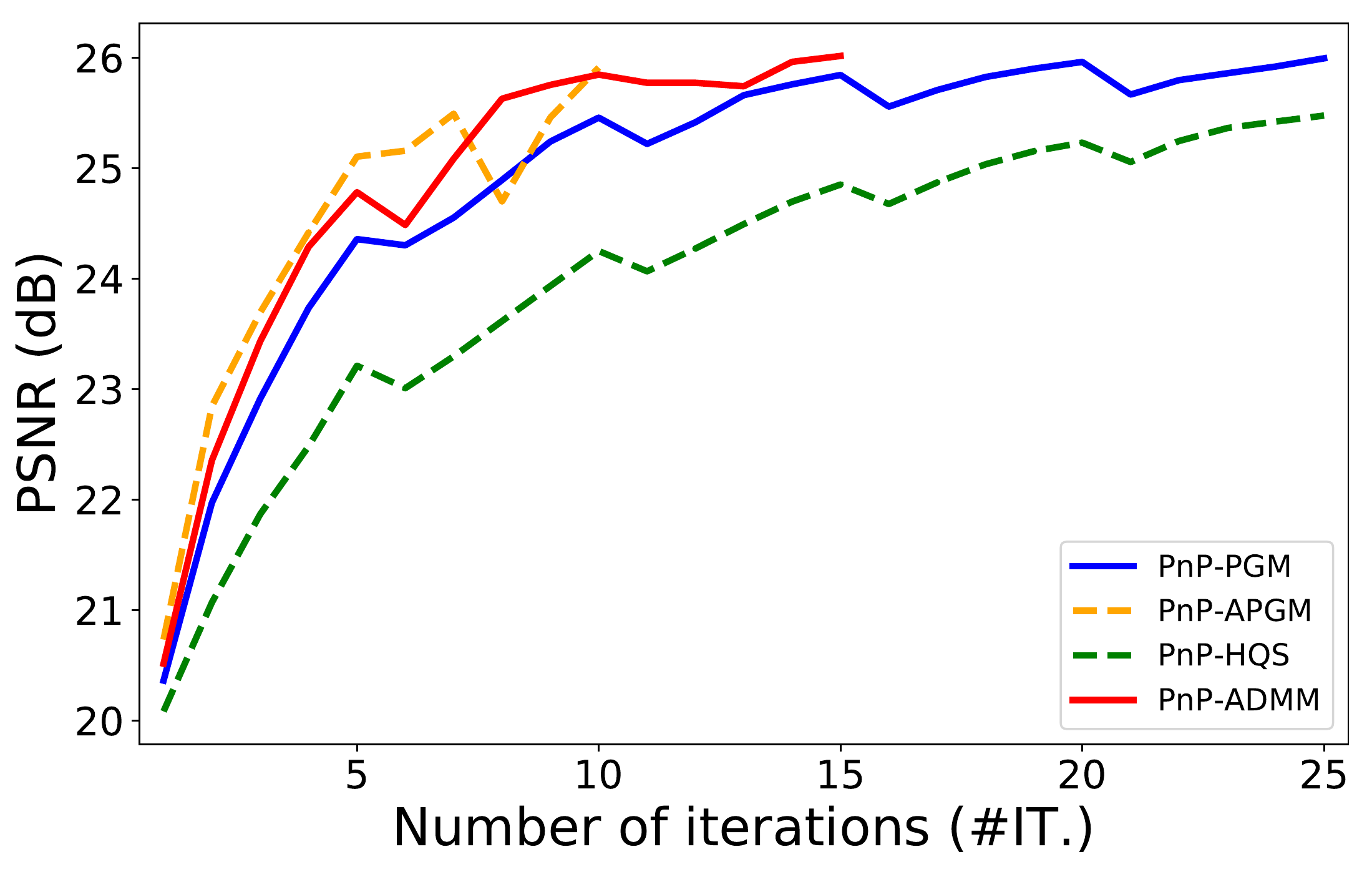} \\				

	\end{tabular} 
	\caption{$R^2$ curve (Left)  and PSNR curve (Right) across iterations for four PnP algorithms using three different policies. }
	\label{fig:PnP-R2}
\end{figure}

\subsection{Iteration Noise of PnP Methods}

The promising results reported in Table~\ref{tb:proximal-comparison} push us to further think of a puzzling question---\textit{Why are there distinctions among different PnP algorithms regarding recovery performance, especially given the fact that these PnP algorithms share the same set of fixed points?}\footnote{The rigorous proof can be found in Remark 3.1 of \citep{Meinhardt_2017_ICCV} and Proposition 3 of \citep{sun2019online}}
Directly answering this question could be rather difficult, but fortunately,
 we find some clues that could be potentially useful to analyze the performance of different PnP methods. 
 To do this, we leverage a concept called \textit{iteration noise}, which can characterize the statistical property of PnP algorithms and shed light on the explanation of the aforementioned question. 

For all PnP algorithms, the key step is the denoising step using the PnP prior, which is
\begin{equation}
\label{eq:key_step}
x_{k+1} =  \mathrm{Prox}_{\sigma_k^2 \mathcal{R}} ( \bar{x}_{k} ) = \mathcal{H}_{\sigma_k} ( \bar{x}_{k} ),
\end{equation}
where $x_{k}$ is the recovered image by the algorithm at the $k$-th iteration, $\bar{x}_{k}$  is the intermediate variable whose form depends on the given PnP algorithm.
For instance,  we have $\bar{x}_{k} := z_k - u_k$ in PnP-ADMM. 

Typically, the PnP denoiser $\mathcal{H}_{\sigma_k}$ is designed or trained for the purpose of Gaussian denoising \citep{venkatakrishnan2013plug}, given the proximal operator $\mathrm{Prox}_{\sigma_k^2 \mathcal{R}}$  sets up a regularized Gaussian denoising problem.  
Such a denoiser is most desired and effective for removing Gaussian noise, but its performance would be severely degraded when tackling with other non-Gaussian noise \citep{plotz2017benchmarking,wei2020physics}.  
In this regard, investigating the \textit{noise} the PnP denoiser intends to suppress, serves as an important clue to analyze  the performance of PnP methods.  To facilitate further analysis, let $x$ be the underlying image that we want to recover, then we can rewrite  \eqref{eq:key_step} to
\begin{equation}\label{eq:key_step_2}
x_{k+1} = \mathcal{H}_{\sigma_k} ( x + \bar{x}_{k} - x) = \mathcal{H}_{\sigma_k} ( x + \epsilon_{k} ), 
\end{equation}
where  $\epsilon_{k} =   \bar{x}_{k} - x$ measures the discrepancy between the current estimate $ \bar{x}_{k} $ and the ground truth $x$ at $k$-th iteration.   
We term this quantity as the \textit{iteration noise} of the PnP algorithm in the $k$-th iteration. It should be acknowledged the similar quantity has been called by \textit{effective noise} in the context of D-AMP algorithm \citep{metzler2016denoising}, but here we elaborate more on its characteristics \textit{varying along iterations}, in the broader context of  general PnP framework. 

Ideally, the PnP denoiser should be designed/trained for removing the iteration noise $\varepsilon_{k}$. However, 
it is unclear what probability distributions $\varepsilon_{k}$ complies in practice, as it can vary across different inverse imaging applications, different proximal algorithms, as well as different internal parameter settings. 
Without any prior information available in advance,  we argue that the Gaussian denoiser is still the most universal choice for the PnP algorithm given the central limit theorem. 

Although theoretical justifications of the iteration noise are highly difficult,  
we can provide empirical analysis on it in the context of PnP methods with Gaussian denoiser. 
Intuitively, the performance of PnP methods relies on the effectiveness of the PnP denoiser to remove the iteration noise. 
The more the iteration noise resembles the Gaussian noise, the more likely it can be suppressed by the Gaussian denoiser. 
To test the normality of the iteration noise, we employ the Gaussian probability plot \citep{Wilk1968Probability} on the iteration noise  for different PnP algorithms\footnote{We do not include the results of RED-ADMM and D-AMP as their fixed points differ from PnP methods.} with oracle policies.  We also evaluate the goodness-of-fit by $R^2$---the coefficient of determination with respect to the resulting probability plot \citep{MORGAN201115}.  The obtained results are shown in Figure~\ref{fig:PnP-QQPlot}, 
from which we observe that: 1) for all PnP algorithms, their corresponding iteration noise shows heavy tails in the beginning stage of the iteration; 2) the degrees of heavy tails are alleviated along iterations for all algorithms 
and 3) PnP-ADMM shows the best improvement over the other three---it improves the $R^2$ to 0.972, whilst the best value from the other three is 0.930 from PnP-APGM.
This also conforms with the performance shown in Table~\ref{tb:proximal-comparison}, where PnP-ADMM performs favorably against others in the oracle policy setting (\eg 26.41 for PnP-ADMM, 25.85 for PnP-PGM). 

To conclude this discussion, 
 we display in Figure~\ref{fig:PnP-R2} that $R^2$ and PSNR vary over iterations for all PnP methods armed with three policies (\ie fixed policy, oracle policy, and learned policy).
It can be seen that compared to the oracle policy, the normality of iteration noise and the recovery performance of PnP-type algorithms (\eg PnP-APGM/PGM) can be significantly improved with our learned policy. It also implies a positive correlation between $R^2$ and PSNR values, suggesting the benefit of Gaussianizing iteration noise.  
We note the importance of Gaussianizing iteration noise has been demonstrated from the AMP literature. In fact, much of the success of AMP is directly attributed to the Gaussian distribution of iteration noise \citep{donoho2009message}. Here, we generalize this conclusion beyond the scope of AMP, to the wide spectrum of PnP methods.

\section{Conclusion and Future Works} \label{sec:conclusion}
In this work, we introduced reinforcement learning (RL) into the PnP framework, yielding a novel tuning-free PnP proximal algorithm for a wide range of inverse imaging problems. 
The main strength of our approach is the policy network, which can customize parameters for different images. 
Through numerical experiments, we demonstrate our learned policy often generates highly effective parameters, which often even reaches comparable performance to the ``oracle" parameters tuned towards the ground truth. 

Our approach takes a big step towards solving the issues of tuning and slow-convergence, known weaknesses of PnP-type methods.  
There is, however, still room left for improvement. 
For example, our proposed algorithm relies on a mixture of a model-free and a model-based RL algorithm for learning the policy network. We believe that recently proposed more advanced RL algorithms can be adopted to our framework and further boost the performance of the learned policy.  Moreover, the  current version of our algorithm can only scale to moderately-sized imaging problems. Making it scalable to large-scale settings is a non-trivial task left for future research. Last but not least, our empirical analysis of the iteration noise reveals some hidden structures of PnP proximal algorithms. Seeking a deep understanding of what makes these algorithms stand out in terms of performance is an interesting and challenging question. 
We foresee a rich pool of future research to further tackle these challenges, aiming  at efficient, effective, and versatile image recovery for inverse imaging applications. 

\acks{This research was supported by the National Natural Science Foundation of China under Grant No. 62171038, No. 61827901, No. 61936011, and No. 62088101.
Authors also gratefully acknowledge the financial support of the CMIH and CCIMI University of Cambridge, and Graduate
school of Beijing Institute of Technology. 
AIAR gratefully acknowledges the financial support of the CMIH and CCIMI University of Cambridge. JL acknowledges support from the Leverhulme Trust, Shanghai Municipal Science and Technology Major Project (2021SHZDZX0102). CBS acknowledges support from the Philip Leverhulme Prize, the Royal Society Wolfson Fellowship, the EPSRC grants EP/S026045/1 and EP/T003553/1, EP/N014588/1, EP/T017961/1, the Wellcome Innovator Award RG98755, the Leverhulme Trust project Unveiling the invisible, the European Union Horizon 2020 research and innovation programme under the Marie Skodowska-Curie grant agreement No. 777826 NoMADS, the Cantab Capital Institute for the Mathematics of Information and the Alan Turing Institute. }

\vskip 0.2in

\end{document}